\documentclass{svjour3}
\usepackage{graphicx}
\usepackage{amsmath,amssymb,amsfonts} 
\usepackage{xspace} 
\usepackage{soul}  

\usepackage{multicol}
\usepackage[table]{xcolor}
 
\newcommand{\glob}[1]{{\cellcolor{red!25} #1}} 

\usepackage{tabu}
\usepackage{rotating}

\usepackage{url}

\newcommand\titlerr{\texttt{T}}
\newcommand\unirr{\texttt{UniRR}}
\newcommand\birr{\texttt{BiRR}}
\newcommand\trirr{\texttt{TriRR}}
\newcommand\casrr{\texttt{CasRR}}
\newcommand\trianglerr{\texttt{TglRR}}
\newcommand\unirrempht{\texttt{UniRR.T$^+$}}
\newcommand\casrrempht{\texttt{CasRR.T$^+$}}
\newcommand\trianglerrempht{\texttt{TglRR.T$^+$}}

\newcommand\LambertaN{\textsf{LamBERTa} (Law article mining based on BERT architecture)\xspace}
\newcommand\LambertaNbold{\textsf{LamBERTa} -- \textbf{L}aw \textbf{a}rticle \textbf{m}ining based on \textbf{BERT} \textbf{a}rchitecture\xspace}
\newcommand\Lamberta{\textsf{LamBERTa}\xspace}

\usepackage{lipsum}                     
\usepackage{xargs}                      
\usepackage[colorinlistoftodos,prependcaption,textsize=small]{todonotes}

\begin{document}
 
\title{Unsupervised Law Article Mining based on 
Deep Pre-Trained Language Representation Models  with Application to  the Italian Civil Code
\thanks{This article was published with the \textit{Artificial Intelligence and Law} journal, Springer Nature, on 15 September 2021, doi: https://doi.org/10.1007/s10506-021-09301-8}}

\titlerunning{Law article mining based on BERT architecture}
 
\author{Andrea Tagarelli \and 
Andrea Simeri   }
 
\authorrunning{A. Tagarelli and A. Simeri}
 
\institute{A. Tagarelli  \and A. Simeri \at Dept. Computer Engineering, Modeling, Electronics, and Systems Engineering (DIMES),\\ University of Calabria, 87036 Rende (CS), Italy\\
Corresponding Author: A. Tagarelli (\email{tagarelli@dimes.unical.it})
}

\date{}

\maketitle              
 
\begin{abstract} 
Modeling law search and retrieval as prediction problems has recently emerged as a predominant approach in law intelligence. Focusing on the law article retrieval task, we present a deep learning framework named LamBERTa, which is designed for civil-law codes, and specifically trained on the Italian civil code. To our knowledge, this is the first study proposing an advanced approach to law article prediction  for the Italian legal system based on a BERT (Bidirectional Encoder Representations from Transformers) learning framework, which has recently attracted increased attention among deep learning approaches, showing outstanding effectiveness in several natural language processing and learning tasks.  
We define LamBERTa models by fine-tuning an Italian pre-trained BERT on the   Italian civil code or its portions,  for law article retrieval as a classification task. 
One key aspect of our LamBERTa framework is that we conceived it to address an extreme classification scenario, which is characterized by a high number of classes, the few-shot learning problem, and the lack of test query benchmarks for Italian legal   prediction tasks. 
 To solve such issues, we define different methods for the unsupervised labeling of the law articles, 
 which can in principle be applied to any law article code system. 
 We provide insights into the explainability and interpretability of our LamBERTa models, 
 and we present an  extensive experimental analysis over query sets of different type, for single-label as well as multi-label evaluation tasks. Empirical evidence has shown the effectiveness of LamBERTa, and also its   superiority against widely used deep-learning text classifiers   and  a few-shot learner conceived for an attribute-aware   prediction task. 

\keywords{law article retrieval \and deep learning \and deep pre-trained language models \and BERT \and text classification  \and Italian civil code}
\end{abstract}

\section{Introduction}

The general purpose of law search is to recognize legal authorities that are relevant to a   question expressing a legal matter~\cite{Dadgostari2020}.  
The interpretative uncertainty in law, particularly that related to the jurisprudential type which is capable of directly affecting citizens, has prompted many to model law search as a prediction problem~\cite{Dadgostari2020}. Ultimately, this would allow lawyers and legal practitioners to explore  the possibility of predicting the outcome of a judgment (e.g., the probable sentence relating to a specific case),  through the aid of computational methods, also sometimes referred to as \textit{predictive justice}~\cite{Viola17}.  
   Predictive justice is currently being developed, to a prevalent extent, following a statistical-jurisprudential approach: the jurisprudential precedents are verified and future decisions are predicted on their basis. However, as stated by legal professionals~\cite{Viola17}, several reasons point out that this  approach should not be preferred,  
because of its limited scope only to cases in which there are numerous precedents, so as to exclude unprecedented cases relating to new regulations, not yet subject to stratified jurisprudential guidelines. 
 In fact, the jurisprudential approach  is not in line with any  \textit{civil law} system ---
 which is adopted in most European Union states and non-Anglophone countries in the world ---  
 with the consequence of a high risk of fallacy (i.e., repetition of errors based on precedents) and risk of standardization (i.e., if   a lawsuit is contrary to many precedents, then no one will propose such a lawsuit).

Clearly, from a major perspective as a \textit{data-driven artificial-intelligence} task, predicting judicial
 decisions is carried out exclusively based on the legal corpora available and the selected  algorithms to use, as remarked by Medvedeva et al.~\cite{MedvedevaVW20}. 
 Moreover, as also witnessed by an increased interest from the artificial intelligence and law research community, a key perspective in legal analysis and problem solving lays on  the opportunities given by advanced, data-driven computational approaches based on natural language processing (NLP), data mining, and machine learning disciplines~\cite{ConradB18}.

Predictive tasks in legal information systems have  often  been  addressed as text classification problems, ranging from case classification and legal judgment prediction~\cite{NallapatiM08,Liu2006,Lin2012,Aletras2016,Sulea2017,WangYNZZN18,MedvedevaVW20}, to legislation norm classification~\cite{Boella2011}, and  statute prediction~\cite{LIU2015194}. 
Early studies have focused on statistical textual features and  machine learning methods, then the progress of \textit{deep learning} methods for text classification~\cite{Goodfellow16,Goldberg17} has prompted the development of deep neural network frameworks, such as recurrent neural networks, 
for single-task learning (e.g., charge prediction~\cite{Luo2017,Ye2018}, sentence modality classification~\cite{ONeillBRO17,ChalkidisAM18}, legal question answering~\cite{DoNTNN17}) or even multi-task learning (e.g.,~\cite{YangJZL19,ZhouZLSS19}).

More recently, \textit{deep pre-trained   language models}, particularly the Bidirectional Encoder Representations from Transformers (BERT)~\cite{DevlinCLT19}, have   emerged showing outstanding effectiveness in several NLP tasks. Thanks to their ability to  learn  a contextual language understanding model, these  models overcome the need  for feature engineering (upon which classic, sparse vectorial representation models rely). Nonetheless, since they are originally trained from generic domain corpora, they should not be directly applied to a specific domain corpus, as the distributional representation (embeddings) of their lexical units may significantly shift from the nuances and peculiarities expressed in domain-specific texts --- and 
 this certainly holds for the legal domain as well, where  interpreting and relating   documents is particularly challenging.

Developing BERT models for legal texts has very recently attracted increased attention,  
mostly concerning  
 classification problems (e.g., \cite{RabeloKG19,ChalkidisAA19,SanchezHMAML20,ShaoMLMSZM20,legalBERT,YoshiokaAS21,abs-2106-13405}). 
 Our research  falls into this context, as we propose a BERT-based framework for   law article retrieval based on civil-law-based corpora.  
 More specifically, as we wanted  to benefit from the essential consultation provided by law professionals in our country,   
  our proposed framework is completely specified using   the Italian Civil Code (ICC) as  the target legal corpus.  
 Notably,  only   few  works have been developed for Italian BERT-based models, such as a retrained BERT for  various NLP tasks on Italian tweets~\cite{PolignanoBGSB19}, and  a    BERT-based  masked-language model for spell correction~\cite{PuccinelliDD19}; however, \textit{no study leveraging BERT for the Italian civil-law corpus has been proposed so far}.

Our main  contributions  in this work are summarized as follows:  
\begin{itemize}
\item 
We push forward research  on law document analysis  for civil law systems, focusing on the modeling, learning and understanding of  logically coherent corpora of law articles, using the Italian Civil Code as case in point.  
\item We study the law article retrieval task as a prediction problem based on the deep machine learning paradigm. More specifically, following the lastest advances in research on deep neural network models for text data,     we  propose a deep pre-trained contextualized language model framework, named  \LambertaN. \Lamberta  is designed to fine-tune an Italian pre-trained BERT on the ICC corpora  for law article retrieval as prediction, i.e., given a natural language query, predict the most relevant   ICC article(s). 
\item Notably, we deal with  a very challenging prediction task, which is characterized not only by a high number (i.e., hundreds) of classes --- as many as the number of  articles --- but also by the issues that arise from  the need for  building suitable training sets given the lack of test query benchmarks for Italian legal article retrieval/prediction tasks.  This also leads to coping with few-shot learning issues (i.e., learning models to  predict the correct class of instances when a small amount of examples are available in the training dataset),  which has been recognized as one of the  so-called \textit{extreme classification} scenarios~\cite{xClass,Chalkidis-Extreme}.  
 We design our \Lamberta framework to solve such issues based on different schemes of \textit{unsupervised training-instance  labeling} that we originally  define for the ICC corpus, although they can easily be generalized to other law code systems.   
\item We address one crucial aspect that typically arises in deep/machine learning models, namely \textit{explainability}, which is clearly  of   interest also in artificial intelligence and law (e.g.,~\cite{BrantingWBPCFPY19,HackerKGN20}). In this regard, we investigate  explainability of our \Lamberta models focusing on the understanding of how they form complex relationships between the textual tokens. 
We further provide insights into the patterns generated by   \Lamberta models through a visual exploratory analysis of the learned representation embeddings.
 \item 
 We present an extensive, quantitative experimental analysis of \Lamberta models by considering:
\begin{itemize}
\item six different types of test queries, which vary  by originating source,  length and lexical characteristics, and include   \textit{comments} about the ICC articles as well as  
\textit{case law decisions} from the civil section of the Italian Court of Cassation that contain significant jurisprudential sentences associated with the ICC articles;
\item   \textit{single-label} evaluation as well as \textit{multi-label} evaluation tasks;
\item different sets of assessment criteria. 
\end{itemize}
   The obtained results have shown    the effectiveness of  \Lamberta, and    its   superiority against (i) widely used deep-learning text classifiers  that have been tested on our different query sets for the article prediction tasks, and against (ii) a few-shot learner conceived for an attribute-aware   prediction task that we have newly designed based on the availability of ICC metadata. 
\end{itemize}

\vspace{2mm}
The remainder of the paper is organized as follows. 
Section~\ref{sec:related} overviews recent works that address legal classification and retrieval problems based on deep learning methods. 
Section~\ref{sec:data} describes the ICC corpus, and Section~\ref{sec:models} 
presents  our proposed framework for the civil-law article retrieval problem. 
Sections~\ref{sec:attention} and \ref{sec:visual-embeddings} are devoted to  qualitative investigations on the explainability and interpretability of \Lamberta models. 
Quantitative experimental evaluation methodology and results are instead presented in Sections~\ref{sec:evaluation} and \ref{sec:results}. 
Finally, 
Section~\ref{sec:conclusion} concludes the paper.

\section{Related Work}
\label{sec:related}

Our work belongs to the corpus of studies that reflect  the recent revival of interest in the role that machine learning, particularly deep neural network models,  can take on  artificial intelligence applications for text data in a variety of domains, including the legal one.  
 In this regard, here   we overview recent research works that employ deep learning methods for addressing computational problems in the legal domain, with a focus on classification and retrieval tasks.  Note that the latter are major categories for the data-driven legal analysis literature review, along with entailment and information extraction based on NLP approaches (e.g.,  named entity recognition, relation extraction, tagging), as extensively studied by Chalkidis and Kampas in~\cite{ChalkidisK19}, to which we refer the interested reader for a broader overview.

Most existing works on deep-learning-based law analysis exploit recurrent neural network models (RNNs) and convolutional neural networks (CNNs), along with the classic multi-layer perceptron (MLP).  For instance, 
O'Neill et al.~\cite{ONeillBRO17} utilize all the above methods for
  classifying deontic modalities in regulatory texts, demonstrating superiority of neural network models over  competitive classifiers including ensemble-based decision tree and largest margin classifiers.   Focusing on obligation and prohibition extraction as a particular case of deontic sentence classification,     
Chalkidis et al.~\cite{ChalkidisAM18} show the benefits of employing a hierarchical attention-based bidirectional LSTM model that considers both the sequence of words in each sentence and the sequence of sentences. 
Branting et al. \cite{BrantingYWMB17}  consider administrative adjudication prediction  in        motion-rulings, Board of Veterans Appeals issue decisions, and World Intellectual Property Organization 
domain name dispute decisions. In this regard, three approaches for prediction are evaluated: maximum 
entropy over token n-grams, SVM over token n-grams, and a hierarchical
attention network~\cite{YangYDHSH16} applied to the full text. While no absolute winner was observed, the study highlights the benefit of using  
feature weights or network attention weights from these
predictive models   to identify salient phrases in motions or contentions and case facts.  
Nguyen et al.~\cite{NguyenNTSS17,NguyenNTSS18} propose  several approaches to train long short term memory  (LSTMs) models and conditional random field (CRF) models for the problem of identifying two key portions of   legal documents, i.e., requisite and effectuation segments,   with evaluation on 
Japanese civil code and Japanese National Pension Law dataset.   
 In~\cite{ChalkidisK19} by Chalkidis and Kampas, a major contribution is the development of word2vec skip-gram embeddings trained on large legal corpora (mostly from European, UK, and US legislations).

Note that our work is clearly different  from the aforementioned ones, since they not only focus on legal corpora other than Italian civil law articles but also they consider machine learning and neural network models that do not exploit the same ability as pre-trained deep language models.

To address the problem of   predicting the final charges according to the fact descriptions in criminal cases,  Hu et al.~\cite{HuLT0S18} propose to exploit    a set of categorical attributes to discriminate among charges (e.g., violence, death, profit purpose, buying and selling). By leveraging these   annotations of charges based on representative attributes, the proposed learning framework aims to predict  attributes and charges of a case simultaneously. 
An attribute attention mechanism is first applied   to select   factual information from facts that are relevant to each particular attribute, so to generate attribute-aware fact representations that can be used to predict the label of an attribute, under  a binary classification task. Then, for the task of charge prediction, the  attribute-aware fact representations aggregated by average pooling are also concatenated  with the  attribute-free  fact representations produced by a conventional LSTM neural network. The training objective is twofold, as it minimizes the cross-entropy loss of charge prediction and the cross-entropy loss of attribute prediction. 

It should be noticed that the above study was especially designed to deal with the typical imbalance of the case numbers of various charges as well as to distinguish related or ``confusing'' charges. In particular, the first aspect corresponds to a challenge of  insufficient training data for some charges, as there are indeed charges with limited cases. This appears to be in analogy with the few-shot learning scenario in law article prediction;  therefore, in Section~\ref{sec:comparison}, we shall present  a comparative evaluation stage with the method in~\cite{HuLT0S18} adapted for the ICC law article prediction task.

Also in the context of legal judgment prediction, Li et al.~\cite{LiZYGF19} propose a multichannel attention-based neural network model, dubbed MANN, that exploits not only the case facts but also information on 
the defendant persona, such as traits that determine the criminal liability (e.g., age, health condition, mental status) and criminal records. 
A two-tier structure is used to empower attention-based 
sequence encoders to hierarchically model the semantic interactions from different parts of case description. 
Results on datasets of criminal cases in mainland China have shown improvements over other neural network models for judgment prediction, althugh MANN may suffer from   the imbalanced classes of prison terms and cannot deal with   criminal cases with multiple defendants.  
More recently, Gan et al.~\cite{GanKYW21} have proposed to  inject legal knowledge
into a neural network model to improve performance and interpretability of legal judgment prediction. The key idea is to model declarative legal knowledge as a set of first-order logic rules and integrate these logic rules into a co-attention network based model (i.e., bidirectional information flows between facts and claims) in an end-to-end way.  
The method has been evaluated on a collection of private loan law cases, where each instance in the dataset consists of a fact description and the plaintiff’s multiple claims, demonstrating some advantage over  
AutoJudge~\cite{LongT0S19}, which  models the   interactions between claims and fact descriptions via pair-wise attention in a judgment prediction task.
 
 The above two works are distant from ours, not only in terms of the target corpora and addressed problems but also since we do not use any type of  information other than the text of the articles, nor any injected knowledge base.

In the last few years, the Competition on Legal Information Extraction/
Entailment (COLIEE) has been an important venue for displaying studies focused on case/statute law retrieval, entailment, and question answering.  
 In most works appeared in the most recent COLIEE editions,   the observed trend is to tackle  the retrieval task by using CNNs and RNNs in the entailment phase, possibly in combination of  additional features produced by applying classic term relevance weighting methods (e.g., TF-IDF, BM25) or statistical topic models (e.g., Latent Dirichlet Allocation). 
 For instance, Kim et al.~\cite{KimXG15} propose a binary CNN-based classifier model   for answering to the legal queries in the entailment phase. The entailment model introduced by Morimoto et al.~\cite{MorimotoKSSM17}  is instead based on MLP incorporating the attention mechanism. 
 Nanda et al.~\cite{NandaACBR17} adopt a combination of   partial string matching and   topic clustering for the retrieval task, while they combine LSTM and CNN models for the entailment phase. 
 Do et al.~\cite{DoNTNN17} propose 
a CNN binary model with additional TF-IDF  and statistical latent semantic features.

The aforementioned studies differ from ours as they mostly focus on CNN and RNN based neural network models, which are indeed used as competing methods against our proposed deep-pretrained language model framework (cf.    Section~\ref{sec:comparison}).

Exploiting BERT for law classification  tasks has recently attracted much attention.   Besides a study on Japanese legal term correction proposed by Yamakoshi et al.~\cite{YamakoshiKOT19},  a few very recent works address sentence-pair classification problems in legal information retrieval and entailment scenarios. 
Rabelo et al.~\cite{RabeloKG19} propose to combine 
 similarity based features and BERT fine-tuned to the task of case law entailment on the   data provided by the   Competition on Legal Information Extraction/Entailment (COLIEE), where the input is an entailed fragment from a case coupled with a candidate entailing paragraph from a noticed case.     
Sanchez et al.~\cite{SanchezHMAML20}  employ BERT in its regression form to learn complex relevance criteria to support legal search over news articles.  Specifically, the input consists of a query-document pair, and the output is a predicted relevance score. Results have shown that BERT trained either on a combined title and  summary field of documents 
or on the documents' contents outperform  a learning-to-rank approach based on LambdaMART equipped with features engineered upon three groups of relevance criteria, namely topical relevance, factual
information, and language quality.  
However,  in legal case retrieval,   the query case is typically much longer and more complex than common keyword queries, and  the definition of relevance between a query case and a supporting case could be beyond general
topical relevance, which makes it  difficult to build  a large-scale case retrieval dataset. To address this challenge, 
Shao et al.~\cite{ShaoMLMSZM20} propose a BERT framework to  model semantic relationships to infer the relevance between  two cases by aggregating paragraph-level dependencies. 
  To this purpose, the BERT model  is fine-tuned
 with a relatively small-scale case law entailment dataset to adapt it to the legal scenario. 
 Experiments conducted on the benchmark of the relevant
case retrieval task in COLIEE 2019 have shown effectiveness of the proposed BERT model.  

We notice that the above two works require to   classify  query-document  pairs (i.e., pairs of query and news article, in~\cite{SanchezHMAML20}, or   pairs of case law documents, in~\cite{ShaoMLMSZM20}), whereas our models are trained by using articles only.  
At the time of submission of this article, we also become aware of a small bunch of works presented at  COLIEE-2020 that compete for the statute law retrieval and question answering (statute entailment) tasks\footnote{https://sites.ualberta.ca/~rabelo/COLIEE2020/.} 
 using BERT~\cite{RabeloKGYKS20}. 
In particular, for the statute law retrieval task, the goal is to read a legal bar exam question and retrieve a subset of Japanese civil code articles to judge whether the question is entailed or not. The BERT-based approach has shown to improve overall retrieval performance, although there are still numbers of questions that are difficult to retrieve by BERT too.  
At COLIEE-2021, which was held in June 2021,\footnote{https://sites.ualberta.ca/~rabelo/COLIEE2021/.}  there has been an increased attention and development of  BERT-based methods to address the statute and case law processing tasks~\cite{YoshiokaAS21,abs-2106-13405}.
 
Again, it should be noted that the above works at COLIEE Competitions assume that the training data are   questions and relevant article pairs, whereas our training data instances are derived from articles only. 
Nonetheless, we recognize that some of the techniques introduced at COLIEE-2021, such as   deploying weighted aggregation on models’ predictions and the iterated self-labeled and fine-tuning process, are worthy of investigation and we shall delve into them in our future research.

 In addition to the COLIEE Competitions, it is worth mentioning the study by 
Chalkidis et al.~\cite{ChalkidisAA19}, which provides a threefold contribution. They release  a new dataset of cases from the European Court of Human Rights (ECHR) for legal judgment prediction, which is larger (about 11.5k cases) than earlier datasets, and such that each case along with its list of facts  is mapped to articles violated (if any) and is assigned an ECHR importance score. The dataset is used to evaluate a selection of neural network models, 
for different tasks, namely binary
classification (i.e., whether a  case violates  a human rights article or not), multi-label classification (i.e., which types of violation, if any),  and case importance detection. 
Results have shown that the neural network models outperform an SVM model  with bag-of-words features, which was previously used in related work such as~\cite{Aletras2016}.  
 Moreover, a hierarchical version of BERT, dubbed HIER-BERT, is proposed to overcome the BERT's maximum length limitation, by first generating fact embeddings and then using them through a self-attention mechanism to produce case embeddings, similarly to a hierarchical attention network model~\cite{YangYDHSH16}.

The latter aspect on the use of a hierarchical attention mechanism, especially when integrated into BERT, is very interesting and useful on long legal documents, such as   case law documents,   to improve the performance of pre-trained language models like BERT that are designed with constraints on the tokenized text length. Nonetheless, as we shall discuss later in Section~\ref{sec:data-preparation}, this contingency does not represent an issue in   the setting of our  proposed framework, due not only to the characteristic length of ICC articles but also to our designed schemes of  unsupervised training-instance labeling.

\section{Data}
\label{sec:data}

The  Italian Civil Code (ICC)  is  
  divided into six, logically coherent books, each in charge of providing rules for a particular civil law   theme:   
\begin{itemize}
\item
\textit{Book-1}, on Persons and the Family, articles 1-455 -- contains the discipline of the juridical capacity of persons, of the rights of the personality, of collective organizations, of the family;
\item
\textit{Book-2}, on Successions, articles 456-809 -- contains the discipline of succession due to death and the donation contract;
\item
\textit{Book-3}, on Property, articles 810-1172 -- contains the discipline of ownership and other real rights;
\item
\textit{Book-4}, on Obligations, articles 1173-2059 -- contains the discipline of obligations and their sources, that is mainly of contracts and illicit facts (the so-called civil liability);
\item
\textit{Book-5},  on Labor, articles 2060-2642 -- contains the discipline of the company in general, of subordinate and self-employed work, of profit-making companies and of competition;
\item
\textit{Book-6}, on the Protection of Rights, articles 2643-2969 -- contains the discipline of the transcription, of the proofs, of the debtor's financial liability and of the causes of pre-emption, of the prescription.
\end{itemize}

The articles of each book are internally organized into a hierarchical structure based on four levels of division, namely (from top to bottom in the hierarchy):  ``titoli'' (i.e., chapters), ``capi'' (i.e., subchapters), ``sezioni'' (i.e., sections), and  ``paragrafi'' (i.e., paragraphs).  
 It should however be emphasized that this hierarchical classification was not  meant as a crisp, ground-truth   organization of the articles' contents: indeed, the topical boundaries of contiguous chapters and subchapters are often quite smooth, as articles in the same group often not only vary in length but can also provide dispositions that are more related to articles in other groups.

The ICC  is obviously publicly available, in various digital formats. 
From one of such sources, we extracted \textit{article id}, \textit{title} and  \textit{content} of each article.  
We   cleaned up the text from non-ASCII characters, removed numbers and date, 
 normalized all variants and abbreviations of frequent keywords such as ``articolo'' (i.e., article),   
 ``decreto legislativo'' (i.e., legislative decree), ``Gazzetta Ufficiale'' (i.e., Official Gazette), and finally we lowercased all letters.  
 
The ICC currently in force was enacted by Royal decree no. 262 of 16 March 1942, and it consists of 2969 articles. This number actually corresponds  to 3225 articles considering all variants and subsequent insertions, which are designated by using Latin-term  suffixes (e.g., ``bis'', ``ter'', ``quater'').  
 However, during its history,  the ICC was revised several  times and subjected to repealings, i.e., per-article partial or total insertions, modifications and removals; to date, 
2294 articles have been repealed. 
  Table~\ref{tab:data-stats} summarizes main statistics on the preprocessed ICC  books.

\begin{table}[t!]
\centering
\caption{Main statistics on the ICC corpus and   its constituent books. }
\label{tab:data-stats}
\scalebox{0.85}{
\begin{tabular}{|l||c|c|c|c|c|c|c|c|c|}
\hline 
ICC  & \# arts. & \multicolumn{4}{|c|}{\# sentences over the articles} & \multicolumn{4}{|c|}{\# words over the articles}  \\ \cline{3-10}
portion &  & \textit{tot.} & \textit{min} & \textit{max} & \textit{mean} (\textit{std}) & \textit{tot.} & \textit{min} & \textit{max} & \textit{mean} (\textit{std})   \\ \hline \hline
 
\textsl{Book-1} & 395 & 1979 & 3 & 21 & 5.010 (2.323) &  32354 & 11 & 569 & 81.909 (71.952)   
\\  \hline 

\textsl{Book-2} & 345 & 1561 & 3 & 13 & 4.525 (1.675) & 24520 & 9 & 354 &  71.072 (51.366)  
\\  \hline 

\textsl{Book-3} & 364 &  1619 & 3 &  24 &  4.448 (1.816) & 25971 & 6 & 893 & 71.349 (65.836)  
\\  \hline 

\textsl{Book-4} &  891 & 3595 & 3 &  12 & 4.035  (1.338) &  50509 & 7 & 365 & 56.688 (38.837)  
\\  \hline 

\textsl{Book-5} &  713 & 3937 &  3 &  37  &  5.522 (3.191) & 75764 &  8 & 1465 & 106.261 (117.393)  
\\  \hline 

\textsl{Book-6} &  331 & 1453 & 3 & 17 &  4.390  (1.895) &  25937 &  12 & 654 & 78.360 (76.954) 
\\  \hline 

All &  3039 & 14131 &  3 & 37 & 4.650 (2.243) & 234945 & 6 & 1465 &  77.310 (78.373)  
\\  \hline 
\end{tabular}
} 
\end{table}

\section{The Proposed \Lamberta Framework}
\label{sec:models}
 
In this section we present our proposed learning framework for   civil-law article retrieval. We first formulate the problem statement in Section~\ref{sec:problem-formulation} and overview the framework in Section~\ref{sec:overview}. We present our devised learning approaches  in Section~\ref{sec:models-app}. We describe the data preparation and preprocessing in Section~\ref{sec:data-preparation}, then in Section~\ref{sec:schemes} we define our unsupervised training-instance labeling methods for the   articles in the target corpus.
Finally, in Section~\ref{sec:learning-config}, we discuss major settings of the proposed framework.

\subsection{Problem setting}
\label{sec:problem-formulation}
 
Our study is concerned with   law article retrieval, i.e.,  finding articles of interest out of a legal corpus that can be recommended as an appropriate response to a query expressing a legal matter.  

To formalize this problem, we assume that any query is expressed  in natural language and discusses a legal subject  that is in principle covered by the target legal corpus (i.e., the ICC, in our context). Moreover, a query is assumed to be free of references to any article identifier in the ICC.  

We address the law article retrieval task based on the supervised machine learning paradigm:  given a new, user-provided instance, i.e., a legal question, the goal is to automatically predict the category associated to the posed question. More precisely, we deal with the more general case in which a probability distribution over all the predefined categories is computed in response to a query. 
The prediction is carried out by a machine learning system that is trained  on a target legal corpus --- whose documents, i.e., articles, are annotated with the actual category they belong to --- in order to learn a computational model, also called classifier, that will then be used to perform the predictions against legal queries by exclusively utilizing the textual information contained in the annotated documents. 

\vspace{1mm}
\noindent 
{\bf Motivations for BERT-based approach.\ } 
In this respect, our objective is to leverage   deep neural-network-based, pre-trained language modeling to solve the law article retrieval task. 
This has a number of key advantages that are summarized as follows. First, like any other deep neural network models, it   totally avoids manual feature engineering, and hence the need for employing feature selection methods as well as feature relevance measures (e.g., TF-IDF). 
Second, like  sophisticated recurrent and convolutional neural networks, it models language semantics and non-linear relationships between terms; however,     better than recurrent and convolutional neural networks and their combinations, it is able to capture subtle and  complex lexical patterns  including the sequential structure and long-term dependencies, thus obtaining   the most comprehensive local and global feature representations of a text sequence. 
Third,  it incorporates the  so-called \textit{attention} mechanism, which allows a learning model  to assign higher weight to text features according to   their higher informativeness or relevance to the learning task.  
Fourth, being a real bidirectional \textit{Transformer} model, it overcomes the main limitations of early deep contextualized models like ELMO~\cite{Peters2018} (whose left-to-right language model and right-to-left language model are actually independently trained) or decoder-based Transformer models,  like OpenAI GPT~\cite{GPT}.

\vspace{1mm}
\noindent 
{\bf Challenges.\ } 
 It should however be noted that, like any other machine learning method,  using deep pre-trained models like BERT for classification tasks normally requires the availability of data annotated with the class labels, so to design the independent training and testing phases for the  classifier. 
 However, this does not apply to our context. 
  
As previously mentioned,   in this work we face three main challenges: 
\begin{itemize}
\item
the first challenge refers to the high number (i.e., hundreds)  of classes, which correspond to  the number of   articles in the ICC corpus, or portion of it, that is used to train a \Lamberta model; 
\item 
the second challenge corresponds  to the so-called  \textit{few-shot learning}   problem, i.e., dealing with a small amount of per-class examples to train a machine learning model, which Bengio et al. recognize as one of    the  ``extreme  classification'' scenarios~\cite{xClass}. 
\item
 the third challenge derives from the unavailability of test query benchmarks for Italian legal article retrieval/prediction tasks. This has prompted us to   define appropriate methods for data annotation, thus for building up training sets for the \Lamberta  framework.  
 To address this problem, we originally define   different schemes of \textit{unsupervised training-instance labeling}; notably, these are not ad-hoc defined for the ICC corpus, rather  they can be adapted to any other law code corpus. 
  \end{itemize}

In the following, we overview our proposed deep pre-trained model based framework that is designed to address all the above challenges.

\subsection{Overview of the \Lamberta framework}
\label{sec:overview} 

\begin{figure}[t!]
\centering
\includegraphics[scale=0.28]{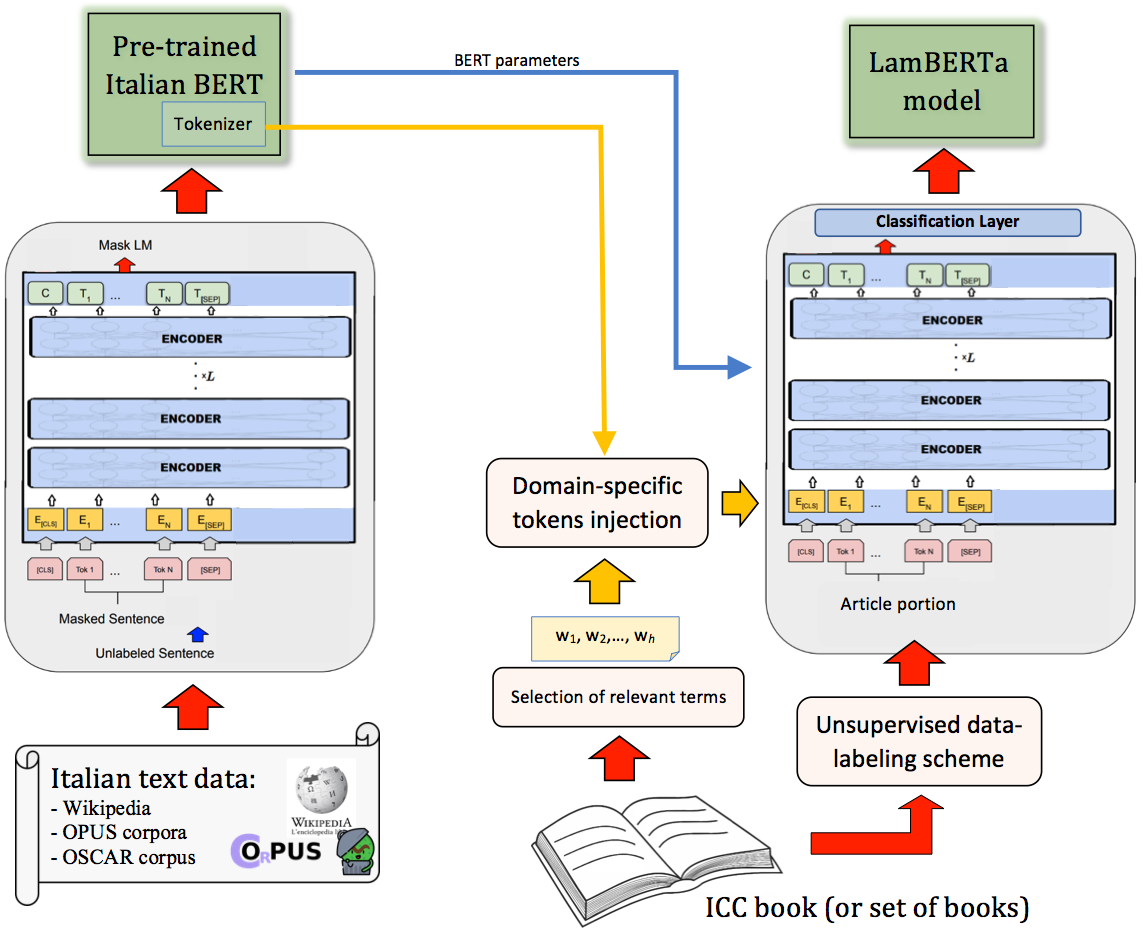}
\caption{An illustration of the conceptual architecture of \Lamberta   designed for the ICC law article retrieval task.}
\label{fig:architecture}
\end{figure}

Figure~\ref{fig:architecture} shows the conceptual architecture of our proposed \LambertaNbold. 
The starting point is a pre-trained Italian BERT model whose source data consists of a recent Wikipedia dump, various texts from the OPUS corpora collection, and the Italian part of the OSCAR corpus; the final training corpus has a size of 81GB and 13\,138\,379\,147 tokens.\footnote{bert-base-italian-xxl-uncased, available at https://huggingface.co/dbmdz/.}  

\Lamberta models are generated by fine-tuning the pre-trained Italian BERT model on a sequence classification task (i.e., BERT with a single linear classification layer on top) given in input the  articles of  the ICC or a portion of it.  

 It is worth noting that the \Lamberta architecture is versatile w.r.t.   the adopted learning approach and the training-instance  labeling scheme for a given corpus of ICC articles. These aspects are elaborated on next.  
 
\subsection{Global and Local learning approaches}
\label{sec:models-app}

We consider two learning approaches, here dubbed   \textit{global} and \textit{local} learning, respectively.  
A \textbf{global model} is trained on the whole ICC, whereas a  \textbf{local model} is trained on a particular book of the ICC.\footnote{From a technical viewpoint,   the considered portion could in principle include any strict subset of the ICC; however, on the legal domain side, it is reasonable to learn local models each dedicated  to a logically coherent subset of the whole civil code, i.e., a single book.}    
 Our rationale underlying this choice is as follows: 
 \begin{itemize}
 \item 
 on the one hand, local models are designed to embed the logical coherence of  the articles within  a particular book and, although limited to its corresponding topical boundaries, they are expected to leverage the multi-faceted semantics  underlying  a specific civil law theme (e.g., inheritage); 
 \item on the other hand, books are themselves part of  the same law code, and hence a global model might be useful  to capture possible interrelations between the single books, however, by embedding different topic signals from different books (e.g. inheritage of Book-2 vs. labor law of Book-5), it could incur the risk of topical dilution over all the ICC. 
 \end{itemize} 

Either type of model is designed to be a \textit{classifier at article level}, i.e., class labels correspond to the articles in the book(s) covered by  the model. 
 Given the one-to-one association between classes and articles,   a question becomes how to create suitable training sets for our \Lamberta models. Our key idea is to adopt an \textit{unsupervised annotation} approach, which is discussed in the next section.

\subsection{Data preparation}
\label{sec:data-preparation}

To tailor the ICC articles to the BERT input format, we initially carried out segmentation of the content of each article into sentences. This was then followed by tokenization of the  sentences  and text encoding, which are described next. 
 
\vspace{1mm}
\noindent 
{\bf Domain-specific terms injection and Tokenization.\ }
BERT was trained using the WordPiece tokenization. This  is an effective way to alleviate the open vocabulary problems in neural machine translation, since a word can be broken down into  sub-word units, 
 which are constructed during the training time and depend on the corpus the model was initially trained on. 
 
 However, when retraining BERT on the ICC articles to learn   global as well as  local models, it is likely that some important terms occurring in the legal texts are missing in the pre-trained lexicon; therefore, to make BERT aware of the domain-specific (i.e., legal) terms and  avoid  subwording such terms thus disrupting their semantics, we injected a selection of terms from the ICC articles into the pre-existing BERT vocabulary before tokenization.\footnote{From the perspective of the BERT architecture, this implies two actions, namely indexing the added terms and initializing their corresponding weights.}   
 
 To select the domain-specific terms to be added,  we carried out the following steps: if we denote with $D$ the  input (portion of) ICC text for the model to learn, we first  preprocessed  the text $D$ as described in Sect.~\ref{sec:data}, then we  removed Italian stopwords,  and finally filtered out overly frequent terms (as occurring in more than 50\% of the articles in $D$) as well as hapax terms.   
   Table~\ref{tab:data-stats-added} reports the number of added tokens and the final number of tokens, for each input corpus to our \Lamberta local and global models.

\begin{table}[t!]
\centering
\caption{Number of domain-specific tokens to inject into the pre-trained BERT vocabulary and the final vocabulary size. }
\label{tab:data-stats-added}
\scalebox{0.85}{
\begin{tabular}{|l||c|c|}
\hline 
ICC  &   \# added    & vocab.  \\  
portion & tokens & size \\ \hline \hline
 
\textsl{Book-1} &     833 & 31935
\\  \hline 

\textsl{Book-2}   & 698 & 31800  
\\  \hline 

\textsl{Book-3}  & 1072 & 32174
\\  \hline 

\textsl{Book-4}   & 1383 & 32485
\\  \hline 

\textsl{Book-5}   & 2048 & 33150
\\  \hline 

\textsl{Book-6}  & 829 & 31931
\\  \hline 

All & 3993 & 35095
\\  \hline 
\end{tabular}
} 
\end{table}

\vspace{1mm}
\noindent 
{\bf Text encoding.\ }  BERT utilizes a fixed sequence size (usually 512) for  each tokenized text, which implies both padding of shorter sequences and truncation of longer sequences. 
 While padding has no side effect, truncation may produce loss of information. 
 
 In learning our models, we wanted to avoid the above aspect, therefore we investigated any condition causing such undesired effect in our input data. We found out that this contingency is very rare in each book of the ICC, even reducing the maximum length to 256 tokens; the only situation that would lead to truncation corresponds to an article sentence that is logically organized in multiple clauses separated by semicolon: for these cases, we treated each of the subsentences as one training unit associated with the same article class-label.

\subsection{Methods for   unsupervised   training-instance labeling}
\label{sec:schemes}

As previously discussed,    our \Lamberta classification models are trained in such a way that it holds  an one-to-one correspondence between articles in a target corpus and class labels. Moreover, the entire corpus must be used to train the model so to embed the whole knowledge therein.  However, given the uniqueness of each article,   the general problem we face is  \textit{how to create as many training instances as possible  for each article} to effectively train the models.

Our key idea is to select and combine portions of each article to generate the training units for it. 
To this aim,   we devise  different unsupervised schemes for data labeling the ICC articles to create the training sets of \Lamberta models.  Our schemes adopt different strategies for selecting and combining portions of each article to derive the training sets, but they share the requirements of generating a minimum number of training units per article, here denoted as   $minTU$; moreover, since each article   is usually comprised of few sentences, and $minTU$ needs to be relatively large (we chose 32 as default value), each of the schemes implements a \textit{round-robin} (RR)  method that iterates over replicas of the same  group of training units per article until at least $minTU$ are generated. 

In the following, we define our methods for unsupervised training-instance labeling of the ICC articles (in square brackets, we indicate the notation that will be used throughout the remainder  of the paper):

\begin{itemize} 
\item  
\textit{Title-only} [\titlerr].  This is the simplest yet lossy scheme, which keeps an article's title while  discarding its content; the round-robin block is just the title of an article.\footnote{We point out that considering also the title of the group to which an article belongs (e.g., chapter, subchapter) would not add further information, since each article's title recalls yet specializes its group's title.}  
\item  
\textit{$n$-gram}. Each training unit   corresponds to $n$ consecutive sentences of an article; the round-robin block starts with the $n$-gram containing  the title and ends with the $n$-gram  containing the last sentence of the article. We set~$n \in \{1,2,3\}$, i.e., we consider a  unigram [\unirr], a bigram [\birr], and a trigram [\trirr] model,~respectively.   
\item   
\textit{Cascade} [\casrr]. The article's sentences are cumulatively selected to form the training units; the round-robin block starts with the first sentence (i.e., the title), then the first two sentences, and so on until all article's  sentences are considered to form  a single training unit. 
\item  
\textit{Triangle} [\trianglerr]. Each training unit is either an unigram, a bigram or a trigram, i.e., the round-robin block contains all  $n$-grams, with $n \in \{1,2,3\}$, that can be extracted from the article's title and description.  
\item   
\textit{Unigram with parameterized emphasis on the title} [\unirrempht]. The set of training units is comprised of one subset containing the article's sentences with round-robin selection, and another subset containing only replicas of the article's title.   More specifically, the two subsets are formed as follows:
\begin{itemize}
\item The first subset is of size equal to the maximum between the number of article's  sentences and the quantity $m \times mean\_s$, where  $m$ is a multiplier (set to 4 as default) and $mean\_s$ expresses the average number of sentences per article, excluding the title. As reported in Table~\ref{tab:data-stats} (sixth column), this mean value can be recognized between 3 and 4 -- recall that the title is excluded from the count --  therefore we set $mean\_s \in \{3,4\}$.
\item The second subset finally contains $minTU - m \times mean\_s$ replicas of the  title. 
\end{itemize} 
\item  
  \textit{Cascade with parameterized emphasis on the title} [{\casrrempht}] and \textit{Triangle with parameterized emphasis on the title} [\trianglerrempht]. These two schemes follow the same approach as \unirrempht except for the composition of the round-robin block, which corresponds to {\casrr}  and \trianglerr, respectively, with the title  left out from this block and replicated in the second block, for each article.
\end{itemize}

\subsection{Learning configuration} 
\label{sec:learning-config}

 An input to \Lamberta  
  is of the form [CLS, $\langle$\textit{text}$\rangle$, SEP], 
 where CLS (stands for `classification') and SEP are special tokens respectively at the beginning of  each sequence and at separation of two parts of the input. 
 These two special tokens are associated with two vectorial dense representations or \textit{embeddings}, denoted as E$_{[\mathrm{CLS}]}$ and E$_{[\mathrm{SEP}]}$, respectively; analogously, there are as many embeddings E$_i$ as the number of tokens  of the input text.   Note also that as   discussed in Sect.~\ref{sec:data-preparation},  each sequence in a mini-batch is padded to the maximum length (e.g., 256 tokens) in the batch. 
 In correspondence of each input embedding E$_i$, an embedding T$_i$ is outputted, and it can be seen as  the contextual representation of the $i$-th  token of the input text. 
 Instead, the final hidden state, i.e., output embedding C,  corresponding to the CLS token captures the high level representation of the entire text to be used as input to further  levels in order to train any sentence-based task, such as sentence classification.  
 This vector, which is by design  of size 768,  
  is fed to a single layer neural network followed by sigmoid activation whose output represents the distribution probability over the law articles of a book.

\Lamberta models were trained using a typical configuration of BERT for masked language modeling, with 12 attention heads and 12 hidden layers, and initial (i.e., pre-trained) vocabulary   of 32\,102 tokens. 
     Each model was trained for 10 epochs, using cross-entropy as loss function, Adam optimizer and initial learning rate selected within [1e-5, 5e-5]  
 on batches of  256  examples. 
 
 It should be noted that our \Lamberta models can be seen as relatively light in terms of computational requirements, since we developed  them under the Google Colab GPU-based    environment with a limited memory of 12 GB and 12-hour limit   for   continuous assignment of the virtual machine.\footnote{https://colab.research.google.com/.}

\section{Explainability of \Lamberta models based on Attention Patterns}
\label{sec:attention}

\begin{figure}[t!]
\centering
\begin{tabular}{ccc}
\hspace{-5mm}
\includegraphics[scale=0.3]{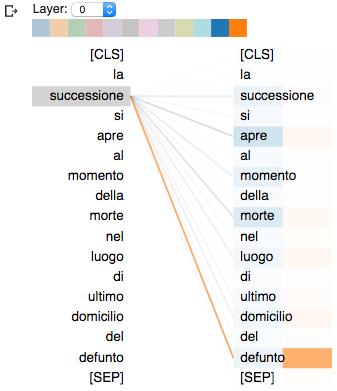} & 
\hspace{-5mm}
\includegraphics[scale=0.33]{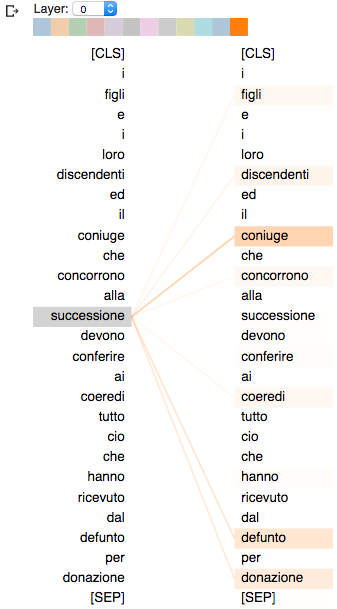} &
\hspace{-5mm}
\includegraphics[scale=0.33]{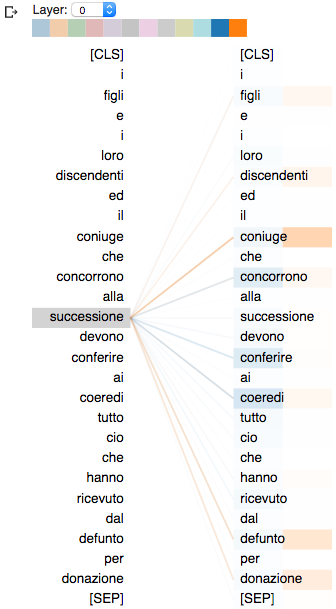}    
\\
(a) & (b) & (c) 
\end{tabular}
\caption{Attention patterns for   ``succession'': (a) two-head pattern from Art. 456 and comparison with  (b) single-head and (c) two-head patterns from Art. 737.}
\label{fig:patterns12}
\end{figure}

 In this section, we     start our understanding of what  is going on inside the ``black box'' of \Lamberta models. 
 One important aspect is the  \textit{explainability} of \Lamberta models, which we investigate here focusing on  the formation of complex relationships between tokens.

Like any BERT-based architecture, our \Lamberta leverages the Transformer paradigm, which allows for processing all elements simultaneously by forming direct connections between individual elements through a mechanism known as \textit{attention}. 
Attention is a way for a model to assign weight to input features (i.e., parts of the texts) based on their high informativeness or importance to some task.  
 In particular, attention enables the model to understand how the words relate to each other in the context of the sentence, by forming composite representations that the model can reason about.

BERT's attention patterns can assume several forms, such as delimiter-focused, bag-of-words, and next-word patterns. 
Through the lens of the \texttt{bertviz} visualization tool,\footnote{https://github.com/jessevig/bertviz.} we 
show how \Lamberta forms its distinctive attention patterns.   
For  this purpose,  
 here we present a selection of examples built upon sentences from Book-2 of the ICC (i.e., relevant to key concepts in inheritance law), which are next reported     both in Italian and English-translated versions.

The  attention-head view in \texttt{bertviz} visualizes  attention patterns  as lines connecting the word being updated (left) with the word(s) being attended to (right), for any given input sequence, where color intensity reflects the attention weight. 
Figure~\ref{fig:patterns12}(a)-(c) shows noteworthy examples focused on the word ``succession'', from the following sentences:

\vspace{1mm}
\begin{quote}
{\scriptsize
from Art. 456: 
\textit{``la successione si apre al momento della morte nel luogo dell'ultimo domicilio del defunto''} (i.e., ``the succession opens at the moment of death in the place of the last domicile of the deceased person'')

\vspace{1mm}
from Art. 737:
\textit{``i figli e i loro discendenti ed il coniuge che concorrono alla successione devono conferire ai coeredi tutto ci\`{o} che hanno ricevuto dal defunto per donazione''} (i.e., the children and their descendants and the spouse who contribute to the succession must give to the co-heirs everything they have received from the deceased person as a donation)

}

\end{quote}

 In Fig.~\ref{fig:patterns12}(a), we observe how the source is connected to a meaningful, non-contiguous set of words, particularly, ``apre'' (``opens''), ``morte'' (``death''), and ``defunto'' (``deceased person''). In addition, in Fig.~\ref{fig:patterns12}(b), we observe how ``successione'' is related to ``coniuge'' (``spouse''), ``donazione'' (``donation''), which is further enriched in the two-head attention patterns with ``coeredi'' (``co-heirs''), ``conferire'' (``give''), and ``concorrono'' (``contribute''), shown in Fig.~\ref{fig:patterns12}(c); moreover, ``successione'' is still connected to ``defunto''. 
Remarkably, these patterns highlight the model's ability not only to mine semantically meaningful patterns that are more complex than next-word or delimiter-focused patterns, but also to build   patterns that consistently hold across various sentences  sharing   words. 
Note that, as shown in our examples, these sentences can belong to different contexts (i.e., different articles), and can significantly vary in length. 
The latter point is particularly evident, for instance, in the following example sentences:

\vspace{1mm}
\begin{quote}
{\scriptsize
from Art. 457: \textit{``l'eredit\`{a} si devolve per legge o per testamento. Non si fa luogo alla successione legittima se non quando manca, in tutto o in parte, quella testamentaria''} (i.e., ``the inheritance is devolved by law or by will. There is no place for legitimate succession except when the testamentary succession is missing, in whole or in part'')

\vspace{1mm}
from Art. 683: 
\textit{``la revocazione fatta con un testamento posteriore conserva la sua efficacia anche quando questo rimane senza effetto perch\'{e} l'erede istituito o il legatario \`{e} premorto al testatore, o \`{e} incapace o indegno, ovvero ha rinunziato all'eredit\`{a} o al legato''} (i.e., `` the revocation made with a later will retains its effectiveness even when this remains without effect because the established heir or legatee is premortal to the testator, or is incapable or unworthy, or has renounced the inheritance or the legacy'')

}
\end{quote}

\noindent 
Focusing now on ``eredit\`{a}'' (``inheritance''),  in   Art. 457 we found attention patterns with ``testamento'' (``testament''), ``legittima'' (``legitimate''), ``testamentaria'' (``testamentary succession''), whereas in the long sentence from Art. 683,  we again found a link to ``testamento'' (``testament'') as well as with the related concept ``testatore'' (``testator''), in addition to  connections with ``premorto'' (``premortal''), 
``indegno'' (``unworthy''), and ``rinunziato'' (``renounced'').

We point out that our investigation of \Lamberta models' attention patterns    was   found to be persistent over different input sequences, besides the above discussed examples. This has    unveiled the ability of \Lamberta models to form different  types of patterns, which     include complex bag-of-words and next-word patterns.

\section{Visualization of ICC \Lamberta Embeddings}
\label{sec:visual-embeddings}

\begin{figure}[t]
\centering
\begin{tabular}{ccc} 
\hspace{-5mm}
\includegraphics[scale=0.27]{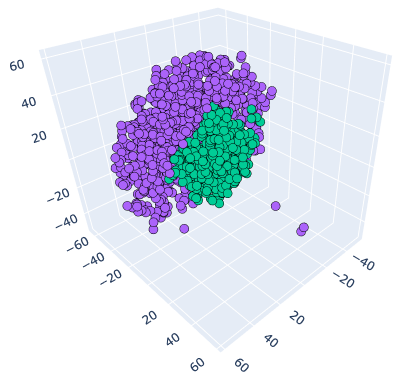} & 
\hspace{-25mm}
\includegraphics[scale=0.27]{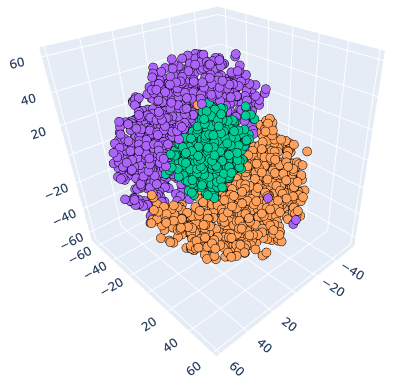} &
\hspace{-25mm}
\includegraphics[scale=0.27]{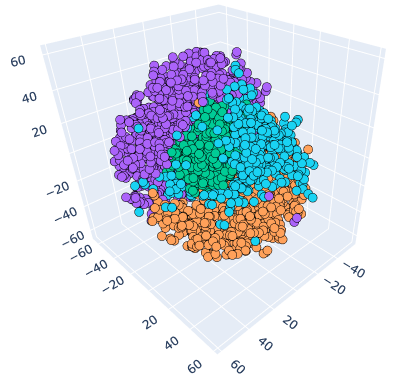} \\ 
\hspace{-5mm} (a) Books 3 and 4 & \hspace{-25mm} (b) Books 3-5 & \hspace{-25mm} (c) Books 3-6
\\
\includegraphics[scale=0.27]{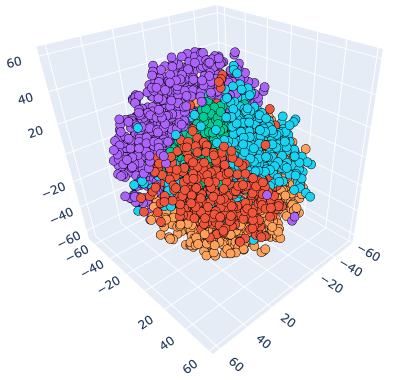} & 
\hspace{5mm}
\includegraphics[scale=0.37]{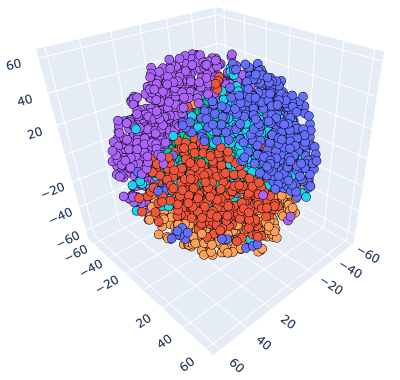} \\
(d) Books 2-6  & \hspace{5mm} (e)  All ICC books
\end{tabular}
\caption{Visualization of the ICC article embeddings produced by \Lamberta local models, transformed onto a 3D t-SNE space. Color codes correspond to ICC books: blue for Book-1, red for Book-2, green for Book-3, purple for Book-4, orange for Book-5, cyan for Book-6}
\label{fig:loc-embeddings}
\end{figure}

\begin{figure}[t]
\centering
\begin{tabular}{ccc} 
\hspace{-5mm}
\includegraphics[scale=0.27]{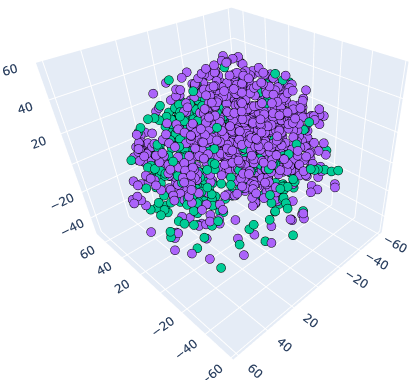} & 
\hspace{-28mm}
\includegraphics[scale=0.27]{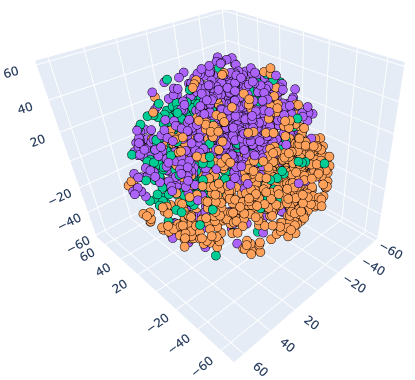} &
\hspace{-28mm}
\includegraphics[scale=0.27]{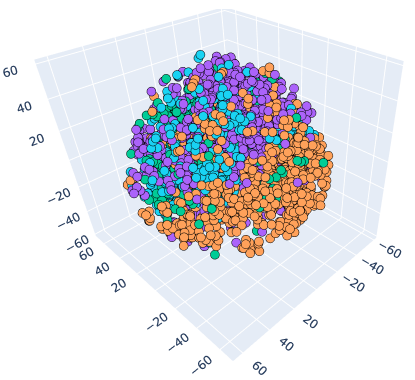} \\ 
\hspace{-5mm} (a) Books 3 and 4 & \hspace{-28mm} (b) Books 3-5 & \hspace{-28mm} (c) Books 3-6 
\\
\includegraphics[scale=0.27]{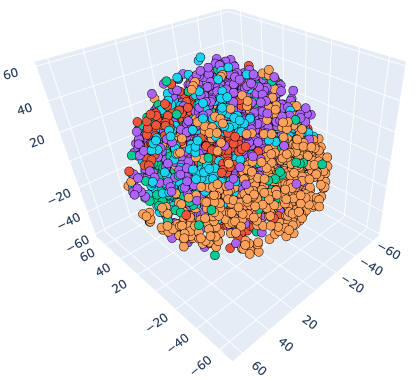} & 
\hspace{5mm}
\includegraphics[scale=0.37]{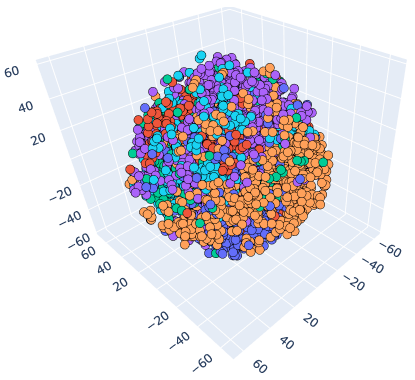} \\
(d) Books 2-6  & \hspace{5mm} (e)  All ICC books
\end{tabular}
\caption{Visualization of the ICC article embeddings produced by \Lamberta global model, transformed onto a 3D t-SNE space. Color coding is the same as in Fig.~\ref{fig:loc-embeddings}}
\label{fig:glob-embeddings}
\end{figure}

Our qualitative evaluation of \Lamberta models is also concerned with   the core outcome of the models, that is, the learned representation  embeddings for the inputted text.  
 
As previously mentioned in Section \ref{sec:learning-config}, for any given text in input, a real-valued vector C of length 768 is produced   downstream of the 12 layers of a BERT model,  in correspondence with the input embedding E$_{[\mathrm{CLS}]}$ relating to the special token CLS. This output embedding C can be seen as an encoded representation of the input text supplied to the model. 

In the following, we present a visual exploratory analysis of \Lamberta embeddings based on a powerful, widely-recognized method for the visualization of high-dimensional data, namely 
 \textit{t-Distributed Stochastic Neighbor Embedding} (t-SNE)~\cite{tsne}. t-SNE is a non-linear technique for dimensionality reduction that is highly effective at providing an intuition of how the data is arranged in a high-dimensional space. 
  The t-SNE algorithm calculates a similarity measure between pairs of instances in the high-dimensional space and in the low-dimensional space, usually 2D or 3D. Then, it converts the pairwise  similarities  to joint probabilities and tries to minimize the Kullback-Leibler divergence between the joint probabilities of the low-dimensional embedding and the high-dimensional data.
 
 It is well-known that  t-SNE outputs provide better and more interpretable results than Principal Component Analysis (PCA) and other linear dimensionality reduction models, which are not effective at interpreting complex polynomial relationships between features.  In particular, by seeking to maximize variance and preserving large pairwise distances, the classic PCA focuses  on placing dissimilar data points far apart in a lower dimension representation; however, in order to represent high-dimensional data on low-dimensional, non-linear manifold, it is important that similar data points must be represented close together. This is ensured by t-SNE, which   preserves   small pairwise distances or local similarities (i.e., nearest-neighbors), so that similar points on the manifold are mapped to similar points in the low-dimensional representation. 
  
 Figure~\ref{fig:loc-embeddings} displays  3D t-SNE representations of the article embeddings generated by \Lamberta local models, for each book of the ICC. (Each point  represents the t-SNE transformation of the embedding of an article onto a 3D space, while colors are used to distinguish the various books.) In the figure, we show progressive combinations of the results from    different books,\footnote{The order of combination of the books is just due to the sake of presentation.}  while the last chart corresponds to the whole ICC. This choice of presentation is motivated to ease the readability of each   book's article embeddings. 
 It can be noted that the \Lamberta local models are able to generate embeddings so that t-SNE can effectively put nearest-neighbor  cases together,  
 with a certain tendency of distributing the points from different books in differest subspaces.  
 
 Let us now compare the above results with those from Figure~\ref{fig:glob-embeddings}, which shows the 3D t-SNE representations of the article embeddings generated by \Lamberta global model.\footnote{In both local and global settings,   t-SNE was carried out until convergence, with the default value of \textit{perplexity}, which is a key parameter to control the number of effective nearest neighbors in the manifold learning. Notably, we tried different values for the perplexity, both within and outside the recommended range (i.e., 5-50), but the   difference in the visual analysis results between the global and local cases still remained the same.   }
 From the comparison, we observe a less compact and localized representation in the global model embeddings w.r.t. the local model ones. This is interesting yet expected, since global models are designed to  go beyond the boundaries of a book, whereas local models focus on the interrelations between articles on the same book. 

It should be noted that while the above remark would hint at preferring the use of local models against global models, at least in terms of interpretability based on visual exploratory analysis, we shall deep our understanding on the effectiveness of the two learning approaches in the next section, which is dedicated to the presentation and discussion of our extensive, quantitative experimental evaluation of \Lamberta models.

\section{Experimental Evaluation}
\label{sec:evaluation}

In this section we present the methodology and results of an experimental evaluation that we have thoroughly carried out on \Lamberta models. 
In the following, we first state our evaluation goals in Section~\ref{sec:evaluation:goals}, then we define the types of test queries and select the  datasets in Section~\ref{sec:evaluation:queries}, finally we  describe our   methodology and assessment criteria  in Section~\ref{sec:evaluation:criteria}.

\subsection{Evaluation goals}
\label{sec:evaluation:goals}

Our main evaluation goals can be summarized as follows: 
\begin{itemize}
\item
To validate and measure the effectiveness of \Lamberta models for  law article retrieval tasks: how do local and global models perform on different evaluation contexts, i.e., against queries of different type, different length, and different lexicon?  (Sect.~\ref{sec:LocalVsGlobal})
\item  
To evaluate  \Lamberta models in  single-label as well as multi-label classification tasks: how do they perform w.r.t. different assumptions on the article relevance to a query, particularly depending on whether a query is  originally associated  with or derived from a particular article, or   by-definition  associated with a group of articles? (Sect.~\ref{sec:LocalVsGlobal}) 
\item 
To understand  how a \Lamberta model's behavior is affected by varying and changing its constituents in terms of training-instance  labeling schemes and learning parameters (Sect.~\ref{sec:ablation}).
 
\item  To demonstrate  the superiority of our classification-based approach to law article retrieval  by comparing \Lamberta to other deep-learning-based text classifiers   (Sect.~\ref{sec:comparison:text}) and to 
 a few-shot learner conceived for an attribute-aware   prediction task that we have newly designed based on the  ICC heading metadata  (Sect.~\ref{sec:comparison:fewshot}).
\end{itemize}

\subsection{Query sets}
\label{sec:evaluation:queries}

A query set is here meant as a collection of natural language texts that discuss legal subjects relating to the ICC articles. For evaluation purposes,  each query as a test instance is associated with one or more article-class labels.  

It should be emphasized that, on the one hand, we cannot devise a training-test split or cross-validation of the target ICC corpus (since we want that our \Lamberta models  embed the   knowledge from all articles therein), and on the other hand, there is a lack of query evaluation benchmarks for Italian civil law documents. 
Therefore, we devise  different  types of query sets, which are aimed at representing  testbeds  at varying difficulty level   for evaluating our \Lamberta models: 

\begin{itemize}
\item
QType-1 -- \textit{book-sentence-queries} 
refer  to a set of queries that correspond to  randomly selected sentences from the articles of a  book.  Each query is derived from a single article, and multiple queries are from the same article.  
\item 
QType-2 -- \textit{paraphrased-sentence-queries} share the same composition of QType-1 queries but differ  from them as the sentences of a book's articles are   paraphrased. To this purpose, we adopt a simple approach based on backtranslation from English (i.e., an original sentence in Italian is first translated to English, then the obtained English sentence is translated to Italian).\footnote{We used the Google Translate service, which is a widely-used yet effective machine translator.}  
\item 
QType-3 -- \textit{comment-queries} are defined to leverage    the  publicly available \textit{comments} on the ICC articles provided by legal experts through the platform ``Law for Everyone''.\footnote{\textit{Law for Everyone} online news portal, available at https://www.laleggepertutti.it.}   
Such comments are delivered to provide annotations about the  interpretation of the meanings and law implications associated to an article, or to  particular terms occurring in an article. 
Each query corresponds to a comment available about one article, which is a paragraph  comprised of about 5 sentences on average. 
\item 
QType-4 -- \textit{comment-sentence-queries} refer  to the same source as QType-3, but the comments are split into sentences, so that each query contains a single sentence of a comment. Therefore, each query will be associated to a single article, and multiple queries will refer to the same article.   
\item  
QType-5 -- \textit{case-queries} refer to a collection of case law decisions from the civil section of the Italian Court of Cassation, which is the highest court in the Italian judicial system. These case law decisions are selected from publicly available corpora of the most significant jurisprudential sentences associated with the ICC articles, spanning over the period 1977-2015.  
\item   QType-6   -- \textit{ICC-heading-queries} are defined by extracting the headings of  chapters, subchapters, and sections of each ICC book. Such headings are   very short, ranging from one to few keywords  used to describe the topic of a particular division of a book. 

\end{itemize}

\begin{table}[t!]
\centering
\caption{Main statistics on the test query sets.}
\label{tab:query-stats}
\scalebox{0.85}{
\begin{tabular}{|l||l|c|c||l|c|c|c|}
\hline 
ICC  &   query & \#queries    & \#words & query & \# queries    & \#words  & \#sentences \\  
portion & type &  & & type & & &\\ \hline \hline
 
\textsl{Book-1} & QType-1 & 790 & 13\,436 & QType-3 & 331  & 38\,383  & 1\,262  
\\  \cline{1-1}\cline{3-4}\cline{6-8} 

\textsl{Book-2}  &   QType-2  & 690 & 10\,998 &   QType-4  & 322 & 30\,226  & 1\,000    
\\   \cline{1-1}\cline{3-4}\cline{6-8} 

\textsl{Book-3} &  & 728 &  11\,942 &  & 286  &  23\,815 & 781  
\\   \cline{1-1}\cline{3-4}\cline{6-8} 

\textsl{Book-4} &  & 1\,774 &  27\,352 &  & 764  &  85\,360 &  2\,640   
\\   \cline{1-1}\cline{3-4}\cline{6-8} 

\textsl{Book-5} &  & 1\,426 & 22\,754  &  & 570  &  87\,094 & 2\,417     
\\   \cline{1-1}\cline{3-4}\cline{6-8} 

\textsl{Book-6} &  & 662 & 11\,993  &  & 294  &  31\,898 &  1\,040    
\\   \hline
\end{tabular}
} 

 \ \\ \vspace{2mm}
\scalebox{0.85}{
\begin{tabular}{|l||l|c|c|c|}
\hline 

ICC  &   query & \#queries    & \#words  & year \\

portion & type &   &   & range \\ \hline \hline
 
\textsl{Book-1} & QType-5 & 333  &   39\,464  & 1978-2014
\\ \cline{1-1}\cline{3-5}

\textsl{Book-2} &   &  347  &    36\,856  & 1979-2014
\\ \cline{1-1}\cline{3-5}

\textsl{Book-3} &  &  371  &  39\,643   & 1979-2014
\\ \cline{1-1}\cline{3-5}

\textsl{Book-4} &   &  975  &    111\,234 & 1978-2015
\\ \cline{1-1}\cline{3-5}

\textsl{Book-5} &  &  1\,037  &  114\,287   & 1977-2015
\\ \cline{1-1}\cline{3-5}

\textsl{Book-6} &  & 720    &   81\,186  & 1978-2015
\\ \hline
\end{tabular}
}

 \ \\ \vspace{2mm}
\scalebox{0.85}{
\begin{tabular}{|l||l|c|c|c|c|}
\hline 
ICC  &   query & \#chapter    & \#subchapter & \#section & \#total  \\
portion & type & queries & queries  & queries  & words \\ \hline \hline
 
\textsl{Book-1} &   QType-6 & 14  & 25  & 22  & 317 
\\  \cline{1-1}\cline{3-6} 

\textsl{Book-2} &  & 5  & 30  & 14  &  160 
\\   \cline{1-1}\cline{3-6} 

\textsl{Book-3}   & &    9 & 21  & 29  &  230 
\\   \cline{1-1}\cline{3-6} 

\textsl{Book-4}  &    & 9  & 51  & 57  & 366   
\\   \cline{1-1}\cline{3-6} 

\textsl{Book-5}  &    & 11  & 11  &  51 &      397
\\   \cline{1-1}\cline{3-6} 

\textsl{Book-6}  &    & 5  &  18 &  38 &    251  
\\   \hline
\end{tabular}
} 
\end{table}

\noindent 
{\bf Characteristics and differences of the query-sets.\ } 
It should be noted that while QType-1 and QType-6  query sets have the same  lexicon as   the corresponding book articles, this does not necessarily hold for   QType-2  due to the paraphrasing process, and the difference becomes more evident with QType-3, QType-4,   and QType-5  query sets. 
Indeed, the latter not only originate from a  corpus different from the ICC, but they have the typical verbosity of annotations or comments about  law articles  as well as of case law decisions; moreover,   they often provide cross-book references, therefore,  QType-3,  QType-4,  and QType-5 query-set contents are not necessarily bounded by a book's context.  

Moreover, QType-3,    QType-5, and QType-6 differ from the other types in terms of length of each query, which corresponds indeed to multiple sentences or a paragraph, in the case of QType-3  and QType-5,  and to a single or few  keywords, in the case of QType-6. Note that, although derived from the ICC books, \textit{the contents of the QType-6 queries were   totally discarded when training our \Lamberta models}; moreover, unlike the other types of queries, each QType-6 query is by definition associated with a group of articles (i.e., according to the book divisions) rather than a single article, therefore we shall use QType-6 queries for the multi-label evaluation task only. 
 Also,  it should be emphasized that the QType-5 queries represent a different       difficult testbed as they contain  real-life, heterogeneous fact descriptions of the case and judicial precedents.

Table~\ref{tab:query-stats} summarizes main statistics on the query sets. In addition, note that the percentage of QType-1 that were paraphrased (i.e., to produce QType-2 queries) resulted in above 85\% for each of the books. 
Also, the number of sentences in a book's query set (last column of the upper subtable) corresponds to the number of queries of QType-4 for that book.

It is also worth emphasizing that all query sets were validated by legal experts.  This is important to ensure not only generic  linguistic requirements but also meaningfulness of the query contents from a legal viewpoint.

\subsection{Evaluation methodology and Assessment criteria}
\label{sec:evaluation:criteria}

Let us denote with $\mathcal{C}$ either a  portion (i.e., a single book) of the ICC, or the whole ICC. 
 We specify  an evaluation context in terms of  a test set (i.e., query set) $Q$ pertinent to  $\mathcal{C}$ and a \Lamberta model  $\mathcal{M}$ learnt from $\mathcal{C}$. 
Note that the pertinence of $Q$ w.r.t. $\mathcal{C}$ is differently determined depending on the type of query set, as previously discussed in Section~\ref{sec:evaluation:queries}.

We consider two multi-class evaluation contexts:   \textit{single-label}  and \textit{multi-label}. In the single-label context, each query is pertinent to only one article, therefore there is only one relevant result for each query.  In the multi-label context, each query can be pertinent to more than one article, therefore there can be multiple relevant results for each query.

\subsubsection{Single-label evaluation context}   
We consider a confusion matrix in which the article  ids correspond to the classes. Upon this matrix,  we measure the following standard statistics for each article $A_i \in \mathcal{C}$:  
the precision for $A_i$ ($P_i$), i.e., the number of times (queries) $A_i$ was correctly predicted out of all predictions of $A_i$, 
the recall for $A_i$ ($R_i$), i.e., the number of times (queries) $A_i$ was correctly predicted out of all queries actually pertinent to $A_i$, 
the F-measure for $A_i$ ($F_i$), i.e., $F_i = 2P_iR_i/(P_i+R_i)$. 
Then, we averaged over all articles to obtain the per-article average \textit{precision} ($P$), \textit{recall} ($R$), and two types of F-measure: \textit{micro-averaged F-measure} ($F^\mu$) as the actual average over all $F_i$s, and \textit{macro-averaged F-measure} ($F^M$) as the harmonic mean of $P$ and $R$, i.e., $F^M = 2PR/(P+R)$.

In addition, we consider two other types of criteria that respectively account for the top-$k$ predictions and the  position (rank) of the correct article in predictions. 
The first aspect is captured in terms of  the fraction of correct article labels that are found in the top-$k$ predictions (i.e., top-$k$-probability results in response to  each query), and averaging over all queries, which is the   \textit{recall@$k$} ($R@k$). The second aspect is measured as   the \textit{mean reciprocal rank} ($MRR$) considering for each query the rank of the correct prediction over the classification probability distribution, and averaging over all queries. 

Moreover, we wanted to understand the \textit{uncertainty of prediction} of \Lamberta models when evaluated over each   query: to this purpose, we measured the \textit{entropy} of the  classification probability distributions obtained for each query evaluation, and averaged over all  queries; in particular, we distinguished between the entropy of each entire distribution ($E$) from the entropy of the distribution corresponding to the top-$k$-probability results ($E@k$).

\subsubsection{Multi-label evaluation context}
 The basic requirement for the multi-label evaluation context is that, for each test query, 
a set of articles is regarded as relevant to the query.  
 In this respect, we consider two different perspectives, depending on whether a query is (i)  originally associated  with or derived from a particular article, or (ii) by-definition  associated with a group of articles. We will refer to the former scenario  as \textit{article-driven} multi-label evaluation, and to the second scenario as  \textit{topic-driven} multi-label evaluation.

\vspace{1mm}
\noindent 
{\bf Article-driven  multi-label evaluation.\ }
For this stage of evaluation, we require that, for each test query associated with article $A_i$, a set of articles \textit{related} to $A_i$ is to  be selected   as the set of articles relevant to the query, together with $A_i$. 
We adopt two approaches to  determine the \textit{article relatedness}, which  rely on a supervised and an unsupervised grouping of the articles, respectively. 
The supervised approach refers to  the logical organization of the articles of each book originally available in the ICC (cf. Section~\ref{sec:data}), whereas the unsupervised approach leverages a content-similarity-based grouping of the articles produced by a document clustering method.

\textit{ICC-classification-based.\ } 
We exploit the ICC-provided classification meta-data of each book to label each article with the terminal division it belongs to. Since a division usually contains a few articles, the same label will be associated to multiple articles. It should be noted that the book hierarchies are quite different to each other, both in terms of maximum branching (e.g., from 5 chapters in Book-2 and Book-6, to 14 chapters in Book-1) and total number of divisions (e.g., from 51 in Book-2 to 135 in Book-4); moreover,  sections,  subchapters   or even chapters can be leaf nodes in the corresponding hierarchy, i.e., they   contain articles without any further division. 
 This leads to the following distribution of number of labels over the various books:  
50 in Book-1, 40 in Book-2, 47 in Book-3, 112 in Book-4, 94 in Book-5, 56 in Book-6.

\textit{Clustering-based.\ } 
As concerns the unsupervised, similarity-based approach, we compute a clustering of the set of articles of a given book, so that each cluster will correspond to a  group of articles that are similar by content. 
Each of the produced clusters will be regarded as the set of relevant articles for each query whose original label article belongs to that set. Therefore, if a cluster contains $n$ articles, then each of the queries originally labeled with any of such $n$ articles, will share the same set of $n$ relevant articles.

To perform the clustering of the article set of a given book, we resort to a widely-used, well-known document clustering method, which consists in applying a centroid-based partitional clustering algorithm~\cite{JainD88} over a document-term matrix modeling a vectorial bag-of-words representation of the documents over the term feature space, using TF-IDF (term-frequency inverse-document-frequency) as term relevance weighting function~\cite{Jones04} and cosine similarity for document comparison.  
This method is also known as \textit{spherical k-means}~\cite{DhillonM01}. 
We used a particularly effective and optimized version of this method, called bisecting k-means~\cite{ZhaoK04}, which is a standard de-facto of document clustering tools based on the classic bag-of-words model.\footnote{http://glaros.dtc.umn.edu/gkhome/cluto/cluto/overview.}

 It should be emphasized that our goal here is  to induce an organization of the articles that is   based on  content affinity while discarding any information on the ICC article labeling. 
In this regard,   
we have deliberately exploited a representation model that, despite its known limitations in being unable to capture latent semantic aspects underlying correlations between words, it is still an effective baseline, yet unbiased with respect to the deep language model  ability  of \Lamberta.

\textit{Computation of relevant sets.\ }
Let us denote with $\mathcal{P}$ a partitioning of $\mathcal{C}$ that corresponds to either the ICC classification of the articles in $\mathcal{C}$ (i.e., supervised organization) or a clustering of $\mathcal{C}$ (i.e., unsupervised organization). For either approach, given a test query $q_i \in Q$ with article label $A$, we detect the relevant article set for $q$ as  the partition $P \in \mathcal{P}$ that contains $A$, then we match this set to the set of    
 top-$|P|$ predictions  of $\mathcal{M}$ to compute precision, recall, and F-measure for $q_i$. Finally, we averaged over all queries to obtain overall precision, recall, micro-averaged F-measure and macro-averaged F-measure.

\vspace{1mm}
\noindent 
{\bf Topic-driven  multi-label evaluation.\ }
Unlike the previously discussed evaluation stage, here we consider queries that are   expressed by one or few keywords  describing the topic associated with a set of articles. 

To this purpose, we again exploit the meta-data of article organization available in the ICC books, so that  the   articles belonging to the same division of a book (i.e., chapter, subchapter, section)   will be regarded as   the set of articles relevant to the query corresponding to the description of that division. 
Depending on the choice of the type of a book's division, i.e., the level of the ICC-classification hierarchy of that book, a different query set will be produced for the book. 

Given a test query $q_i \in Q$ with article labels $A_{i_1}, \ldots, A_{i_n}$,  we match this set to the set of top-$i_n$ predictions of $\mathcal{M}$ to compute precision, recall, and F-measure for $q_i$. Finally, we averaged over all queries to obtain overall precision, recall,  micro-averaged F-measure and macro-averaged F-measure.   
 Moreover, we measured the fraction of  top-$k$ predictions that are relevant to a query, and averaging over all queries, i.e.,  \textit{precision@$k$} ($P@k$).

\section{Results}
\label{sec:results}

We organize the presentation of results into three parts. 
Section~\ref{sec:LocalVsGlobal} describes the comparison between global and local models under both the single-label and multi-label evaluation contexts; note that we will use notations $G$ and $L_i$ to refer to the global model and the local model corresponding to the $i$-th book, respectively, for any given test query-set. 
Section~\ref{sec:ablation} is devoted to an ablation study of our models, focusing on the analysis of the various unsupervised data labeling methods and the effect of the training size on the models' performance. 
Finally, Section~\ref{sec:comparison} presents results obtained by competing deep-learning methods.

\subsection{Global vs. Local models}
\label{sec:LocalVsGlobal}

\subsubsection{Single-label evaluation} 
\begin{figure}[t!]
\centering
\begin{tabular}{ccc}
\hspace{-5mm}
\includegraphics[scale=0.27]{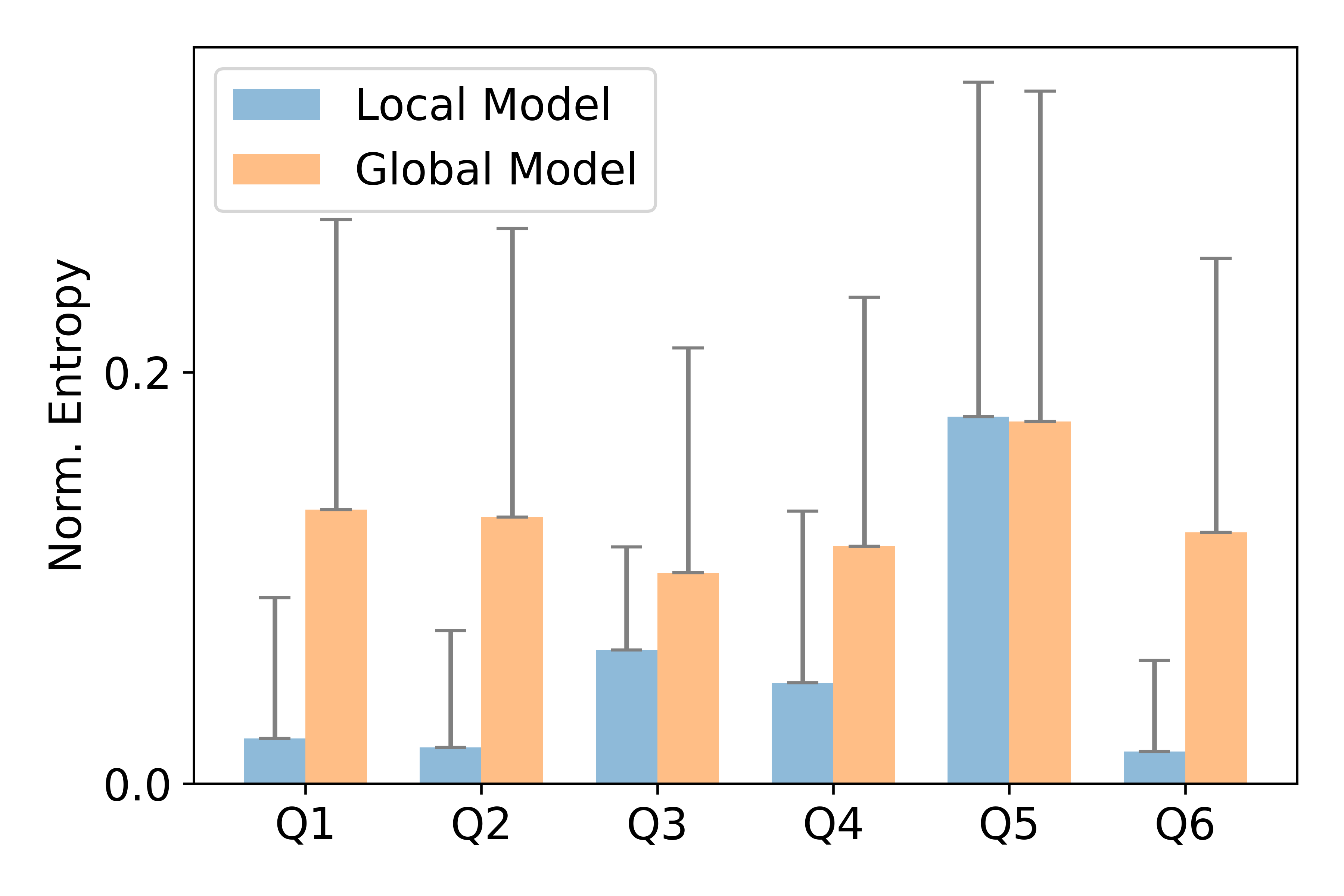} & 
\hspace{-3mm}
\includegraphics[scale=0.27]{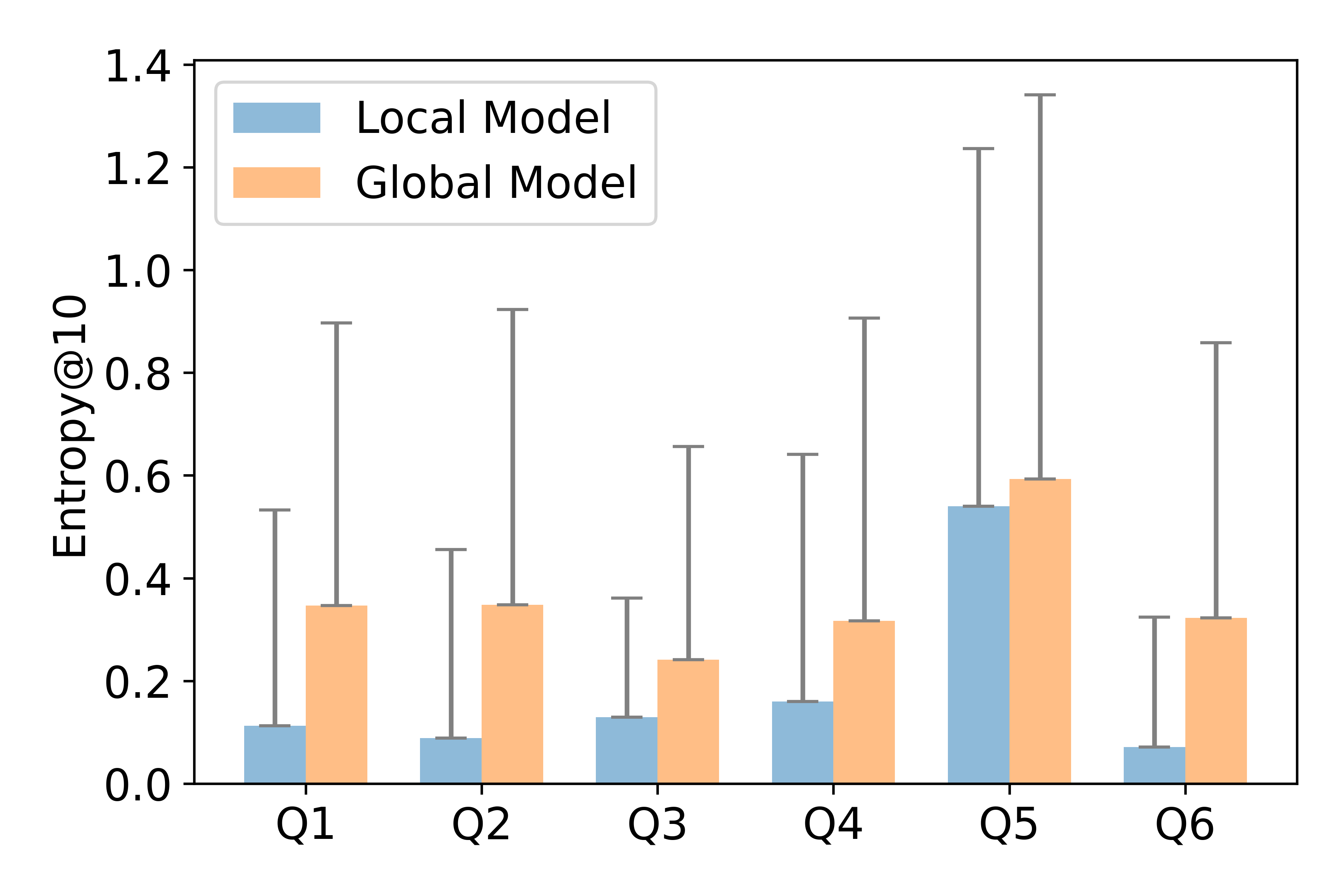} &
\hspace{-3mm}
\includegraphics[scale=0.27]{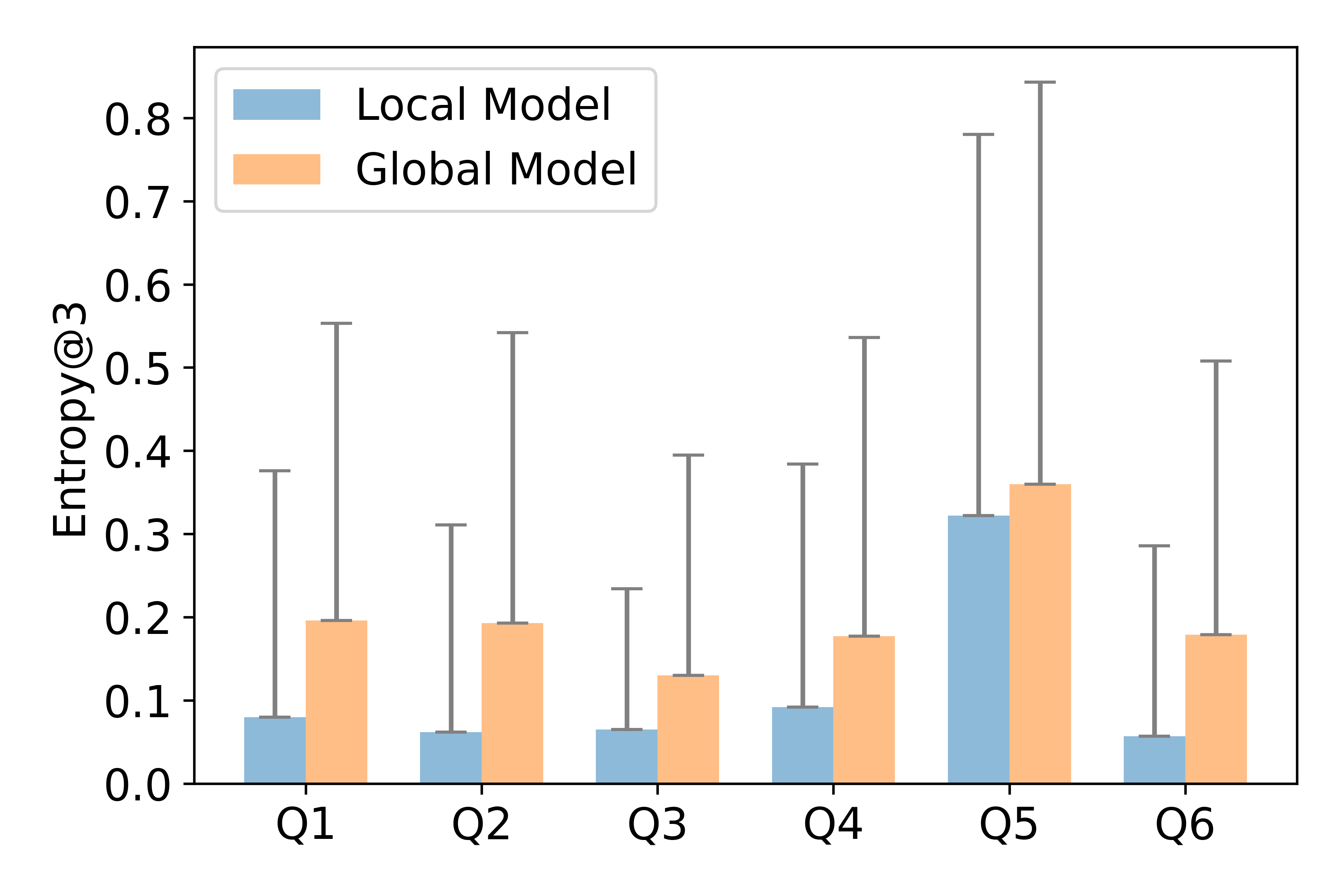}    
\\
(a) $E$ & (b) $E@10$ & (c) $E@3$ \\ 
\hspace{-5mm}
\includegraphics[scale=0.27]{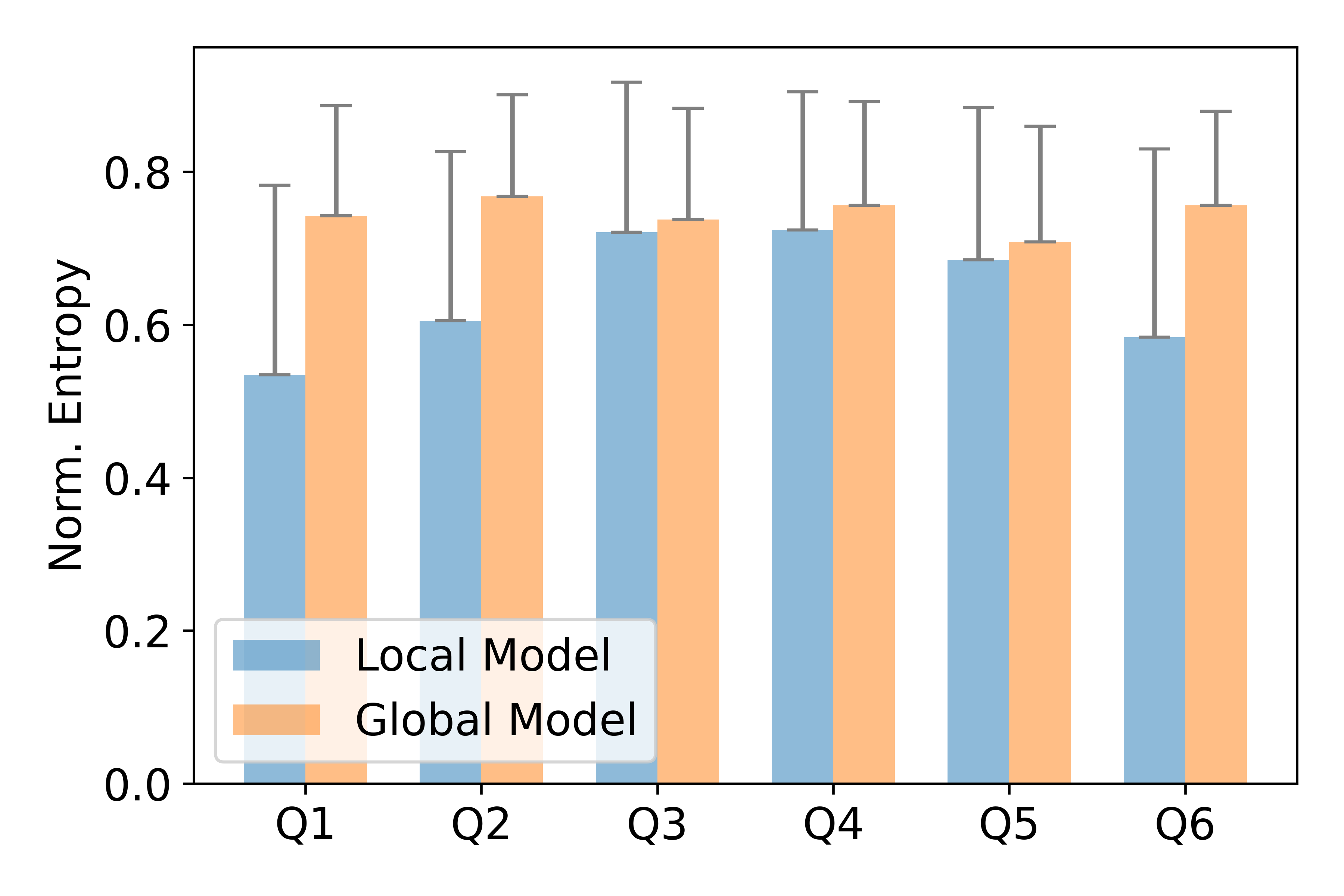} & 
\hspace{-3mm}
\includegraphics[scale=0.27]{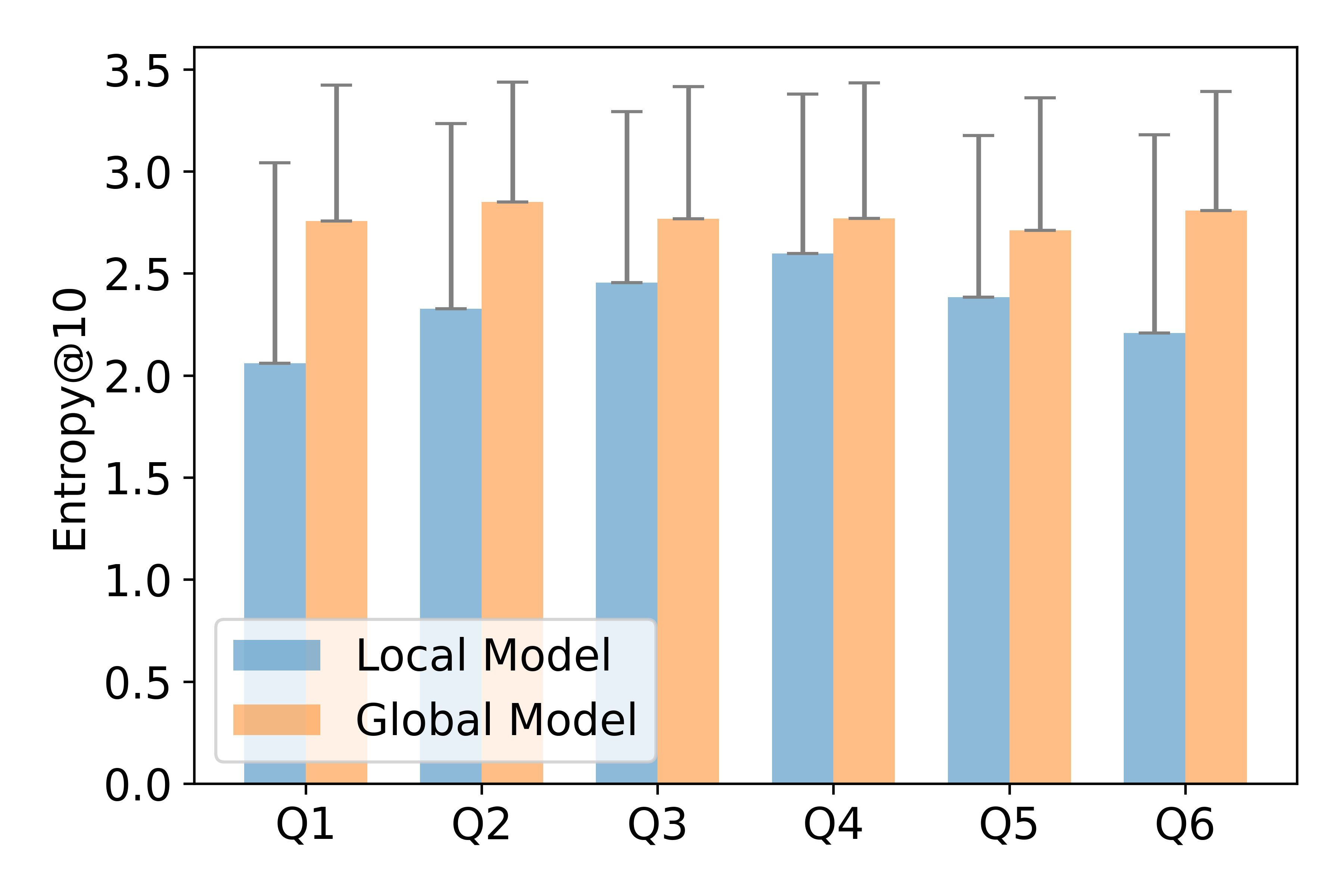} &
\hspace{-3mm}
\includegraphics[scale=0.27]{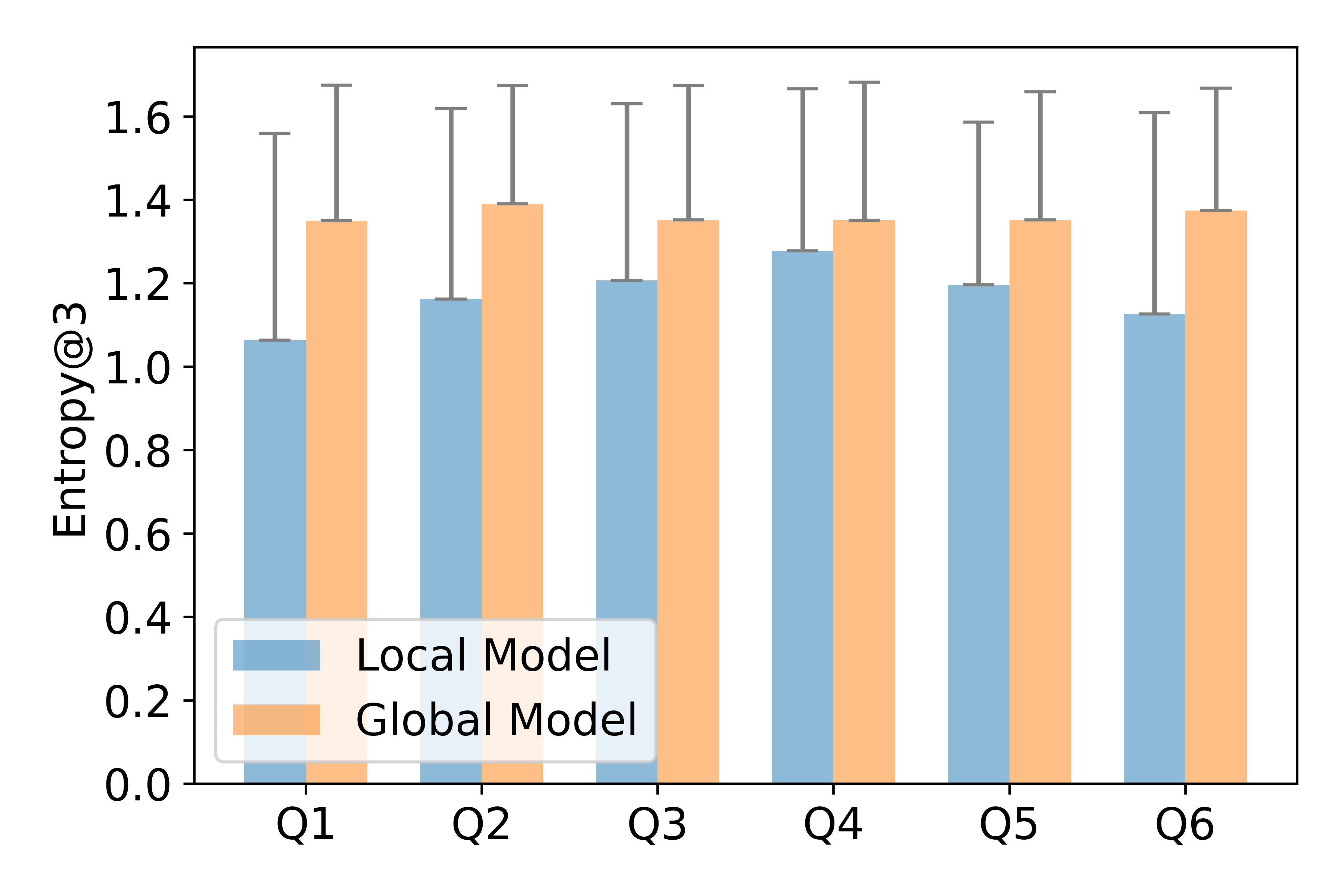}    
\\
(d) $E$ & (e) $E@10$ & (f) $E@3$ \\ 
\end{tabular}
\caption{Global vs. local models, for all sets of book-sentence-queries (a-c) and comment-sentence-queries (d-f): Mean and standard deviation of  the normalized Entropy  and of the  Entropy@$k$   values  of the classification probability distributions}
\label{fig:modelliv5_UniRRemphT4_QT2_entropy}
\end{figure}

Table~\ref{tab:modelliv5_UniRRemphT4_QT2} and  Fig.~\ref{fig:modelliv5_UniRRemphT4_QT2_entropy} compare  global and local models, for all books of the ICC, according to the   criteria selected for the single-label evaluation context.  
For the sake of brevity yet   representativeness, here we show results corresponding to the  
{\unirrempht} labeling scheme --- as we shall discuss in Section~\ref{sec:ablation}, this choice of training-instance labeling scheme is justified as being in general the best-performing scheme for the query sets; 
nonetheless, analogous findings  were drawn by using other types of queries and labeling schemes.

\setlength{\tabcolsep}{3.5pt}
\begin{table}[t!]
\centering
\caption{Global vs. local models, 
for all sets of book-sentence-queries (QType-1, upper subtable),    paraphrased-sentence-queries (QType-2, second upper subtable), comment-queries (QType-3, third upper subtable),    comment-sentence-queries (QType-4, second bottom subtable),   and case-queries (QType-5, bottom subtable): Recall, Precision, micro- and macro-averaged F-measures, Recall$@k$, and MRR. (Bold values correspond to the best model for each query-set and evaluation criterion)}
\label{tab:modelliv5_UniRRemphT4_QT2}
\scalebox{0.8}{
\begin{tabular}{|c|cc|cc|cc|cc||cc|cc|cc|}
\hline
 & 
   \multicolumn{2}{|c|}{$R$} & 
   \multicolumn{2}{|c|}{$P$} & 
   \multicolumn{2}{|c|}{$F^{\mu}$} &
   \multicolumn{2}{|c||}{$F^{M}$} & 
   \multicolumn{2}{|c|}{$R@3$} & 
   \multicolumn{2}{|c|}{$R@10$} & 
   \multicolumn{2}{|c|}{$MRR$} 
   \\
\hline 
$i$ & $L_i$ & $G$ & $L_i$ & $G$ & $L_i$ & $G$ & $L_i$ & $G$ & $L_i$ & $G$ & $L_i$ & $G$ & $L_i$ & $G$\\ \hline  \hline
$Q_1$ &  
\textbf{0.962}	&	0.949	&
0.975	&	\textbf{0.979}	&
\textbf{0.961}	&	0.955	&
\textbf{0.969}	&	0.964	&
0.989	&	\textbf{0.992}	&
\textbf{0.999}	&	0.997	&
\textbf{0.976}	&	0.970	 \\  
$Q_2$  &   
\textbf{0.972}	&	0.954	&
0.981	&	\textbf{0.985}	&
\textbf{0.971}	&	0.961	&
\textbf{0.977}	&	0.969	&
\textbf{0.994}	&	0.981	&
\textbf{0.999}	&	0.986	&
\textbf{0.983}	&	0.968	\\
$Q_3$ &   
\textbf{0.990}	&	0.978	&
\textbf{0.994}	&	0.994	&
\textbf{0.990}	&	0.981	&
\textbf{0.992}	&	0.986	&
\textbf{1.000}	&	0.995	&
\textbf{1.000}	&	0.997	&
\textbf{0.995}	&	0.987	\\ 
$Q_4$ &   
\textbf{0.961}	&	0.948	&
0.983	&	\textbf{0.985}	&
\textbf{0.964}	&	0.958	&
\textbf{0.972}	&	0.966	&
\textbf{0.984}	&	0.977	&
\textbf{0.990}	&	0.986	&
\textbf{0.975}	&	0.966	\\
$Q_5$ &   
\textbf{0.910}	&	0.905	&
0.944	&	\textbf{0.947}	&
\textbf{0.909}	&	0.908	&
\textbf{0.927}	&	0.925	&
\textbf{0.984}	&	0.974	&
\textbf{0.998}	&	0.988	&
\textbf{0.947}	&	0.940	\\
$Q_6$ & 
\textbf{0.979}	&	0.961	&
\textbf{0.986}	&	0.986	&
\textbf{0.978}	&	0.966	&
\textbf{0.982}	&	0.973	&
\textbf{1.000}	&	0.985	&
\textbf{1.000}	&	0.992	&
\textbf{0.989}	&	0.974	\\
\hline
\hline

$Q_1$ & \textbf{0.841}			&	0.730 			&
\textbf{0.866}			&	0.823 			&
\textbf{0.828}			&	0.745 			&
\textbf{0.853}		&	0.774 			&
\textbf{0.905}			&	0.829 			&
\textbf{0.949}			&	0.881 			&
\textbf{0.881}			&	0.784 			\\

$Q_2$ & \textbf{0.828}			&	0.639 			&
\textbf{0.856}			&	0.766 			&
\textbf{0.814}			&	0.669 			&
\textbf{0.841}		&	0.697 			&
\textbf{0.992}			&	0.742 			&
\textbf{0.941}			&	0.812 			&
\textbf{0.871}			&	0.701 			\\

$Q_3$ & \textbf{0.861}			&	0.728 			&
\textbf{0.886}			&	0.843 			&
\textbf{0.851}			&	0.756 			&
\textbf{0.873}		&	0.781 			&
\textbf{0.922}			&	0.816 			&
\textbf{0.942}			&	0.842 			&
\textbf{0.896}			&	0.778 			\\

$Q_4$ & \textbf{0.736}			&	0.635 			&
\textbf{0.756}			&	0.706 			&
\textbf{0.713}			&	0.639 			&
\textbf{0.746}		&	0.668 			&
\textbf{0.806}			&	0.716 			&
\textbf{0.861}			&	0.783 			&
\textbf{0.779}			&	0.685 			\\

$Q_5$ & \textbf{0.718}			&	0.684 			&
\textbf{0.759}			&	0.742 			&
\textbf{0.710}			&	0.686 			&
\textbf{0.738}		&	0.712 			&
\textbf{0.843}			&	0.808 			&
\textbf{0.908}			&	0.867 			&
\textbf{0.790}			&	0.755 			\\

$Q_6$ & \textbf{0.841}			&	0.704 			&
\textbf{0.874}			&	0.817 			&
\textbf{0.833}			&	0.730 			&
\textbf{0.857}		&	0.756 			&
\textbf{0.914}			&	0.783 			&
\textbf{0.941}			&	0.845 			&
\textbf{0.882}			&	0.756 			\\

\hline
\hline

$Q_1$ & \textbf{0.349}			&	0.25 			&
\textbf{0.248}			&	0.196 			&
\textbf{0.274}			&	0.211 			&
\textbf{0.290}		&	0.220 			&
\textbf{0.494}			&	0.422 			&
\textbf{0.675}			&	0.530 			&
\textbf{0.455}			&	0.355 			\\

$Q_2$ & \textbf{0.313}			&	0.177 			&
\textbf{0.213}			&	0.145 			&
\textbf{0.239}			&	0.155 			&
\textbf{0.253}		&	0.159 			&
\textbf{0.494}			&	0.307 			&
\textbf{0.655}			&	0.441 			&
\textbf{0.445}			&	0.265 			\\

$Q_3$ & \textbf{0.396}			&	0.260 			&
\textbf{0.298}			&	0.216 			&
\textbf{0.325}			&	0.228 			&
\textbf{0.340}		&	0.236 			&
\textbf{0.577}			&	0.426 			&
\textbf{0.704}			&	0.574 			&
\textbf{0.507}			&	0.371 			\\

$Q_4$ & \textbf{0.336}			&	0.247 			&
\textbf{0.239}			&	0.190 			&
\textbf{0.264}			&	0.206 			&
\textbf{0.279}		&	0.215 			&
\textbf{0.487}			&	0.387 			&
\textbf{0.622}			&	0.522 			&
\textbf{0.438}			&	0.343 			\\

$Q_5$ & 0.171 			&	\textbf{0.252} 			&
0.128 			&	\textbf{0.190} 			&
0.137 			&	\textbf{0.205} 			&
0.147 		&	\textbf{0.216} 			&
0.364			&	\textbf{0.433} 			&
0.562 			&	\textbf{0.590} 			&
0.302 			&	\textbf{0.375} 			\\

$Q_6$ & \textbf{0.387}			&	0.235 			&
\textbf{0.291}			&	0.178 			&
\textbf{0.317}			&	0.193 			&
\textbf{0.332}		&	0.203 			&
\textbf{0.588}			&	0.388 			&
\textbf{0.745}			&	0.551 			&
\textbf{0.515}			&	0.340 			\\

\hline
\hline

$Q_1$ & \textbf{0.190}			&	0.136			&
\textbf{0.197}			&	0.183			&
\textbf{0.170}			&	0.135			&
\textbf{0.194}		&	0.156			&
\textbf{0.363}			&	0.265			&
\textbf{0.502}			&	0.376			&
\textbf{0.333}			&	0.239			\\
$Q_2$ &  \textbf{0.216}			&	0.100			&
\textbf{0.191}			&	0.148			&
\textbf{0.173}			&	0.106			&
\textbf{0.203}			&	0.119			&
\textbf{0.343}			&	0.172			&
\textbf{0.473}			&	0.257			&
\textbf{0.315}			&	0.157			\\
$Q_3$ & \textbf{0.241}			&	0.153			&
\textbf{0.223}			&	0.188			&
\textbf{0.204}			&	0.145			&
\textbf{0.231}			&	0.169			&
\textbf{0.433}			&	0.268			&
\textbf{0.569}			&	0.405			&
\textbf{0.395}			&	0.250			\\
$Q_4$ &  	\textbf{0.189}			&	0.144			&
\textbf{0.176}			&	0.173			&
\textbf{0.156}			&	0.137			&
\textbf{0.182}			&	0.157			&
\textbf{0.324}			&	0.253			&
\textbf{0.450}			&	0.345			&
\textbf{0.293}			&	0.228			\\
$Q_5$ &   0.092			&	\textbf{0.119}			&
0.132			&	\textbf{0.148}			&
0.090			&	\textbf{0.113}			&
0.108			&	\textbf{0.132}			&
0.256			&	\textbf{0.278}			&
0.414			&	\textbf{0.426}			&
0.222			&	\textbf{0.248}			\\
$Q_6$ &  \textbf{0.224}			&	0.144			&
\textbf{0.211}			&	0.209			&
\textbf{0.192}			&	0.147			&
\textbf{0.218}			&	0.171			&
\textbf{0.372}			&	0.237			&
\textbf{0.508}			&	0.336			&
\textbf{0.333}			&	0.210			\\

\hline
\hline

$Q_1$ & \textbf{0.228}			&	0.118 			&
\textbf{0.233}			&	0.121 			&
\textbf{0.210}			&	0.101 			&
\textbf{0.230}		&	0.120 			&
\textbf{0.494}			&	0.304 			&
\textbf{0.813}			&	0.488 			&
\textbf{0.430}			&	0.263 			\\

$Q_2$ & \textbf{0.292}			&	0.162 			&
\textbf{0.316}			&	0.232 			&
\textbf{0.274}			&	0.172 			&
\textbf{0.303}		&	0.191		&
\textbf{0.501}			&	0.280 			&
\textbf{0.726}			&	0.444 			&
\textbf{0.465}			&	0.270 			\\

$Q_3$ & \textbf{0.284}			&	0.190 			&
\textbf{0.323}			&	0.217 			&
\textbf{0.273}			&	0.180 			&
\textbf{0.302}		&	0.203 			&
\textbf{0.566}			&	0.330 			&
\textbf{0.856}			&	0.474 			&
\textbf{0.489}			&	0.286 			\\

$Q_4$ & \textbf{0.259}			&	0.159 			&
\textbf{0.299}			&	0.187 			&
\textbf{0.250}			&	0.150 			&
\textbf{0.278}		&	0.172 		&
\textbf{0.528}			&	0.314 			&
\textbf{0.803}			&	0.493 			&
\textbf{0.458}			&	0.279 			\\

$Q_5$ & \textbf{0.354}			&	0.256 			&
\textbf{0.401}			&	0.307			&
\textbf{0.342}			&	0.252 			&
\textbf{0.376}		&	0.279 		&
\textbf{0.641}			&	0.484 			&
\textbf{0.884}			&	0.647 			&
\textbf{0.564}			&	0.426 			\\

$Q_6$ & \textbf{0.392}			&	0.241 			&
\textbf{0.445}			&	0.313 			&
\textbf{0.372}			&	0.247 			&
\textbf{0.417}		&	0.273 			&
\textbf{0.665}			&	0.376 			&
\textbf{0.885}			&	0.520 			&
\textbf{0.590}			&	0.347 			\\
\hline
\end{tabular}
}
\end{table}

Looking at the table, let us first consider the results obtained by the \Lamberta models against QType-1 queries. At a first sight, we observe   the outstanding  scores  achieved by all models; this was obviously expected since QType-1 queries are directly extracted from sentences of the book articles used to train the models. More interesting is the comparison between  the  local models and the global model: for any query-set from   the $i$-th book, the corresponding local  model behaves significantly  better than   the global model in most cases. It should be noticed  that the slightly  better results of the global model in terms of precision are actually determined by its  occasional wrongly  predictions of articles from different books in place of  articles of Book-$i$.   

 The outperformance of local models over the global one is also confirmed in terms of the entropy results, which are always lower, and hence better,  than the global's ones.   Moreover,  while the plots in Fig.~\ref{fig:modelliv5_UniRRemphT4_QT2_entropy} are representative for  QType-1 and QType-4 queries, the above result for the comparison between local and global models in terms of entropy behavior equally holds regardless of the type of query  set.   
  Note that, in  Fig.~\ref{fig:modelliv5_UniRRemphT4_QT2_entropy}(a) and (d), the values of entropy are normalized within the interval [0, 1] to get a fair  comparison between the local models and global models.\footnote{For each book, the entropy value of the local model is divided by the base-2 logarithm of the number of articles of that book, whereas the entropy value of the global model is always divided by the base-2 logarithm of the total number of articles in the ICC.}    
This indicates that the predictions made by a local model   are generally more  certain than those by the global model.

Turning back to the results reported in Table~\ref{tab:modelliv5_UniRRemphT4_QT2}, we observe that   high performance scores are still obtained  for the paraphrased-sentence-based  queries,  which indicates a remarkable robustness of \Lamberta models w.r.t. lexical variants of queries.

 Considering now the comment-based queries, both in the paragraph-size (i.e., QType-3) and sentence-size versions (i.e., QType-4), we observe a much lower performance  of the models. This is nonetheless not surprising since  
 QType-3 and QType-4 queries are lexically and linguistically different from the ICC contents (cf. Section~\ref{sec:evaluation:queries}). Besides that, local models show to be consistently better than the global ones, despite the cross-corpora learning ability of the global model which, in principle, could be beneficial for queries that may contain references to articles of different books. One exception corresponds to Book-5: as confirmed by a qualitative inspection of the legal experts who assisted us, this is  due to an accentuation in the queries for Book-5 of a tendency in more broadly covering subjects of articles from other books.  
 This is also supported by quantitative statistics about the uniqueness of the terms in the vocabulary of a book: in fact, we found out that the  percentage of a book's vocabulary terms that are unique is minimum w.r.t. Book-5; in detail, 39\% of Book-1 and Book-2, 48\% of Book-3, 43\% of Book-4, and 36\% of Book-6.   
 A  further interesting remark can be drawn from the comparison of  the    QType-3 scores with the  QType-4 scores, as these appear to be generally  higher for the former queries, which provide more topical context than sentence-size queries. This would hence  suggest that relatively long queries can  be handled by \Lamberta models effectively.  
  
  Notably, the above finding on the ability of handling long, contextualized queries is also confirmed by inspectioning the results obtained against the queries stating case law decisions (i.e., QType-5 queries). 
  Indeed, despite the particular difficulty of such a testbed,  \Lamberta models show performance scores that are generally comparable to those obtained for the QType-3 comment  queries;  moreover, once again, local models behave better than the corresponding global ones, according to all assessment criteria (and even  without exceptions as opposed to QType-3 comment  queries).

\setlength{\tabcolsep}{4pt}
\begin{table}[t!]
\centering
\caption{Article-driven, Clustering-based multi-label evaluation: global vs. local models,  with  labeling scheme \unirrempht, for all sets of book-sentence-queries (QType-1),    paraphrased-sentence-queries (QType-2), comment-queries (QType-3),   comment-sentence-queries (QType-4),   and case-queries (QType-5). (Bold values correspond to the best model for each query-set and evaluation criterion)}
\label{tab:modelliv5_UniRRemphT4_multilabel-cb}
\scalebox{0.8}{
\begin{tabular}{|l|c|cc|cc|cc|cc|}
\hline 
query-set & & 
   \multicolumn{2}{|c|}{$R$} & 
   \multicolumn{2}{|c|}{$P$} & 
   \multicolumn{2}{|c|}{$F^{\mu}$} &
   \multicolumn{2}{|c|}{$F^{M}$} 
   \\
\cline{2-10}
& $i$ & $L_i$ & $G$ & $L_i$ & $G$ & $L_i$ & $G$ & $L_i$ & $G$ 
\\ \hline  \hline
QType-1 & $Q_1$  & \textbf{0.615}	&	0.592	&
\textbf{0.617}	&	0.591	&
\textbf{0.615}	&	0.591	&
\textbf{0.616}	&	0.591	\\

& $Q_2$  &  \textbf{0.674} &  0.636 &    
 \textbf{0.684}  & 0.640 &   
  \textbf{0.676}   &  0.637 &   
  \textbf{0.679}    & 0.638  \\	

& $Q_3$ &  \textbf{0.623}	&	0.604	&
\textbf{0.626}	&	0.603	&
\textbf{0.624}	&	0.603	&
\textbf{0.625	}&	0.603	\\

& $Q_4$ &  \textbf{0.331}	&	0.313	&
\textbf{0.326}	&	0.312	&
\textbf{0.324}	&	0.312	&
\textbf{0.328}	&	0.312	\\

& $Q_5$ & 0.631	&	\textbf{0.632}	&
0.633	&	\textbf{0.634}	&
0.628	&	\textbf{0.631}	&
0.632	&	\textbf{0.633}	\\

& $Q_6$ &   \textbf{0.711}	&	0.675	&
\textbf{0.720}	&	0.677	&
\textbf{0.710}	&	0.675	&
\textbf{0.715}	&	0.676	\\

\hline  \hline

QType-2 & $Q_1$  & \textbf{0.557}	&	0.473 	&
\textbf{0.559}	&	0.474 	&
\textbf{0.557}	&	0.472 	&
\textbf{0.558}	&	0.473 	\\

& $Q_2$  &  \textbf{0.595} &  0.442  &    
 \textbf{0.604}  & 0.445  &   
  \textbf{0.598}   &  0.443  &   
  \textbf{0.599}    & 0.444   \\	

& $Q_3$ &  \textbf{0.563}	&	0.477 	&
\textbf{0.566}	&	0.479 	&
\textbf{0.563}	&	0.477 	&
\textbf{0.564} &	0.478 	\\

& $Q_4$ &  0.172 	&	\textbf{0.236} 	&
0.213 	&	\textbf{0.237} 	&
0.186 	&	\textbf{0.236} 	&
0.190 	&	\textbf{0.237} 	\\

& $Q_5$ & \textbf{0.522} 	&	0.498 	&
\textbf{0.526} 	&	0.499 	&
\textbf{0.522} 	&	0.497 	&
\textbf{0.524} 	&	0.498 	\\

& $Q_6$ &   \textbf{0.624}	&	0.511 	&
\textbf{0.628}	&	0.512 	&
\textbf{0.625}	&	0.511 	&
\textbf{0.626}	&	0.513 	\\

\hline
\hline

QType-3 & $Q_1$  & \textbf{0.312}	&	0.241  	&
\textbf{0.315}	&	0.242  	&
\textbf{0.312}	&	0.241  	&
\textbf{0.313}	&	0.242  	\\

 & $Q_2$  &  \textbf{0.296} &  0.164   &    
 \textbf{0.300}  & 0.165   &   
  \textbf{0.297}   &  0.164   &   
  \textbf{0.298}    & 0.164    \\	

& $Q_3$ &  \textbf{0.332}	&	0.226  	&
\textbf{0.337}	&	0.228  	&
\textbf{0.333}	&	0.226  	&
\textbf{0.334} &	0.227  	\\

& $Q_4$ &  \textbf{0.234}  	&   0.174  	&
\textbf{0.241}  	&	0.175  	&
\textbf{0.236}  	&	0.174  	&
\textbf{0.237} 	&	0.174  	\\

& $Q_5$ & 0.209   	&	\textbf{0.249}  	&
0.214  	&	\textbf{0.252}  	&
0.209  	&	\textbf{0.250}  	&
0.211  	&	\textbf{0.251}  	\\

& $Q_6$ &   \textbf{0.361}	&	0.234  	&
\textbf{0.365}	&	0.236  	&
\textbf{0.361}	&	0.234  	&
\textbf{0.363}	&	0.235  	\\

\hline
\hline

QType-4 & $Q_1$ & \textbf{0.227}	&	0.154	&
\textbf{0.227}	&	0.155	&
\textbf{0.226}	&	0.154	&
\textbf{0.227}	&	0.154	\\
			
& $Q_2$  & \textbf{0.205}  & 0.098  &  
\textbf{0.207} &  0.103 &   		
\textbf{0.205} & 0.096  &  
\textbf{0.206} &  0.100 \\
      			
& $Q_3$ & \textbf{0.259	}&	0.153	&
\textbf{0.261}	&	0.154	&
\textbf{0.258}	&	0.153	&
\textbf{0.260}	&	0.153	\\
					
& $Q_4$ &\textbf{ 0.163	}&	0.123	&
\textbf{0.164}	&	0.127	&
\textbf{0.164}	&	0.123	&
\textbf{0.163}	&	0.125	\\

& $Q_5$ & 0.151	&	\textbf{0.172}	&
0.152	&	\textbf{0.179}	&
0.150	&	\textbf{0.174}	&
0.151	&	\textbf{0.175}	\\

& $Q_6$ & \textbf{0.236}	&	0.147	&
\textbf{0.238}	&	0.149	&
\textbf{0.235}	&	0.147	&
\textbf{0.237}	&	0.148	\\

\hline
\hline

QType-5 & $Q_1$  & \textbf{0.283}	&	0.179  	&
\textbf{0.289}	&	0.183  	&
\textbf{0.284}	&	0.180  	&
\textbf{0.286}	&	0.181  	\\

  & $Q_2$  & \textbf{0.294}	&	0.170  	&
\textbf{0.304}	&	0.175  	&
\textbf{0.298}	&	0.172  	&
\textbf{0.299}	&	0.173  	\\

  & $Q_3$  & \textbf{0.298}	&	0.171  	&
\textbf{0.302}	&	0.174  	&
\textbf{0.299}	&	0.171  	&
\textbf{0.300}	&	0.172  	\\

  & $Q_4$  & \textbf{0.274}	&	0.192  	&
\textbf{0.276}	&	0.193  	&
\textbf{0.275}	&	0.191  	&
\textbf{0.275}	&	0.192  	\\

  & $Q_5$  & \textbf{0.378}	&	0.296  	&
\textbf{0.382}	&	0.298  	&
\textbf{0.378}	&	0.295  	&
\textbf{0.380}	&	0.297  	\\

  & $Q_6$  & \textbf{0.404}	&	0.226  	&
\textbf{0.408}	&	0.229  	&
\textbf{0.405}	&	0.226  	&
\textbf{0.406}	&	0.227  	\\
 
\hline
\end{tabular}
 }
\end{table}

\setlength{\tabcolsep}{4pt}
\begin{table}[t!]
\centering
\caption{Article-driven, ICC-classification-based multi-label evaluation: global vs. local models,  with  labeling scheme \unirrempht, for all sets of book-sentence-queries (QType-1),    paraphrased-sentence-queries (QType-2), comment-queries (QType-3),    comment-sentence-queries (QType-4),  and case-queries (QType-5). (Bold values correspond to the best model for each query-set and evaluation criterion)}
\label{tab:modelliv5_UniRRemphT4_multilabel-icc}
\scalebox{0.8}{
\begin{tabular}{|l|c|cc|cc|cc|cc|}
\hline 
query-set & & 
   \multicolumn{2}{|c|}{$R$} & 
   \multicolumn{2}{|c|}{$P$} & 
   \multicolumn{2}{|c|}{$F^{\mu}$} &
   \multicolumn{2}{|c|}{$F^{M}$} 
   \\
\cline{2-10}
& $i$ & $L_i$ & $G$ & $L_i$ & $G$ & $L_i$ & $G$ & $L_i$ & $G$ 
\\ \hline  \hline
QType-1 & $Q_1$ &  

\textbf{0.247}	&	0.220	&
\textbf{0.312}	&	0.269	&
\textbf{0.268}	&	0.236	&
\textbf{0.276}	&	0.242	\\

& $Q_2$  &   
      
  \textbf{0.198}      & 0.187  &   
     \textbf{0.256}     & 0.237  &   
      \textbf{0.216}      & 0.203  &   
        \textbf{0.223}      &  0.209  \\	

& $Q_3$ &   

\textbf{0.217}	&	0.192	&
\textbf{0.287}	&	0.250	&
\textbf{0.238}	&	0.209	&
\textbf{0.247}	&	0.217	\\

& $Q_4$ &  

\textbf{0.191}	&	0.179	&
\textbf{0.234}	&	0.217	&
\textbf{0.205}	&	0.191	&
\textbf{0.210}	&	0.196	\\

& $Q_5$ &  

\textbf{0.201}	&	0.191	&
\textbf{0.253}	&	0.240	&
\textbf{0.218}	&	0.207	&
\textbf{0.224}	&	0.213	\\

& $Q_6$ &   

\textbf{0.268}	&	0.243	&
\textbf{0.306}	&	0.283	&
\textbf{0.279}	&	0.254	&
\textbf{0.286}	&	0.262	\\

\hline
\hline

QType-2 & $Q_1$  & \textbf{0.244}	&	0.194  	&
\textbf{0.310}	&	0.239  	&
\textbf{0.265}	&	0.209  	&
\textbf{0.273}	&	0.214  	\\

& $Q_2$  &  \textbf{0.215} &  0.160   &    
 \textbf{0.278}  & 0.200   &   
  \textbf{0.235}   &  0.173   &   
  \textbf{0.246}    & 0.178    \\	

& $Q_3$ &  \textbf{0.206}	&	0.164  	&
\textbf{0.280}	&	0.224  	&
\textbf{0.229}	&	0.182  	&
\textbf{0.238  	}&	0.189  	\\

& $Q_4$ &  \textbf{0.172}  	&	 0.144   	&
\textbf{0.213}  	&	 0.178   	&
\textbf{0.186}  	&	 0.156   	&
\textbf{0.190}  	&	 0.159   	\\

& $Q_5$ & \textbf{0.186} 	&	0.168  	&
\textbf{0.235} 	&	0.213  	&
\textbf{0.202} 	&	0.183  	&
\textbf{0.208} 	&	0.188  	\\

& $Q_6$ &   \textbf{0.257}	&	0.212  	&
\textbf{0.297}	&	0.248  	&
\textbf{0.268}	&	0.222  	&
\textbf{0.276}	&	0.228  	\\

\hline
\hline

QType-3 & $Q_1$  & \textbf{0.245}	&	0.167  	&
\textbf{0.319}	&	0.215  	&
\textbf{0.269}	&	0.183  	&
\textbf{0.277}	&	0.188  	\\

& $Q_2$  &  \textbf{0.170} &  0.096   &    
 \textbf{0.239}  & 0.134   &   
  \textbf{0.192}   &  0.108   &   
  \textbf{0.199}    & 0.112    \\	

& $Q_3$ &  \textbf{0.189}	&	0.133  	&
\textbf{0.265}	&	0.186  	&
\textbf{0.212}	&	0.149  	&
\textbf{0.220  	}&	0.155  	\\

& $Q_4$ &  \textbf{0.169}  	&	 0.126  	&
\textbf{0.215}  	&	 0.161   	&
\textbf{0.184}  	&	 0.138   	&
\textbf{0.189}  	&	 0.141   	\\

& $Q_5$ &  \textbf{0.154}   	&	0.146  	&
 \textbf{0.204}  	&	0.200  	&
 \textbf{0.169}   	&	0.165  	&
 \textbf{0.176}   	&	0.169  	\\

& $Q_6$ &   \textbf{0.240}	&	0.167  	&
\textbf{0.285}	&	0.211  	&
\textbf{0.252}	&	0.178  	&
\textbf{0.261}	&	0.186  	\\

\hline
\hline

QType-4 & $Q_1$ &  
			
\textbf{0.188}	&	0.122	&
\textbf{0.241}	&	0.153	&
\textbf{0.205}	&	0.132	&
\textbf{0.211}	&	0.136	\\
			
& $Q_2$  &  
 
\textbf{0.123}&  0.065 &  
\textbf{0.174} & 0.090  &  
\textbf{0.139} & 0.073  &  
\textbf{0.144} &  0.076 \\
      			
& $Q_3$ &  
			
\textbf{0.156}	&	0.094	&
\textbf{0.230}	&	0.133	&
\textbf{0.176}	&	0.105	&
\textbf{0.184}	&	0.110	\\

& $Q_4$ &  
			
\textbf{0.126}	&	0.094	&
\textbf{0.164}	&	0.122	&
\textbf{0.139}	&	0.103	&
\textbf{0.142}	&	0.106	\\

& $Q_5$ &  
			
\textbf{0.118}	&	0.115	&
\textbf{0.188}	&	0.176	&
\textbf{0.140}	&	0.134	&
\textbf{0.145}	&	0.139	\\

& $Q_6$ &  
			
\textbf{0.173}	&	0.116	&
\textbf{0.204}	&	0.143	&
\textbf{0.182}	&	0.123	&
\textbf{0.187}	&	0.129	\\

\hline
\hline

QType-5 & $Q_1$  & \textbf{0.309}	&	0.199  	&
\textbf{0.381}	&	0.237  	&
\textbf{0.334}	&	0.212  	&
\textbf{0.341}	&	0.216  	\\

  & $Q_2$  & \textbf{0.193}	&	0.110  	&
\textbf{0.265}	&	0.151  	&
\textbf{0.215}	&	0.122  	&
\textbf{0.223}	&	0.127  	\\

  & $Q_3$  & \textbf{0.244}	&	0.167  	&
\textbf{0.274}	&	0.179  	&
\textbf{0.255}	&	0.172  	&
\textbf{0.258}	&	0.173  	\\

  & $Q_4$  & \textbf{0.233}	&	0.158  	&
\textbf{0.296}	&	0.199  	&
\textbf{0.254}	&	0.172  	&
\textbf{0.261}	&	0.176  	\\

  & $Q_5$  & \textbf{0.214}	&	0.174  	&
\textbf{0.272}	&	0.224  	&
\textbf{0.234}	&	0.191  	&
\textbf{0.240}	&	0.196  	\\

  & $Q_6$  & \textbf{0.245}	&	0.159  	&
\textbf{0.276}	&	0.185  	&
\textbf{0.253}	&	0.165  	&
\textbf{0.259}	&	0.171  	\\
\hline
\end{tabular}
}
\end{table}

\subsubsection{Multi-label evaluation} 

Our investigation on the   effectiveness of global and local models was further deepened under the different multi-label evaluation contexts we devised.

{\bf Article-driven  multi-label evaluation.\ }
Table~\ref{tab:modelliv5_UniRRemphT4_multilabel-cb} and 
Table~\ref{tab:modelliv5_UniRRemphT4_multilabel-icc} report results according to the article-driven clustering-based and ICC-classification-based analyses, respectively. For the clustering-based approach, results correspond to  
 a number of clusters over the articles of each book that was set  so to have mean cluster size close to three.

At a first glance, the superiority of local models stands out, thus confirming the findings drawn from the single-label evaluation results.  
Upon a focused inspection, we observe few exceptions in Table~\ref{tab:modelliv5_UniRRemphT4_multilabel-cb} corresponding to results on Book-5  
(similarly as we already found in Table~\ref{tab:modelliv5_UniRRemphT4_QT2})   for the QType-1, QType-3, and QType-4, and on Book-4 for QType-2;  
however, this does not occur in Table~\ref{tab:modelliv5_UniRRemphT4_multilabel-icc}. 
We tend to ascribe this to the fact that, being driven by a content-similarity-based approach, the clustering of the articles is clearly conditioned on the higher topic variety observable in Book-5 or Book-4, which may  favor  a global model against a local one in better capturing topics that are outside the boundaries of that particular book.  
By contrast, this does not hold for the ICC-classification-based grouping of the articles, as it relies on an externally provided organization of the articles in a particular book that is not constrained by the   topic patterns (e.g., word co-occurrences) that might be distinctive of that  book.

Notably, looking at the QType-5 results, the performance scores obtained by local and global models are  generally comparable to or even  higher than those respectively obtained on QType-3 (or  QType-4) queries, which is particularly evident for   the ICC-classification-based grouping of the articles (Table~\ref{tab:modelliv5_UniRRemphT4_multilabel-icc}).

As a further point of investigation, we explored whether the article-driven clustering-based multi-label evaluation is sensitive to how the query relevant sets (i.e., clusters of articles) were formed, with a focus on the content representation model of the articles. In particular, we replaced the TF-IDF vectorial representation of the articles with the article embeddings generated by our \Lamberta models, while keeping the same clustering methodology and setting as used in our previous analysis. 
From the comparison  results reported in the Appendix, Table~\ref{tab:modelliv5_UniRRemphT4_multilabel-cb-Lambertaembeddings},    it stands out that using the embeddings generated by \Lamberta models to represent the articles in the clustering process is always beneficial to the multi-label classification performance of \Lamberta local models,   according to all criteria and query sets. As expected, the improvements are generally more evident for QType-1 and QType-2 query sets, since these queries have lexicons close to the training data. More interestingly,   we also observe that the performance gain by \Lamberta embeddings tends to be higher for the largest books, i.e., Book-4 and Book-5, with peaks of about 300\% percentage increase reached with the QType-2 query-set relevant for Book-4. 
Nonetheless, apart from these particular cases,   discovering clusters over a set of articles represented by their  \Lamberta embeddings does not bring in general to outstanding boost of performance against the TF-IDF vectorial representation: indeed, this can be ascribed to the very small size of the clusters produced, and hence the TF-IDF vectorial representation can well approximate and match the clusters detected over  the \Lamberta embeddings of the articles,  especially in smaller collections of articles (i.e., ICC books).

\setlength{\tabcolsep}{3.5pt}
\begin{table}[t!]
\centering
\caption{Topic-driven multi-label evaluation: global vs. local models,  with  labeling scheme \unirrempht, for all sets of ICC-heading-queries (i.e., QType-6 query sets). Column `a\_{cs}' stands for average class size, i.e., the average no. of articles belonging to each query label. (Bold values correspond to the best model for each query-set and evaluation criterion)}
\label{tab:modelliv5_UniRRemphT4_multilabel-topic}
\scalebox{0.8}{
\begin{tabular}{|c|c|cc|cc||c|cc|cc||c|cc|cc|}
\hline 
  & 
   \multicolumn{5}{|c||}{chapter} & 
   \multicolumn{5}{|c||}{subchapter} & 
   \multicolumn{5}{|c|}{section} 
   \\
\hline
 $i$ & a\_{cs}  & \multicolumn{2}{|c|}{$F^{\mu}$} & \multicolumn{2}{|c||}{$P@3$}   & a\_{cs}   & \multicolumn{2}{|c|}{$F^{\mu}$} & \multicolumn{2}{|c||}{$P@3$}  & a\_{cs}   & \multicolumn{2}{|c|}{$F^{\mu}$} & \multicolumn{2}{|c|}{$P@3$}  \\
 \cline{3-6} \cline{8-11} \cline{13-16}
   &   & $L_i$ & $G$ & $L_i$ & $G$ &   & $L_i$ & $G$ & $L_i$ & $G$ &   & $L_i$ & $G$ & $L_i$ & $G$ \\
 \hline  \hline
$Q_1$ & 26.1 & \textbf{0.314} & 0.251 & \textbf{0.867} & 0.733 & 14.4 & \textbf{0.307} & 0.253 & \textbf{0.875} & 0.792 &   7.5 & \textbf{0.348} & 0.205 & \textbf{0.722} & 0.556 \\ 
$Q_2$  & 69.0 & \textbf{0.256} & 0.247 & 0.800 & 0.800 & 11.6 & \textbf{0.167} & 0.145 & \textbf{0.519} & 0.370 & 11.8 & \textbf{0.234} & 0.089 & \textbf{0.500} & 0.333 \\
 $Q_3$  & 40.3 & \textbf{0.287} & 0.225 & \textbf{0.667} & 0.556 & 17.6 & \textbf{0.185} & 0.145 & \textbf{0.778} & 0.389 & 8.5 & \textbf{0.270} & 0.219  & \textbf{0.577} & 0.500 \\
$Q_4$  & 98.4 & \textbf{0.238} & 0.083 & \textbf{0.444} & 0.333  &  17.1 & \textbf{0.216} & 0.144 & \textbf{0.673} & 0.531 & 8.3 & \textbf{0.218} & 0.149 &  \textbf{0.653} & 0.531 \\
 $Q_5$  & 64.8 & \textbf{0.174} & 0.163 & 0.545 & \textbf{0.727} & 22.2 & \textbf{0.263}  & 0.237 & 0.733 & 0.733 & 10.9 & 0.215 & \textbf{0.219} & \textbf{0.558} & 0.512\\
 $Q_6$  & 66.2 & \textbf{0.402}& 0.198 & \textbf{1.000} & 0.600 & 21.4 & \textbf{0.319}  & 0.202 & \textbf{0.667} & 0.467 & 6.5 & \textbf{0.290}  & 0.248 & \textbf{0.571} & 0.486  \\
\hline  
\end{tabular}
}
\end{table}

{\bf Topic-driven  multi-label evaluation.\ }
Let us now focus on the topic-driven multi-label evaluation results, based on QType-6 queries,  which are shown in Table~\ref{tab:modelliv5_UniRRemphT4_multilabel-topic} (details on precision and recall values are omitted for the sake of presentation, but this does not change the remarks being discussed). 
Three major remarks arise here.  The first one is again about the higher effectiveness of local models against the global one, for each of the book query-sets, with the usual exception corresponding to Book-5, which is more evident as the type of division is finer-grain.   
The second remark concerns a comparison between the methods' performance by varying the types of book division: the chapter, subchapter and section cases are indeed quite different to each other, which should be noticed through the different values of the average number of articles ``covered'' by each query label-class; in general, higher values correspond to more abstract divisions. Moreover, the roughly monotone variation of this statistic over all books cannot be coupled with a monotonic variation of the performances: for instance, the highest F-measure   values correspond to the division at section level in Book-1, whereas they correspond to the chapter level in Books-2, 3, 4 and 6. 
Finally, it is worth noticing that the performances significantly increase when precision@3 scores are considered, often reaching very high values: this would suggest that, despite the intrinsic complexity of this evaluation task,  the models are able to guarantee a high fraction of top-3 predictions that correspond to the articles that are  relevant for each query.

\subsection{Ablation study}
\label{sec:ablation}

\setlength{\tabcolsep}{3pt}
\begin{table}[t!]
\centering
\caption{Evaluation of different local models $L_2$ for all types of query-sets pertinent to articles of Book-2. (Bold values correspond to the best model for each query-set and evaluation criterion)}
\label{tab:ablation}
\scalebox{0.8}{
\begin{tabular}{|l|l|cccc||cc||cc|}
\hline
query-set & method    & $R$ & $P$ & $F^{\mu}$ & $F^{M}$  &  $R@10$ & $MRR$ & $E$ & $E@10$ 
   \\  \hline  \hline
QType-1 &  \titlerr  	&	0.530	&	0.855	&	0.623	&	0.655	&	0.619	&	0.564	&	3.023	&	1.335	\\
&  \unirr  	&	0.675	&	0.928	&	0.755	&	0.782	&	0.851	&	0.744	&	0.343	&	0.167	\\
&  \birr  	&	0.552	&	0.775	&	0.621	&	0.645	&	0.745	&	0.635	&	0.790	&	0.342	\\
&  \trirr  	&	0.584	&	0.763	&	0.637	&	0.661	&	0.801	&	0.674	&	1.149	&	0.455	\\
&  \casrr  	&	0.555	&	0.901	&	0.653	&	0.687	&	0.712	&	0.608	&	1.422	&	0.808	\\
&  \trianglerr  	&	0.900	&	0.955	&	0.910	&	0.927	&	0.980	&	0.929	&	0.459	&	0.281	\\
&  \unirrempht  	&	\textbf{0.972}	&	\textbf{0.981} 	&	\textbf{0.971}	&	\textbf{0.977}	&	\textbf{0.999}	&	\textbf{0.983}	&	\textbf{0.149}	&	\textbf{0.089}	\\
&  \casrrempht  	&	0.823	&	0.929	&	0.843	&	0.873	&	0.903	&	0.853	&	0.863	&	0.457	\\
&  \trianglerrempht  	&	0.919	&	0.972	&	0.927	&	0.945	&	0.977	&	0.940	&	0.492	&	0.298	\\ 
\hline \hline

QType-2 &  \titlerr  	&	0.410	&	0.598	&	0.456	&	0.487	&	0.551	&	0.463	&	3.892	&	1.706	\\
&  \unirr  	&	0.549	&	0.744	&	0.605	&	0.632	&	0.754	&	0.624	&	\textbf{1.310}	&	\textbf{0.585}	\\
&  \birr  	&	0.342	&	0.470	&	0.374	&	0.396	&	0.581	&	0.436	&	2.084	&	0.899	\\
&  \trirr  	&	0.442	&	0.620	&	0.494	&	0.516	&	0.699	&	0.544	&	2.253	&	0.881	\\
&  \casrr  	&	0.383	&	0.620	&	0.444	&	0.473	&	0.597	&	0.455	&	2.289	&	1.232	\\
&  \trianglerr  	&	0.612	&	0.722	&	0.626	&	0.662	&	0.823	&	0.684	&	2.101	&	1.107	\\
&  \unirrempht  	&	\textbf{0.828}	&	\textbf{0.856}	&	\textbf{0.814}	&	\textbf{0.841}	&	\textbf{0.941}	&	\textbf{0.871}	&	1.620	&	0.732	\\
&  \casrrempht  	&	0.625	&	0.740	&	0.639	&	0.677	&	0.784	&	0.685	&	2.221	&	1.074	\\
&  \trianglerrempht  	&	0.632	&	0.729	&	0.642	&	0.677	&	0.854	&	0.705	&	2.311	&	1.174	\\
\hline \hline

QType-3 &  \titlerr  	&	0.023	&	0.008	&	0.010	&	0.012	&	0.171	&	0.073	&	6.312	&	2.816	\\
&  \unirr  	&	0.296	&	0.208	&	0.230	&	0.244	&	0.618	&	0.425	&	4.368	&	2.051	\\
&  \birr  	&	0.212	&	0.140	&	0.156	&	0.169	&	0.494	&	0.321	&	5.185	&	2.349	\\
&  \trirr  	&	0.194	&	0.113	&	0.133	&	0.143	&	0.478	&	0.295	&	5.885	&	2.523	\\
&  \casrr  	&	0.110	&	0.075	&	0.082	&	0.089	&	0.363	&	0.199	&	\textbf{4.034}	&	\textbf{1.800}	\\
&  \trianglerr  	&	0.261	&	0.186	&	0.203	&	0.217	&	0.624	&	0.394	&	4.048	&	1.939	\\
&  \unirrempht  	&	\textbf{0.313}	&	\textbf{0.213}	&	\textbf{0.239}	&	\textbf{0.253}	&	\textbf{0.655}	&	\textbf{0.445}	&	4.938	&	2.266	\\
&  \casrrempht  	&	0.232	&	0.157	&	0.176	&	0.187	&	0.525	&	0.339	&	4.717	&	2.172	\\
&  \trianglerrempht 	&	0.217	&	0.162	&	0.173	&	0.185	&	0.593	&	0.347	&	4.254	&	2.100	\\
\hline \hline

QType-4 & \titlerr  	&	0.037	&	0.042	&	0.028	&	0.039	&	0.137	&	0.076	&	6.379	&	2.800	\\
&  \unirr  	&	0.175	&	0.199	&	0.160	&	0.186	&	0.439	&	0.271	&	4.000	&	1.834	\\
&  \birr  	&	0.104	&	0.125	&	0.094	&	0.114	&	0.309	&	0.185	&	4.592	&	2.065	\\
&  \trirr  	&	0.090	&	0.101	&	0.077	&	0.096	&	0.275	&	0.163	&	5.357	&	2.241	\\
&  \casrr  	&	0.056	&	0.068	&	0.049	&	0.061	&	0.219	&	0.115	&	\textbf{3.268}	&	\textbf{1.615}	\\
&  \trianglerr  	&	0.154	&	0.178	&	0.141	&	0.165	&	0.416	&	0.253	&	4.174	&	1.991	\\

&  \unirrempht  	&	\textbf{0.216}	&	0.191	&	\textbf{0.173}	&	\textbf{0.203}	&	\textbf{0.473}	&	\textbf{0.315}	&	5.077	&	2.311	\\
&  \casrrempht  	&	0.135	&	0.161	&	0.119	&	0.147	&	0.347	&	0.211	&	4.586	&	2.130	\\
&  \trianglerrempht  	&	0.144	&	\textbf{0.202}	&	0.143	&	0.168	&	0.398	&	0.226	&	4.148	&	2.027	\\
\hline

QType-5 & \titlerr  	&	0.036 	&	0.025 	&	0.022 	&	0.029 	&	0.127 	&	0.081 	&	6.570 	&	 2.968	\\

&  \unirr  	&	0.260 	&	0.288 	&	0.232 	&	0.273 	&	0.666 	&	0.436 	&	\textbf{4.446} 	&	\textbf{2.074} 	\\

&  \birr  	&	0.232 	&	0.236 	&	0.197 	&	0.234 	&	0.548 	&	0.357 	&	4.873 	&	2.198 	\\

&  \trirr  	&	0.225 	&	0.248 	&	0.204 	&	0.236 	&	0.568 	&	0.380 	&	5.498 	&	2.408 	\\

&  \casrr  	&	0.082 	&	0.089 	&	0.068 	&	0.086 	&	0.334 	&	0.187 	&	5.429 	&	2.400 	\\

&  \trianglerr  	&	0.278 	&	\textbf{0.318} 	&	0.258 	&	0.297 	&	0.686 	&	0.434 	&	4.560 	&	2.150 	\\

&  \unirrempht  	&	\textbf{0.292}	&	0.316	&	\textbf{0.274}	&	\textbf{0.303}	&	\textbf{0.726}	&	\textbf{0.465}	&	4.972	&	2.299	\\

&  \casrrempht  &	0.238 	&	0.295 	&	0.233 	&	0.263 	&	0.579 	&	0.380 	&	4.822 	&	2.158 	\\

&  \trianglerrempht  &	0.283 	&	0.316 	&	0.273 	&	0.299 	&	0.700 	&	0.463 	&	4.638 	&	2.137 	\\
\hline

\end{tabular}
}
\end{table}

 \setlength{\tabcolsep}{3pt}
\begin{table}[t!]
\centering
\caption{Single-label evaluation of local model $L_2$ \unirrempht with $minTU=64$, for each type of query-set. Percentage values  correspond to the increase/decrease percentage of performance criteria when using $minTU=64$ compared to  $minTU=32$. (Highlighted in red are the values corresponding to a worsening when using $minTU=64$.)}
\label{tab:ablation-TU64}
\scalebox{0.8}{
\begin{tabular}{|l|cccc||cc||cc|}
\hline
query-set   & $R$ & $P$ & $F^{\mu}$ & $F^{M}$  &  $R@10$ & $MRR$ & $E$ & $E@10$ 
   \\  \hline  \hline
QType-1 &   0.971	&	0.982	&	0.971	&	0.977	&	0.996	&	0.982	&	0.090	&	0.072	\\  
 & \scriptsize \glob{-0.10\%}  &   	\scriptsize 	{+0.13\%}  & \scriptsize   		\glob{-0.04\%}  &   	\scriptsize 	\glob{-0.04\%}  &   	\scriptsize 	\glob{-0.34\%}  &   \scriptsize 		\glob{-0.13\%}  &   	\scriptsize 	{-39.40\%}  &   \scriptsize 		{-19.41\%}   \\
 \hline \hline
QType-2 & 0.797	&	0.825	&	0.781	&	0.811	&	0.928	&	0.845	&	1.266	&	0.657	\\
& \scriptsize \glob{-3.73\%}  &   	\scriptsize 	\glob{-3.67\%}  &  \scriptsize  		\glob{-4.09\%}  &   \scriptsize 		\glob{-3.62\%}  &  \scriptsize  		\glob{-1.43\%}  &   \scriptsize 		\glob{-2.97\%}  &  \scriptsize  		{-21.85\%}  &   \scriptsize 		{-10.30\%}      \\
 \hline \hline
QType-3 &  0.351	&	0.254	&	0.280	&	0.295	&	0.666	&	0.481	&	3.661	&	1.814	\\
& \scriptsize {+12.14\%}  &   	\scriptsize 	{+19.25\%}  &  \scriptsize  		{+17.15\%}  &   	\scriptsize 	{+16.60\%}  &   	\scriptsize 	{+1.68\%}  &   \scriptsize 		{+8.09\%}  &  \scriptsize  		{-25.86\%}  &   	\scriptsize 	{-19.95\%}      \\
 \hline \hline
QType-4 & 0.221	&	0.204	&	0.185	&	0.212	&	0.482	&	0.308	&	3.889	&	1.928	\\
& \scriptsize {+2.31\%}  &   \scriptsize 		{+7.37\%}  &  \scriptsize  		{+6.94\%}  &  \scriptsize  		{+4.95\%}  &  \scriptsize  		{+2.55\%}  &  \scriptsize  		\glob{-0.96\%}  &   \scriptsize 		{-23.40\%}  &   	\scriptsize 	{-16.57\%}    \\

 \hline \hline
 
QType-5 & 0.342	&	0.352 &	0.320	&	0.347	&	0.740	&	0.460	&	3.615 &	1.872	\\
 
& \scriptsize {+17.13\%}  &   \scriptsize 		{+11.33\%}  &  \scriptsize  		{+16.67\%}  &  \scriptsize  		{+14.53\%}  &  \scriptsize  		{+1.93\%}  &  \scriptsize  		\glob{-1.08\%}  &   \scriptsize 		{-27.27\%}  &   	\scriptsize 	{-18.58\%}    \\
\hline
\end{tabular}
}
\end{table}

 \setlength{\tabcolsep}{3pt}
\begin{table}[t!]
\centering
\caption{Multi-label evaluation of local model $L_2$ \unirrempht with $minTU=64$, for each type of query-set. Percentage values  correspond to the increase/decrease percentage of performance criteria when using $minTU=64$ compared to  $minTU=32$. (Highlighted in red are the values corresponding to a worsening when using $minTU=64$.)}
\label{tab:ablation-TU64-multilabel}
\scalebox{0.8}{
\begin{tabular}{|l|cccc||cccc|}
\hline
query-set & \multicolumn{4}{|c||}{Clustering-based} & \multicolumn{4}{|c|}{ICC-Classification-based}  \\ \hline 
   & $R$ & $P$ & $F^{\mu}$ & $F^{M}$  & $R$ & $P$ & $F^{\mu}$ & $F^{M}$ 
   \\  \hline  \hline
QType-1  & 0.658 &  0.668 &  0.662  &  0.663  & 0.197 & 0.252  &  0.215  &  0.221   \\

& \scriptsize	\glob{-2.08\%}	& \scriptsize	\glob{-2.20\%}	& \scriptsize	\glob{-2.07\%}	& \scriptsize	\glob{-2.21\%}	& \scriptsize	\glob{-0.51\%}	& \scriptsize	\glob{-1.56\%}	& \scriptsize	\glob{-0.46\%}	& \scriptsize	\glob{-0.90\%}	\\
 \hline \hline
 
QType-2   &  0.556  &  0.565  &  0.559 &  0.560  &  0.198  &  0.252  &  0.215   &   0.222   \\
 
& \scriptsize	\glob{-6.55\%}	& \scriptsize	\glob{-6.46\%}	& \scriptsize	\glob{-6.52\%}	& \scriptsize	\glob{-6.51\%}	& \scriptsize	\glob{-8.04\%}	& \scriptsize	\glob{-9.34\%}	& \scriptsize	\glob{-8.45\%}	& \scriptsize	\glob{-8.61\%}	\\
 \hline \hline
QType-3   &  0.322  &   0.327 & 0.324  &  0.324  &  0.166  &  0.228  & 0.185    &   0.192   \\
 
& \scriptsize	+8.78\%	& \scriptsize	+9.00\%	& \scriptsize	+9.09\%	& \scriptsize	+8.72\%	& \scriptsize	\glob{-2.35\%}	& \scriptsize	\glob{-4.60\%}	& \scriptsize	\glob{-3.65\%}	& \scriptsize	\glob{-3.52\%}	\\
 \hline \hline
QType-4   &  0.202  &  0.206  &  0.203 &  0.204  &  0.131  &  0.183  & 0.148    &   0.153   \\    
 
& \scriptsize	\glob{-0.98\%}	& \scriptsize	+0.00\%	& \scriptsize	\glob{-0.98\%}	& \scriptsize	\glob{-0.49\%}	& \scriptsize	+6.50\%	& \scriptsize	+5.17\%	& \scriptsize	+6.47\%	& \scriptsize	+6.25\%	\\
 \hline \hline
   
QType-5   &  0.312   &  0.327   &  0.319  &  0.320   &  0.181   &  0.247   & 0.201     &   0.209    \\    
   
& \scriptsize	{+6.04\%}	& \scriptsize	+7.72\%	& \scriptsize	{+7.12\%}	& \scriptsize	+7.09\%	& \scriptsize	\glob{-6.22\%}	& \scriptsize	\glob{-6.79\%}	& \scriptsize \glob{-6.51\%}	& \scriptsize	\glob{-6.28\%}	\\
\hline
\end{tabular}
}
\end{table}

\subsubsection{Training-instance labeling schemes} 
Besides the settings of BERT learning parameters, our \Lamberta models are expected to work differently depending on the choice of training-instance labeling scheme (cf. Section~\ref{sec:schemes}). Understanding how this aspect relates to the performance is fundamental, as it impacts on the  complexity of the induced model.  
 In this respect, we compared our defined   labeling schemes, using the induced local models of Book-2 (i.e., $L_2$) as case in point.

 Table~\ref{tab:ablation} shows single-label evaluation results for all types of query-sets. 
As expected, we observe that the scheme based on an article's title only is largely the worst-performing method (with the exception of QType-2 where, due to  the little impact of paraphrasing on the article  titles, it happens that the performance of {\titlerr} is   slightly better than {\casrr}).   More interestingly, lower size of $n$-gram seems to be beneficial to the effectiveness of the model, especially on QType-3, QType-4,   and QType-5  queries, indeed   {\unirr} always outperforms both {\birr} and {\trirr}; also, {\unirr} behaves better than the cascade scheme (\casrr) as well, and again the gap is more evident in the comment-based queries.  
The combination of $n$-grams of varying size reflected by the {\trianglerr} scheme leads to a significant increase in the performance  over all previously mentioned schemes,  for QType-1, QType-2,   and QType-5  queries. This would suggest that more sophisticated labeling schemes   
can lead to higher effectiveness in the learned model. 
 Nonetheless, superior performance is obtained by  considering the schemes with emphasis on the title, which all improve upon the corresponding schemes not emphasizing the title, with {\unirrempht} being the best-performing method by far according to all criteria.  
 
 It also should be noted that   the comment    and case law query  testbeds confirmed to be more difficult than the other types of queries. Also,  QType-3   and QType-5  results are found to be generally higher than those obtained for QType-4 queries, which would  indicate that, by providing a more informative context,  long (i.e., paragraph-like)  queries are better handled by \Lamberta models w.r.t.  their shorter (i.e., sentence-like)  counterparts.

\subsubsection{Training units per article} 
 The setting of the number of training units per article ($minTU$, whose default is 32), which impacts on the size of the training sets, is another model parameter that in principle deserves attention. We investigated this aspect according to both single-label and multi-label evaluation contexts, whose results are reported in    Table~\ref{tab:ablation-TU64} and Table~\ref{tab:ablation-TU64-multilabel},  respectively, for the local model learned from Book-2; note that, once again, this is for the sake of presentation, as we found out analogous remarks for other books.

  Our goal was to understand whether and to what extent doubling the $minTU$ value could lead to improve the model performance. 
  In this respect,   considering the single-label evaluation case,   we observe two different outcomes when setting $minTU$ to 64: the one corresponding to QType-1 and QType-2 queries, which unveils a general  degradation of the performance, and the other one corresponding to 
   QType-3,  QType-4, QType-5 queries, which conversely shows significant improvements according to most criteria.   
   This is remarkable as it would suggest that increasing the number of replicas in building the books' training sets is unnecessary, or even detrimental, for queries lexically close to the book articles; by contrast, it turns out to be useful for improving the classifier performance against more difficult query-testbeds like comment and case  queries.  Looking at the multi-label evaluation results, we can draw analogous considerations for the clustering-based evaluation approach (since results over QType-4 are only slightly decreased), whereas using less training replicas appears to be preferable for the   ICC-classification-based evaluation approach.

\subsection{Comparative analysis}
\label{sec:comparison}

\subsubsection{Text-based law article prediction}
\label{sec:comparison:text}
We conducted a comparative analysis of \Lamberta models with state-of-the-art text classifiers based on deep learning architectures: 
\begin{itemize}
\item 
\textit{BiLSTM}~\cite{BiLSTM,BiLSTMA}, a bidirectional LSTM model as sequence encoder. LSTM models have been widely used in text classification as   they can capture contextual information while representing the sentence by fixed size vector. 
 The model exploited in this evaluation utilizes 2 layers of BiLSTM, with 32 hidden units for each BiLSTM layer.  
 
 \item 
 \textit{TextCNN}~\cite{TextCNN}, a convolutional-neural-network-based model with multiple filter widths for text encoding and classification. 
 Every   sentence is represented as a bidimensional tensor of shape $(n,d)$, where $n$ is the sentence length and $d$ is the dimensionality of the word embedding vectors. The \textit{TextCNN} model utilizes three different filter  windows of sizes $\{3,4,5\}$, 100 feature maps for the windows' sizes,     ReLU activation function and max-pooling.

\item 
 \textit{TextRCNN}~\cite{TextRCNN}, a bidirectional LSTM with a pooling layer on the last sequence output.  
  \textit{TextRCNN} therefore combines   the recurrent neural network and convolutional network to leverage the advantages of the individual models in capturing the text  semantics. The model  first exploits a recurrent structure to learn word representations for every word in the text, thus 
  capturing  the contextual information;  
  afterwards, max-pooling is applied to determine which features are important for the classification task.

 \item 
\textit{Seq2Seq-A}~\cite{DuH18,BahdanauCB14}, a Seq2Seq model with attention  mechanism. 
 Seq2Seq models have been widely used in   machine translation and document summarization due to their capability to generate new sequences based on observed text data. For text classification, here the \textit{Seq2Seq-A}  model utilizes a  single layer BiLSTM as encoder with  32 hidden units. This encoder   learns the hidden representation for every word in an input sentence, and its final state   is then used to learn attention scores for each word in the sentence. After learning the attention weights, the weighted sum of encoder hidden states (hidden states of words) gives the attention output vector. The latter is then concatenated to the hidden representation and passed to a linear layer  to produce the final classification. 
 
\item 
\textit{Transformer} model for text classification, which is adapted from the model  originally proposed for the task of machine translation in~\cite{VaswaniSPUJGKP17}. 
The key aspect in this model is the use of an attention mechanism to deal with long range dependencies, but without resorting to RNN models.  
The encoder part of the original Transformer model is used for classification. This encoder is composed of 6 layers, each having two sub-layers, namely a multi-head attention layer, and a 2-layer   feed-forward network. 
Compared to BERT,  residual connection,  layer normalization, and    masking are discarded.    
\end{itemize}

\setlength{\tabcolsep}{2.0pt}
\begin{table}[t!]
\begin{sideways}
  \begin{minipage}{0.95\textheight}
 \centering
\caption{Competing models trained with the individual ICC books annotated with the {\unirrempht} labeling scheme, and tested over all sets of book-sentence-queries (QType-1, upper subtable),  paraphrased-sentence-queries (QType-2, second upper subtable), comment-sentence-queries (QType-4, third upper subtable),   and case-queries (QType-5, bottom subtable): best-performing values of precision, recall, and   micro-averaged F-measure. (Bold values correspond   to the best performance obtained by a competing method,  for each query-set and evaluation criterion;   \Lamberta performance values, formatted in italic,  are also reported from Table~\ref{tab:modelliv5_UniRRemphT4_QT2} to ease the comparison with the competing methods)}
\label{tab:competing_UniRRemphT4}
\begin{tabular}{|c|ccc|ccc|ccc|ccc|ccc|ccc|}
\hline 
 & 
   \multicolumn{3}{|c|}{\Lamberta} &
   \multicolumn{3}{|c|}{\textit{TextCNN}} & 
   \multicolumn{3}{|c|}{\textit{BiLSTM}}   &
    \multicolumn{3}{|c|}{\textit{TextRCNN}} &
   \multicolumn{3}{|c|}{\textit{Seq2Seq-A}} &
   \multicolumn{3}{|c|}{\textit{Transformer}} 
   \\ 
\hline
 $i$  & $P$ & $R$ & $F^{\mu}$   & $P$ & $R$ & $F^{\mu}$    & $P$ & $R$ & $F^{\mu}$    & $P$ & $R$ & $F^{\mu}$     & $P$ & $R$ & $F^{\mu}$     & $P$ & $R$ & $F^{\mu}$   
\\ \hline  \hline
  $Q_1$  
  &  \textit{0.975} &  \textit{0.962}  & \textit{0.961}   
& 0.942 & 0.911  & 0.910       & 0.752 & 0.712 & 0.680       

& 0.894  & 0.838  & 0.835     & 0.620 & 0.629 & 0.564    
  & \textbf{0.959} & \textbf{0.950} & \textbf{0.948} \\

 $Q_2$  
   &  \textit{0.981} &  \textit{0.972}  &  \textit{0.971}  
& 0.959 &  0.940 & 0.936       & 0.735 & 0.730 & 0.681      

&  0.894 & 0.852  & 0.845     &  0.630 & 0.670 & 0.585    
  & \textbf{0.965} & \textbf{0.955} & \textbf{0.954} \\

 $Q_3$ 
   &  \textit{0.994} &  \textit{0.990}   & \textit{0.990}   
& 0.962 & 0.951  & 0.951       & 0.727 & 0.724 & 0.674       
  
  & 0.905  & 0.865  & 0.861     & 0.631 & 0.641 & 0.573   
  & \textbf{0.970} & \textbf{0.965} & \textbf{0.965} \\

$Q_4$ 
  &  \textit{0.983} &  \textit{0.961}  &  \textit{0.964}  
& 0.971 & 0.933  &  0.938      & 0.789 & 0.772 & 0.736      
   
   &  0.919 & 0.875  &  0.872    & 0.727 & 0.703 & 0.665    
   & \textbf{0.969} &  \textbf{0.947}& \textbf{0.951} \\

 $Q_5$ 
   &  \textit{0.944} &  \textit{0.910}  & \textit{0.909}    
& 0.895 & 0.857  & 0.848       & 0.689 & 0.670 & 0.626      
  
  & 0.817  & 0.763  & 0.750     & 0.669 & 0.646 & 0.602    
   & \textbf{0.936} & \textbf{0.903} & \textbf{0.903} \\

 $Q_6$ 
   &  \textit{0.986} &  \textit{0.979}  &  \textit{0.978}  
& 0.949 & 0.938  &  0.935      & 0.745 & 0.723 & 0.677      
   
  & 0.903  & 0.866  &  0.858    & 0.757 & 0.718 & 0.685    
    & \textbf{0.955} & \textbf{0.957} & \textbf{0.956} 
      
\\ \hline  \hline

  $Q_1$  
    & \textit{0.866}  &  \textit{0.841}  &  \textit{0.828}    
& 0.800    & 0.769   & 0.750        & 0.360  & 0.412  & 0.349        

& 0.674   & 0.646   & 0.619      & 0.345  & 0.381  & 0.324   
  & \textbf{0.801} & \textbf{0.774}  & \textbf{0.754} \\

  $Q_2$  
    &  \textit{0.856} & \textit{0.828}   & \textit{0.814}   
& 0.760  & 0.726   & 0.706        & 0.409  & 0.442  & 0.382        
    
& 0.646   & 0.643   & 0.601      & 0.353  & 0.384  & 0.327      
  & \textbf{0.813} & \textbf{0.787}  & \textbf{0.762} \\

    $Q_3$  
      &  \textit{0.886}  &  \textit{0.861}  & \textit{0.851}   
& 0.817  & 0.765   & 0.758        & 0.464  & 0.493  & 0.435          
 
 & 0.707   & 0.672   & 0.648      & 0.361  & 0.409  & 0.342      
  & \textbf{0.852} & \textbf{0.804}  & \textbf{0.795} \\

    $Q_4$  
      &  \textit{0.756} &  \textit{0.736}  &  \textit{0.713}   
& 0.729  & 0.643   & 0.669   
     & 0.310  & 0.330  & 0.292         
  
  & 0.643   & 0.607   & 0.582      & 0.278  & 0.297  & 0.258   
  & \textbf{0.733} & \textbf{0.651}  & \textbf{0.674}
   \\

    $Q_5$  
      &  \textit{0.759} &  \textit{0.718}  &  \textit{0.710}  
& 0.714  & 0.666   & 0.649        & 0.381  & 0.399  & 0.353         
  
  & 0.637   & 0.610   & 0.580      & 0.377  & 0.393  & 0.345     

    & \textbf{0.734} & \textbf{0.715}  & \textbf{0.697}
     \\

    $Q_6$  
      & \textit{0.874} &  \textit{0.841}  &  \textit{0.833}   
& 0.788  & 0.737   & 0.725        & 0.451  & 0.451  & 0.405        
  
  & 0.662   & 0.625   & 0.603      & 0.378  & 0.438  & 0.364   
  & \textbf{0.795} & \textbf{0.756}  & \textbf{0.742} \\
  
\hline
\hline

 $Q_1$ 
   &  \textit{0.197} &  \textit{0.190} &  \textit{0.170}  
& \textbf{0.158} & \textbf{0.152} & \textbf{0.124}       & 0.013 &  0.018 & 0.013   

   & 0.146  & 0.147  &  0.120    & 0.034  & 0.019 & 0.018    & 0.114 & 0.100 & 0.089 \\
			
 $Q_2$  
   &  \textit{0.191} &  \textit{0.216}  &  \textit{0.173}  
& \textbf{0.170} & 0.155  &  \textbf{0.136}      & 0.015 & 0.037  & 0.009     

 &  0.166 & \textbf{0.167}  &  0.133    & 0.011  & 0.034 & 0.012    & 0.121 & 0.093 & 0.092 \\
      			
 $Q_3$ 
   &  \textit{0.223} &  \textit{0.241}  & \textit{0.204}   
& 0.201 &  \textbf{0.208} &  0.173      & 0.006 & 0.021  & 0.007     

 & \textbf{0.205}  & 0.204  &   \textbf{0.176}   & 0.019  & 0.029 & 0.019    & 0.118 & 0.112 & 0.102 \\
 
 $Q_4$ 
   &  \textit{0.176} &  \textit{0.189}  &  \textit{0.156}  
& 0.159 & 0.149  &  0.126      & 0.006 &  0.007 & 0.005    

&  \textbf{0.174} & \textbf{0.173}  &  \textbf{0.144}    & 0.006  & 0.016 & 0.004    & 0.073 & 0.071 & 0.061 \\
			
 $Q_5$ 
   & \textit{0.132}  &  \textit{0.092}  &  \textit{0.090}  
& \textbf{0.109} & 0.080  &  \textbf{0.090}      & 0.013 & 0.017  & 0.008       

 & 0.104  & \textbf{0.092}  &  0.085    & 0.004  & 0.087 & 0.006    & 0.089 & 0.075 & 0.070 \\
			
 $Q_6$ 
   & \textit{0.211}  &  \textit{0.224}  &  \textit{0.192}  
& 0.183 & 0.166  & 0.155       & 0.014 & 0.024  & 0.014   

& \textbf{0.186}  &  \textbf{0.185} &  \textbf{0.157}    &  0.017 & 0.065 & 0.020    & 0.120 & 0.103 & 0.090 \\

\hline
\hline

 $Q_1$ 
   & \textit{0.233}   & \textit{0.228}   &  \textit{0.210}     & 0.147   & 0.137  & 0.123     & 0.009  & 0.009  & 0.008  
   
    & \textbf{0.155}   & \textbf{0.144}   &  \textbf{0.123}     & 0.001   & 0.005  & 0.002      &  0.102  &  0.089  &   0.086 \\

 $Q_2$  
   & \textit{0.316}   & \textit{0.292}   &  \textit{0.274}     
   &  \textbf{0.186}    &  0.163     &  0.132   
    & 0.009  & 0.023  & 0.009 
    
    & 0.172   & \textbf{0.169}   &  \textbf{0.143}     & 0.023   & 0.012  & 0.016     & 0.118  & 0.092  & 0.104  \\

 $Q_3$ 
   & \textit{0.323}   & \textit{0.284}   &  \textit{0.273}     & \textbf{0.221}   & \textbf{0.267}  & \textbf{0.219}     & 0.006  & 0.013  & 0.008  
   
   & 0.178   & 0.173   &  0.151     & 0.010   & 0.019  & 0.012     & 0.193  & 0.115  & 0.124  \\

 $Q_4$ 
   & \textit{0.299}   & \textit{0.259}   &  \textit{0.250}     & \textbf{0.216}   & \textbf{0.187}  & \textbf{0.170}     & 0.007  & 0.009  & 0.004 
   
   & 0.115   & 0.112   &  0.095     & 0.021   & 0.011  & 0.011     & 0.106  & 0.088  & 0.097  \\

 $Q_5$ 
   & \textit{0.401}   & \textit{0.354}   &  \textit{0.342}     & 0.271   & \textbf{0.251}  & 0.222     & 0.016  & 0.016  & 0.009  
   
   & \textbf{0.288}   & 0.232   &  \textbf{0.226}     & 0.005   & 0.020  & 0.006     & 0.186  & 0.144  & 0.152  \\

 $Q_6$ 
   & \textit{0.445}   & \textit{0.392}   &  \textit{0.372}     & \textbf{0.357}   & \textbf{0.322}  & \textbf{0.294}     & 0.045  & 0.050  & 0.033  
   
   & 0.301   & 0.303   &  0.259     & 0.022   & 0.029  & 0.016     & 0.201  & 0.130  & 0.137  \\  

\hline
\end{tabular}

 \end{minipage}  
  \end{sideways}
  \end{table}

\vspace{2mm}
\noindent 
{\bf Evaluation requirements.\ } 
We designed this comparative evaluation to fulfill three main requirements: (i) significance of the selected competing methods, (ii) robustness of their evaluation, and (iii)  relatively fair comparison among the  competing methods and our \Lamberta models.

To meet the first requirement, we focused on deep neural network models that have been widely used for text classification and are often referred to as   ``predecessors'' of deep pre-trained language models like BERT. 

To ensure a robust   evaluation, we carried out an extensive parameter-tuning phase of each competing methods, by varying all main parameters within recommended or reasonable range values, which include:  
dropout probability (within [0.1,0.4]), 
maximum sentence length (fixed to 60, or flexible), 
batch size (base-2 powers from 16 to 128), 
number of epochs (from 10 to 100). 
The results we will present here refer to the best performances achieved by each of the models.

We considered main aspects that are shared between our models and the competing ones to achieve a fair comparison.  
 First, as each of the competing models, except the Transformer, needs word vector initialization,  they were provided with  Italian Wikipedia pre-trained Glove embeddings --- recall that Italian Wikipedia is a major constituent of the pre-trained Italian BERT used for our \Lamberta models.  
Each model was then   fine-tuned over the individual ICC books following the same data labeling schemes used for the \Lamberta models. 
Moreover, since we have to train a classification problem with as many classes as the number of articles for a given ICC book, all the   models    use an appropriate yet identical setting as in \Lamberta for the optimizer (i.e., Adam) and the loss function (i.e., cross entropy).

\vspace{2mm}
\noindent
{\bf Experimental results.\ }
The goal of this evaluation stage was to demonstrate the superiority of \Lamberta against the selected competitors for  case law retrieval as  classification task,   using different types of test queries.  
For each competitor, we used the individual books' {\unirrempht} labeled data for training. We  carried out several runs by varying the main parameters as previously discussed, and  eventually, we selected the best-performing results for each competitor and query-set, which are shown in Table~\ref{tab:competing_UniRRemphT4}.

Looking at the table, there are a few   remarks that stand out about the competitors. 
First,  the \textit{Transformer} model consistently excels over the other competitors in QType-1 and QType-2 query sets; this hints at a better effectiveness of the \textit{Transformer} approach against queries that  have some lexical affinity with the training text data. 
By contrast, when this does not hold as for comment (i.e., QType-4) and case law (i.e., QType-5) queries,  \textit{TextCNN} and \textit{TextRCNN} tend to perform better than the others. In fact, for such types of queries,  the lack of masked language modeling   in \textit{Transformer} seems not to be compensated by the attention mechanism as it happens for the QType-1   and QType-2  queries.   
Moreover, the RNN models achieve lower scores than CNN-based or more advanced models, on all query sets. This indicates  that   the ability  of \textit{BiLSTM} and \textit{Seq2Seq-A} of handling more distant token information (i.e.,    long range semantic dependency) and achieving complete abstraction at their bottom layers (without requiring multiple layer stacking like for CNNs) appear  to vanish without a deep pre-training.  In addition,   CNN models are    effective in   extracting local and position-invariant features, and it has indeed been shown   (e.g.,~\cite{0001KYS17})  that CNNs can successfully learn to classify a sentence (like QType-4 comment queries) or a paragraph (like QType-5 case law queries).

Our major finding is that, when comparing to the results obtained by \Lamberta models   (cf. first  column  of Table~\ref{tab:competing_UniRRemphT4}),   
  the best among the above competing  methods 
turn out to be outperformed by the corresponding \Lamberta models,   on all types of query.   
This confirms our initial expectation on the superiority of \Lamberta in learning    classification models from   few labeled examples per-class  
 under a tough multi-class classification scenario, having  
 a very large (i.e., in the order of hundreds) number of classes.

\subsubsection{Attribute-aware law article prediction}
\label{sec:comparison:fewshot}

\vspace{-2mm}
As discussed in Section~\ref{sec:related}, the method in~\cite{HuLT0S18} was conceived for attribute-aware charge prediction, and was evaluated on collected criminal  cases using  two types of text data: the fact of each case and the charges extracted from the penalty results of each case.  Moreover, a   number of attributes were defined as to distinguish the confusing charge pairs previously selected from the confusion matrix obtained by a charge prediction model. 

 \begin{figure}[t!]
\centering
\begin{tabular}{cc}
\includegraphics[scale=0.4]{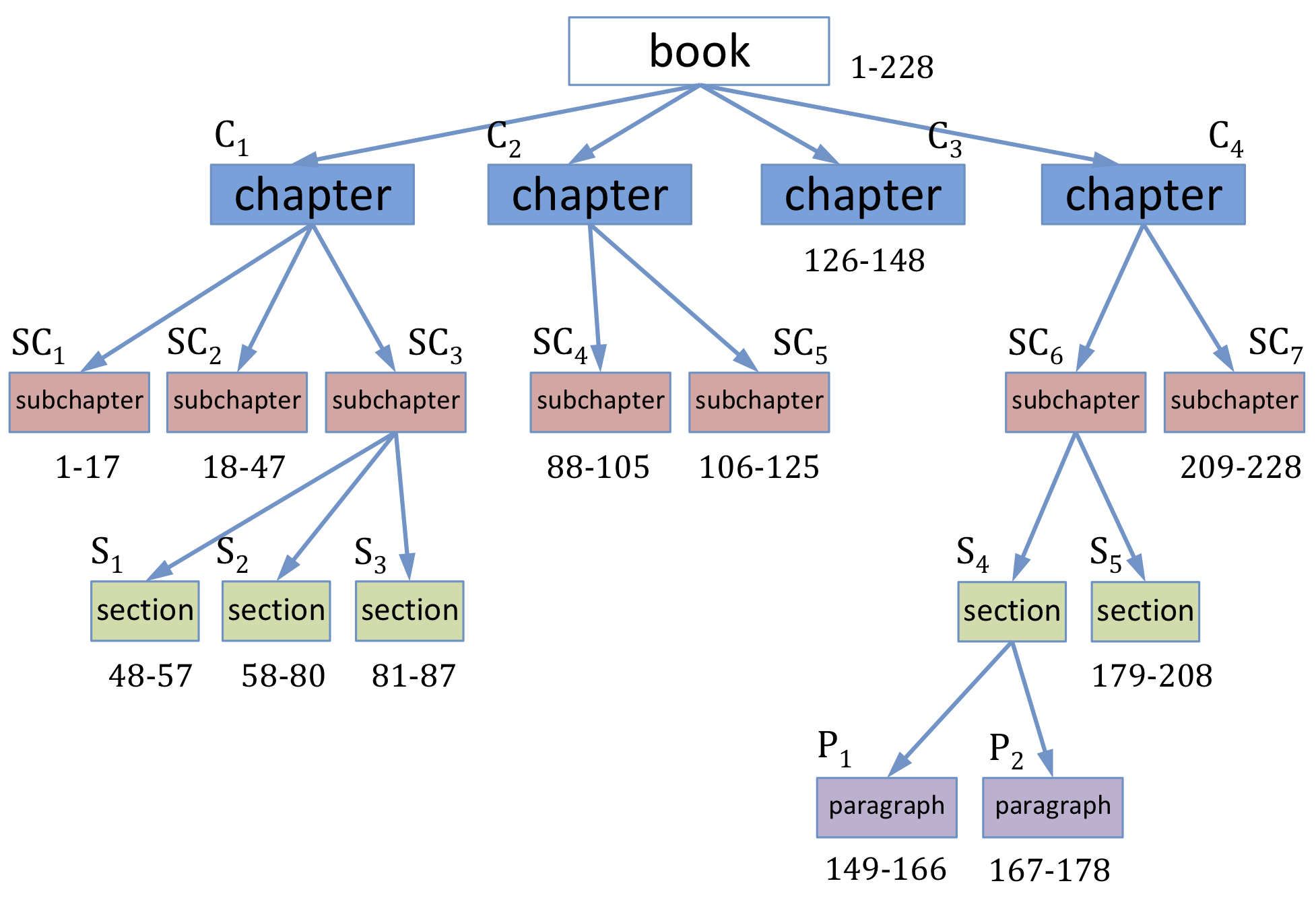} & \hspace{-3mm}
\includegraphics[width=5.5cm,height=5cm]{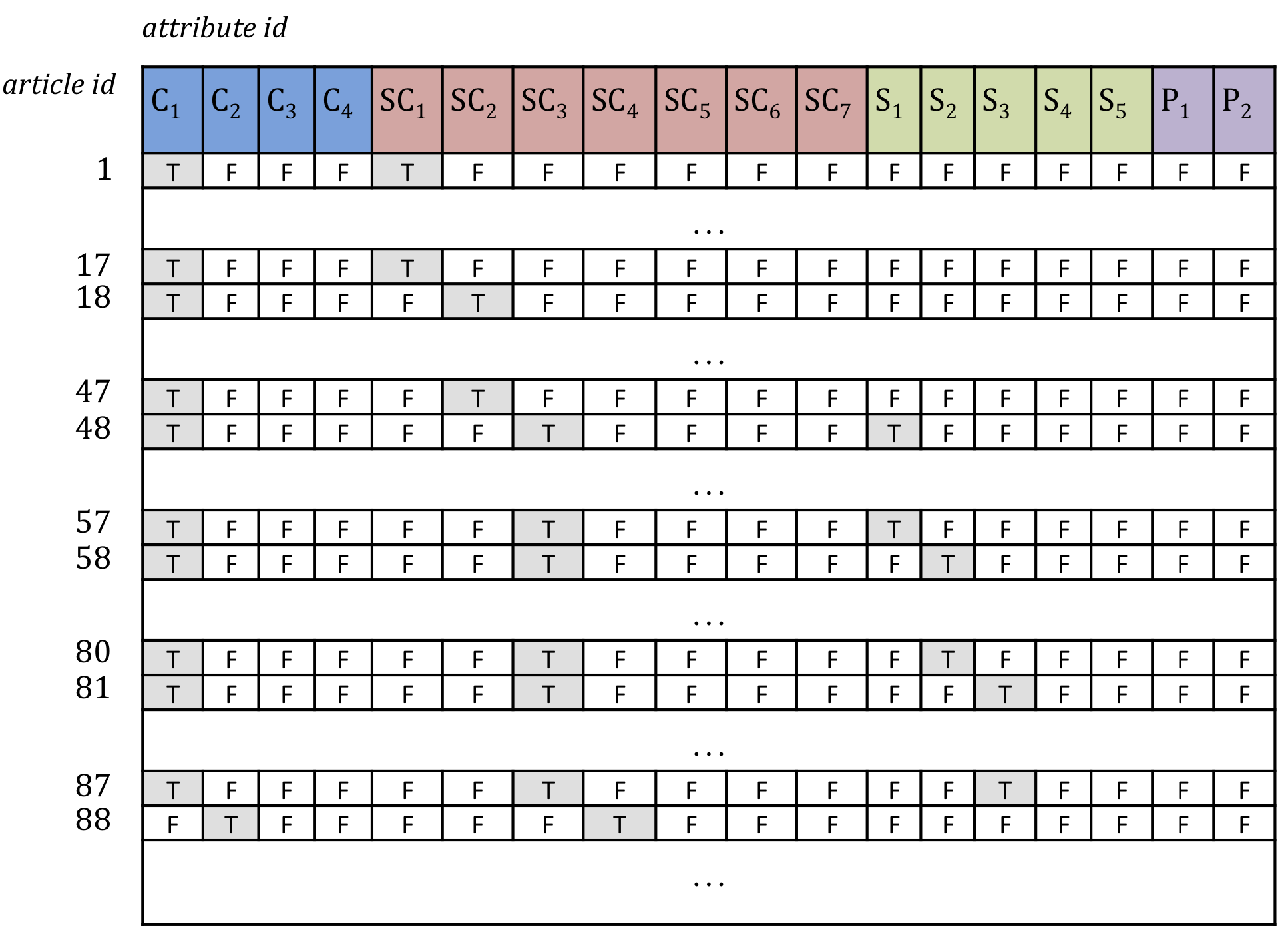} 
\end{tabular}
\caption{Attribute-aware article prediction: Example book hierarchical organization (on the left) and excerpt of its corresponding attribute-aware article representation (on the right)}
\label{fig:exampleICCtree}
\end{figure}

To apply the above   method to the ICC law article prediction task,   
we define    an \textit{attribute-aware representation} of the ICC article  data by exploiting the available ICC hierarchical labeling of the articles of each book (cf. Section~\ref{sec:data}).   
Our objective is to leverage the ICC-classification based attributes   as explicit knowledge about how to distinguish related groups of articles within the same book.  
 
 To this purpose, we adopt the following methodology. From the tree modeling the hierarchical organization into chapters, subchapters, sections and paragraphs of any given book, we treat each node as a boolean attribute and  a   complete path  (from the chapter level  to a leaf node) as an attribute-set assignment for each of the articles under that subtree path.   
  We illustrate our defined procedure in Figure~\ref{fig:exampleICCtree}, 
  where for the sake of simplicity,  the example shows  a hypothetical book containing 228 articles and  organized into four chapters (i.e., $C_1, \ldots, C_4$), seven subchapters (i.e., $SC_1, \ldots, SC_7$),   five sections (i.e., $S_1, \ldots, S_5$), and two    paragraphs (i.e., $P_1, P_2$). Thus, eighteen attributes are defined in total, and each article in the book is  associated with a binary vector (i.e., 1 means that an attribute is representative for the article, 0 otherwise); note also that the bunch of articles within the same hierarchical level subdivision share the   attribute representation. 
  
  The ICC-heading attributes resulting from the application of the above    procedure on each of the ICC books are reported in Appendix, Tables~\ref{tab:attributes_book1}--\ref{tab:attributes_book6}.

\setlength{\tabcolsep}{2.5pt}
\begin{table}[t!]
\centering
\caption{Attribute-aware  article prediction:  The \textit{A-FewShotAP} model~\cite{HuLT0S18}   trained with each  ICC book  annotated with the {\unirrempht} labeling scheme, and tested over all sets of   paraphrased-sentence-queries (QType-2,  upper subtable), comment-sentence-queries (QType-4, second upper subtable),   and case-queries (QType-5, bottom subtable). The table shows best-performing values of precision, recall, and   macro-averaged F-measure. (\Lamberta performance values, formatted in italic,  are also reported from Table~\ref{tab:modelliv5_UniRRemphT4_QT2}  to ease the comparison with the competing method)}
\label{tab:competing_ac_UniRRemphT4}
\scalebox{0.9}{
\begin{tabular}{|c|ccc||ccc|}
\hline 

 & 
   \multicolumn{3}{|c||}{\Lamberta} &
   \multicolumn{3}{|c|}{\textit{A-FewShotAP}~\cite{HuLT0S18}} 
   \\
\hline

 $i$  & $P$ & $R$ & $F^{M}$   & $P$ & $R$ & $F^{M}$   
\\ \hline  \hline

  $Q_1$  
    & \textit{0.866}  &  \textit{0.841}  &  \textit{0.853}   
& 0.355   & 0.410    & 0.381         
 \\

  $Q_2$  
    &  \textit{0.856} & \textit{0.828}   & \textit{0.841}   
& 0.328   & 0.378    & 0.351    
  \\

    $Q_3$  
      &  \textit{0.886}  &  \textit{0.861}  & \textit{0.873}   
& 0.450   & 0.514    & 0.480    
  \\

    $Q_4$  
      &  \textit{0.756} &  \textit{0.736}  &  \textit{0.746}  
& 0.319   & 0.384    & 0.348      
  \\

    $Q_5$  
      &  \textit{0.759} &  \textit{0.718}  &  \textit{0.738}  
& 0.382   & 0.444    & 0.411         
  \\

    $Q_6$  
      & \textit{0.874} &  \textit{0.841}  &  \textit{0.857}   
& 0.412   & 0.474    & 0.441     
  \\
  
\hline
\hline

 $Q_1$ 
   &  \textit{0.197} &  \textit{0.190} &  \textit{0.194}  
&    0.018  &  0.031  & 0.023    \\
			 
 $Q_2$  
   &  \textit{0.191} &  \textit{0.216}  &  \textit{0.203}  
&    0.023  &  0.043  & 0.030    \\
      		 	
 $Q_3$ 
   &  \textit{0.223} &  \textit{0.241}  & \textit{0.231}   
&    0.027  &  0.038  & 0.032    \\
 
 $Q_4$ 
   &  \textit{0.176} &  \textit{0.189}  &  \textit{0.182}  
&    0.029  &  0.047  & 0.036    \\
		 
 $Q_5$ 
   & \textit{0.132}  &  \textit{0.092}  &  \textit{0.108}
&    0.035  &  0.042  & 0.038    \\
		 	
 $Q_6$ 
   & \textit{0.211}  &  \textit{0.224}  &  \textit{0.218}  
&    0.031  &  0.037  & 0.034    \\

\hline
\hline

 $Q_1$ 
   & \textit{0.233}   & \textit{0.228}   &  \textit{0.230}    &  0.006   &  0.013    &  0.009    \\

 $Q_2$  
   & \textit{0.316}   & \textit{0.292}   &  \textit{0.303}     &  0.016   &  0.021    &  0.018    \\

 $Q_3$ 
   & \textit{0.323}   & \textit{0.284}   &  \textit{0.302}  &  0.016   &  0.018    &  0.017    \\

 $Q_4$ 
   & \textit{0.299}   & \textit{0.259}   &  \textit{0.278}   &  0.007   &  0.010    &  0.008    \\

 $Q_5$ 
   & \textit{0.401}   & \textit{0.354}   &  \textit{0.376}   &  0.018  &  0.027    &  0.022    \\

 $Q_6$ 
   & \textit{0.445}   & \textit{0.392}   &  \textit{0.417}    &   0.025  & 0.037     &  0.030    \\

\hline

\end{tabular}
}
\end{table}

\vspace{2mm}
\noindent
{\bf Experimental settings and results.\ }
We resorted to the software implementation of the method in~\cite{HuLT0S18} provided by the authors\footnote{https://github.com/thunlp/attribute\_charge} and made minimal adaptations to the code so to enable the method to work with our ICC data. We hereinafter refer to this method as \textit{attribute-aware few-shot article prediction} model, \textit{A-FewShotAP}.  
 
Analogously to the previous stage of comparative evaluation, 
 we conducted an extensive parameter-tuning phase of \textit{A-FewShotAP},  by considering both default values and variations for the main parameters, such as number of layers, learning rate, number of epochs, batch size. 
 Moreover, like for the other competitors, we used Italian Wikipedia pre-trained embeddings.

Table~\ref{tab:competing_ac_UniRRemphT4} summarizes the best-performing results obtained by \textit{A-FewShotAP} over different types of query sets, namely paraphrased-sentence-queries, comment-sentence-queries, and case-queries. 
  
As it can be observed, \textit{A-FewShotAP} performance scores reveal to be  never comparable to those produced by our \Lamberta models. Indeed, despite no information on article attributes is exploited in  \Lamberta models,    the latter   achieve a large effectiveness gain over  \textit{A-FewShotAP}, regardless of the type of query. 

Nonetheless, it is also worth noticing the beneficial effect of the attribute-aware article prediction when comparing  \textit{A-FewShotAP} with  other RNN-based competing methods, particularly \textit{BiLSTM} and \textit{Seq2Seq-A}. Notably, although  \textit{A-FewShotAP} builds on a conventional LSTM --- unlike   \textit{BiLSTM} and \textit{Seq2Seq-A}, which  utilize a bidirectional LSTM --- it can achieve comparable or even higher performance than the other two methods (cf.  Table~\ref{tab:competing_UniRRemphT4}): in fact, on QType-4 queries,  \textit{A-FewShotAP} behaves  better   than \textit{BiLSTM} and (especially in terms of precision) than   \textit{Seq2Seq-A}, while it performs  better than \textit{Seq2Seq-A} and is generally as good as 
\textit{BiLSTM} on   QType-2 and QType-5 queries.


\section{Conclusions}
\label{sec:conclusion}

We presented \Lamberta, a novel BERT-based language understanding framework for law article  retrieval as a prediction task.   
One key aspect of \Lamberta is that it is  designed to deal with a challenging learning scenario, where the  multi-class classification setting is characterized by hundreds of classes and very few, per-class training instances that are generated in an unsupervised fashion. 

 The purpose of our research is to show that a deep-learning-based civil-law article retrieval method  can be helpful not only to legal experts to reduce their workload,   but also to citizens who can benefit from such a  system to save their search and consultation needs.  
 Note also that, while focusing on  the Italian Civil Code in its current version, the \Lamberta architecture can easily be generalized to learn from other law code systems.

We are currently working on an important extension of \Lamberta  
 to enhance its capability of understanding patterns between different parts from a collection of law corpora (as it's the case of the books within the ICC), but also to enable \Lamberta to learn from external, heterogeneous legal sources, particularly to embed comments on the code and judicial decisions, also for entailment tasks. 
 We believe this next step will contribute to the  
development of powerful tools that can model both sides of the same coin in predictive justice, so to deliver unprecedented solutions to different problems in artificial intelligence for law.

The ICC corpus and evaluation data are made available to the research community at 
\url{https://people.dimes.unical.it/andreatagarelli/ai4law/}.

\newpage 
\section*{Appendix}
\label{sec:appendix}

Table~\ref{tab:modelliv5_UniRRemphT4_multilabel-cb-Lambertaembeddings} compares article-driven multi-label performance results based on clusters of articles modeled either through TF-IDF vectors or \Lamberta embeddings. Note that, to ease the comparison, the results under the `TF-IDF vectors' columns correspond to the performance of local models  reported in Table~\ref{tab:modelliv5_UniRRemphT4_multilabel-cb}.

\setlength{\tabcolsep}{4pt}
\begin{table}[t!]
\centering
\caption{Article-driven, Clustering-based multi-label evaluation: comparison of results obtained by local models,  with  labeling scheme \unirrempht, 
based on clusters over TF-IDF representation vs. clusters over \Lamberta embeddings of the ICC articles, for all sets of book-sentence-queries (QType-1),    paraphrased-sentence-queries (QType-2), comment-queries (QType-3),   comment-sentence-queries (QType-4),  and case-queries (QType-5). (Bold values correspond to the best model for each query-set and evaluation criterion) }
\label{tab:modelliv5_UniRRemphT4_multilabel-cb-Lambertaembeddings}
\scalebox{0.8}{
\begin{tabular}{|l|c|cc|cc|cc|cc|}
\hline 
 
query-set & & 
   \multicolumn{2}{|c|}{$R$} & 
   \multicolumn{2}{|c|}{$P$} & 
   \multicolumn{2}{|c|}{$F^{\mu}$} &
   \multicolumn{2}{|c|}{$F^{M}$} 
   \\
\cline{2-10}
 
& $i$ & TF-IDF & \Lamberta & TF-IDF & \Lamberta & TF-IDF & \Lamberta & TF-IDF & \Lamberta
\\ 
 
&     &  vectors  &   embeddings &  vectors  &   embeddings &  vectors  &   embeddings &  vectors  &   embeddings
\\ \hline  \hline
 
QType-1 & $Q_1$  &   {0.615}	&	\textbf{0.641} 	&
{0.617}	&	\textbf{0.643} 	&
 {0.615}	&	\textbf{0.641} 	&
 {0.616}	&	\textbf{0.642} 	\\

& $Q_2$  &   {0.674} &  \textbf{0.682}  &    
  {0.684}  & \textbf{0.687}  &   
   {0.676}   &  \textbf{0.684}  &   
  {0.679}    & \textbf{0.685}   \\	
 
& $Q_3$ &  0.623	&	\textbf{0.677} 	&
{0.626}	&	\textbf{0.674} 	&
{0.624}	&	\textbf{0.675} 	&
{0.625}&	\textbf{0.675} 	\\

& $Q_4$ &  {0.331}	&	\textbf{0.454} 	&
{0.326}	&	\textbf{0.456} 	&
{0.324}	&	\textbf{0.454} 	&
{0.328}	&	\textbf{0.455} 	\\

& $Q_5$ & 0.631	&	\textbf{0.719}	&
0.633	&	\textbf{0.722}	&
0.628	&	\textbf{0.719}	&
0.632	&	\textbf{0.720}	\\

& $Q_6$ &   0.711	&	\textbf{0.781} 	&
0.720	&	\textbf{0.784} 	&
0.710	&	\textbf{0.782} 	&
0.715	&	\textbf{0.782} 	\\

\hline  \hline

QType-2 & $Q_1$  & 0.557	&	\textbf{0.575}  	&
0.559	&	\textbf{0.577}  	&
0.557	&	\textbf{0.575}  	&
0.558	&	\textbf{0.576}  	\\
 
& $Q_2$  &  0.595 &  \textbf{0.662}   &    
 0.604  & \textbf{0.666}   &   
  0.598   &  \textbf{0.663}   &   
  0.599    & \textbf{0.664}    \\	
 
& $Q_3$ &  0.563	&	\textbf{0.859}  	&
0.566	&	\textbf{0.884}  	&
0.563	&	\textbf{0.849}  	&
0.564 &	\textbf{0.871}  	\\
 
& $Q_4$ &  0.172 	&	\textbf{0.748} 	&
0.213 	&	\textbf{0.769} 	&
0.186 	&	\textbf{0.727} 	&
0.190 	&	\textbf{0.759} 	\\
 
& $Q_5$ & 0.522 	&	\textbf{0.606}  	&
0.526 	&	\textbf{0.609}  	&
0.522 	&	\textbf{0.606}  	&
0.524 	&	\textbf{0.607}  	\\
 
& $Q_6$ &   0.624	&	\textbf{0.682}  	&
0.628	&	\textbf{0.690}  	&
0.625	&	\textbf{0.684}  	&
0.626	&	\textbf{0.686}  	\\

\hline
\hline

QType-3 & $Q_1$  & {0.312}	&	\textbf{0.329}   	&
{0.315}	&	\textbf{0.333}   	&
{0.312}	&	\textbf{0.331}   	&
{0.313}	&	\textbf{0.331}   	\\
 
 & $Q_2$  &  {0.296} &  \textbf{0.320}    &    
 {0.300}  & \textbf{0.322}    &   
  {0.297}   &  \textbf{0.320}    &   
  {0.298}    & \textbf{0.321}     \\	
 
& $Q_3$ &  {0.332}	&	\textbf{0.349}   	&
{0.337}	&	\textbf{0.353}   	&
{0.333}	&	\textbf{0.350}   	&
{0.334} &	\textbf{0.351}   	\\
 
& $Q_4$ &   0.234   	&   \textbf{0.249}   	&
 0.241   	&	\textbf{0.253}   	&
 0.236    	&	\textbf{0.250}   	&
 0.237  	&	\textbf{0.251}   	\\
 
& $Q_5$ & 0.209   	&	\textbf{0.330}  	&
0.214  	&	\textbf{0.333}  	&
0.209  	&	\textbf{0.330}  	&
0.211  	&	\textbf{0.331}  	\\
 
& $Q_6$ &   {0.361}	&	\textbf{0.372}   	&
{0.365}	&	\textbf{0.373}   	&
{0.361}	&	\textbf{0.371}   	&
{0.363}	&	\textbf{0.372}   	\\

\hline
\hline
 
QType-4 & $Q_1$ & 0.227	&   \textbf{0.230} 	&
0.227	&	\textbf{0.234} 	&
0.226	&	\textbf{0.231} 	&
0.227	&	\textbf{0.232} 	\\
			 
& $Q_2$  & 0.205  & \textbf{0.222}   &  
0.207 &  \textbf{0.223}  &   		
0.205 & \textbf{0.223}   &  
0.206 &  \textbf{0.223}  \\
      		 
& $Q_3$ & 0.259 &	\textbf{0.265} 	&
0.261	&	\textbf{0.268} 	&
0.258	&	\textbf{0.265} 	&
0.260	&	\textbf{0.266} 	\\
			 	
& $Q_4$ & 0.163 &	\textbf{0.171} 	&
0.164	&	\textbf{0.175} 	&
0.164	&	\textbf{0.172} 	&
0.163	&	\textbf{0.173} 	\\

& $Q_5$ & 0.151	&	\textbf{0.180}	&
0.152	&	\textbf{0.183}	&
0.150	&	\textbf{0.180}	&
0.151	&	\textbf{0.181}	\\

& $Q_6$ & 0.236	&	\textbf{0.246} 	&
0.238	&	\textbf{0.250} 	&
0.235	&	\textbf{0.247} 	&
0.237	&	\textbf{0.248} 	\\

\hline
\hline

QType-5 & $Q_1$  & {0.283}	&	\textbf{0.290}   	&
{0.289}	&	\textbf{0.294}   	&
{0.284}	&	\textbf{0.292}   	&
{0.286}	&	\textbf{0.292}   	\\
  & $Q_2$  &  0.294 	&	\textbf{0.300}   	&
 0.304 	&	\textbf{0.311}   	&
 0.298 	&	\textbf{0.305}   	&
 0.299 	&	\textbf{0.274}   	\\
  & $Q_3$  & {0.298}	&	\textbf{0.393}   	&
{0.302}	&	\textbf{0.391}   	&
{0.299}	&	\textbf{0.392}   	&
{0.300}	&	\textbf{0.392}   	\\
  & $Q_4$  & 0.274	&	\textbf{0.282}   	&
0.276	&	\textbf{0.285}   	&
0.275	&	\textbf{0.282}   	&
0.275	&	\textbf{0.283}   	\\
  & $Q_5$  & 0.378	&	\textbf{0.407}   	&
0.382	&	\textbf{0.411}   	&
0.378	&	\textbf{0.408}   	&
0.380	&	\textbf{0.409}   	\\
  & $Q_6$  & 0.404	&	\textbf{0.454}   	&
{0.408}	&	\textbf{0.456}   	&
{0.405}	&	\textbf{0.454}   	&
{0.406}	&	\textbf{0.455}   	\\ 
\hline
 
\end{tabular}
 }
\end{table}

Tables~\ref{tab:attributes_book1}--\ref{tab:attributes_book6} list the ICC-heading attributes, for  each of the ICC books, which were used in the task of attribute-aware law prediction.

\begin{table}[t!]
\centering
\caption{Unique headings of book subdivisions originally provided in the ICC and used in the attribute-aware article prediction evaluation task}
\label{tab:attributes_book1}
\begin{tabular}{|c|l|l|}
\hline 
 & attributes & attributes (translated in English)\\
 \hline \hline  
\textsl{Book-1}  & 
\begin{minipage}[t]{0.42\columnwidth}%
\tiny
{1: `persone fisiche', 2: `persone giuridiche', 3: `disposizioni generali', 4: `associazioni e fondazioni', 5: `associazioni non riconosciute e comitati', 6: `domicilio e residenza', 7: `assenza e dichiarazione di morte presunta', 8: `assenza', 9: `dichiarazione di morte presunta', 10: `ragioni eventuali che competono alla persona di cui si ignora l'esistenza o di cui \`{e} stata dichiarata la morte presunta', 11: `parentela e affinit\`{a}', 12: `matrimonio', 13: `promessa di matrimonio', 14: `matrimonio celebrato davanti a ministri del culto cattolico e matrimonio celebrato davanti a ministri dei culti ammessi nello stato', 15: `matrimonio celebrato davanti all'ufficiale dello stato civile', 16: `condizioni necessarie per contrarre matrimonio', 17: `formalit\`{a} preliminari del matrimonio', 18: `opposizioni al matrimonio', 19: `celebrazione del matrimonio', 20: `matrimonio dei cittadini in paese straniero e degli stranieri nel regno', 21: `nullit\`{a} del matrimonio', 22: `prove della celebrazione del matrimonio', 23: `disposizioni penali', 24: `diritti e doveri che nascono dal matrimonio', 25: `scioglimento del matrimonio e separazione dei coniugi', 26: `regime patrimoniale della famiglia', 27: `fondo patrimoniale', 28: `comunione legale', 29: `comunione convenzionale', 30: `regime di separazione dei beni', 31: `impresa familiare', 32: `stato di figlio', 33: `presunzione di paternit\`{a}', 34: `prove della filiazione', 35: `azione di disconoscimento e azioni di contestazione e di reclamo dello stato di figlio', 36: `riconoscimento dei figli nati fuori dal matrimonio', 37: `dichiarazione giudiziale della paternit\`{a} e della maternit\`{a}', 38: `legittimazione dei figli naturali', 39: `adozione di persone maggiori di et\`{a}', 40: `adozione di persone maggiori di et\`{a} e suoi effetti', 41: `forme dell'adozione di persone di maggiore et\`{a}', 42: `responsabilit\`{a} genitoriale e diritti e doveri del figlio', 43: `diritti e doveri del figlio', 44: `esercizio della responsabilit\`{a} genitoriale a seguito di separazione, scioglimento, cessazione degli effetti civili, annullamento, nullit\`{a} del matrimonio ovvero all'esito di procedimenti relativi ai figli nati fuori del matrimonio', 45: `ordini di protezione contro gli abusi familiari', 46: `tutela e emancipazione', 47: `tutela dei minori', 48: `giudice tutelare', 49: `tutore e protutore', 50: `esercizio della tutela', 51: `cessazione del tutore dall'ufficio', 52: `rendimento del conto finale', 53: `emancipazione', 54: `affiliazione e affidamento', 55: `misure di protezione delle persone prive in tutto od in parte di autonomia', 56: `amministrazione di sostegno', 57: `interdizione, inabilitazione e incapacit\`{a} naturale', 58: `alimenti', 59: `atti dello stato civile'

\

}
\end{minipage}
& 
\begin{minipage}[t]{0.42\columnwidth}%
\tiny
{
1: `natural persons', 2: `legal persons', 3: `general provisions', 4: `associations and foundations', 5: `unrecognized associations and committees', 6: `domicile and residence', 7: `absence and declaration of presumed death', 8: `absence', 9: `declaration of presumed death', 10: `possible reasons for the person whose existence is unknown or whose existence has been declared presumed death', 11: `kinship and affinity', 12: `marriage', 13: `promise of marriage', 14: `marriage celebrated before Catholic ministers and marriage celebrated before ministers of worship admitted in the state', 15: `marriage celebrated before the registrar', 16: `necessary conditions for marriage', 17: `preliminary formalities for marriage', 18: `opposition to marriage', 19: `celebration of marriage', 20: `marriage of citizens in a foreign country and foreigners in the kingdom', 21: `nullity of marriage', 22: `proof of celebration of marriage', 23: `penal provisions', 24: `rights and duties arising from marriage', 25: `dissolution of marriage and separation of spouses', 26: `family property regime', 27: `patrimonial fund', 28: `legal community', 29: `conventional community', 30: `property separation regime', 31: `family business', 32: `child status', 33: `presumption of paternity', 34: `evidence of filiation', 35: `action of denial and actions of contestation and complaint of the child status', 36: `recognition of children born out of wedlock', 37: `judicial declaration of paternity and maternity', 38: `legitimation of natural children', 39: `adoption of older people', 40: `adoption of older people and its effects', 41: `forms of adoption of older people', 42: `parental responsibility and rights and duties of the child', 43: `rights and duties of the child', 44: `practice of parental responsibility following separation, dissolution, termination of civil effects, annulment, nullity of marriage or the outcome of proceedings relating to children born out of wedlock', 45: `protection orders against family abuse', 46: `guardianship and emancipation', 47: `guardianship of minors', 48: `tutelary judge', 49: `guardian and protutore', 50: `practice of guardianship', 51: `termination of the guardian from office', 52: `return of the final account', 53: `emancipation', 54: `affiliation and custody', 55: `protection measures for persons lacking in whole or in part autonomy', 56: `support administration', 57: `disqualification, disability and natural incapacity', 58: `alimony', 59: `civil status documents'	
}
\end{minipage} 
\\ \hline 
\end{tabular}
\end{table}

\begin{table}[t!]
\centering
\caption{\textit{(cont.)} Unique headings of book subdivisions originally provided in the ICC and used in the attribute-aware article prediction evaluation task}
\label{tab:attributes_book2}
\begin{tabular}{|c|l|l|}
\hline 
 & attributes & attributes (translated in English)\\
 \hline \hline  
\textsl{Book-2}  & 
\begin{minipage}[t]{0.42\columnwidth}%
\tiny
{1: `disposizioni generali sulle successioni', 2:  'apertura della successione, delazione e acquisto dell'eredit\`{a}', 3: `capacit\`{a} di succedere', 4: `indegnit\`{a}', 5: `rappresentazione', 6: `accettazione dell'eredit\`{a}', 7: `disposizioni generali', 8: `beneficio d'inventario', 9: `separazione dei beni del defunto da quelli dell'erede', 10: `rinunzia all'eredit\`{a}', 11: `eredit\`{a} giacente', 12: `petizione di eredit\`{a}', 13: `legittimari', 14: `diritti riservati ai legittimari', 15: `reintegrazione della quota riservata ai legittimari', 16: `successioni legittime', 17: `successione dei parenti', 18: `successione del coniuge', 19: `successione dello stato', 20: `successioni testamentarie', 21: `capacit\`{a} di disporre per testamento', 22: `capacit\`{a} di ricevere per testamento', 23: `forma dei testamenti', 24: `testamenti ordinari', 25: `testamenti speciali', 26: `pubblicazione dei testamenti olografi e testamenti segreti', 27: `istituzione di erede e legati', 28: `disposizioni condizionali, a termine e modali', 29: `legati', 30: `diritto di accrescimento', 31: `revocazione delle disposizioni testamentarie', 32: `sostituzioni', 33: `sostituzione ordinaria', 34: `sostituzione fedecommissaria', 35: `esecutori testamentari', 36: `divisione', 37: `collazione', 38: `pagamento dei debiti', 39: `effetti della divisione e garanzia delle quote', 40: `annullamento e rescissione in materia di divisione', 41: `patto di famiglia', 42: `donazioni', 43: `capacit\`{a} di disporre e di ricevere per donazione', 44: `forma e effetti della donazione', 45: `revocazione delle donazioni'

\

}
\end{minipage}
& 
\begin{minipage}[t]{0.42\columnwidth}%
\tiny
{
1: `general provisions on succession', 2: `opening of succession, delation and acquisition of inheritance', 3: `ability to succeed', 4: `unworthiness', 5: `representation', 6: `acceptance of the inheritance', 7: `general provisions', 8: `inventory benefit', 9: `separation of the deceased's assets from those of the heir', 10: `disclaim of inheritance', 11: `lying inheritance', 12: `petition of inheritance', 13: `heirs', 14: `rights reserved for the heirs', 15: `reinstatement of the quota reserved for the heirs', 16: `legitimate succession', 17: `succession of relatives', 18: `succession of spouse', 19: `succession of the state', 20: `testamentary successions', 21: `ability to dispose by testament', 22: `ability to receive by testamemt', 23: `form of testaments', 24: `ordinary testaments ', 25: `special testaments', 26: `publication of holographic testaments and secret testaments', 27: `establishment of heirs and legacies', 28: `conditional provisions, forward and modal', 29: `legacies', 30: `right of accretion', 31: `revocation of testamentary dispositions', 32: `substitutions', 33: `ordinary substitutions', 34: `fidecommissary substitutions', 35: `testamentary executors', 36: `division', 37: `collation', 38: `payment of debts', 39: `effects of division and guarantee of quotas', 40: `annulment and rescission of division', 41: `family pact', 42: `donations', 43: `ability to dispose and receive by donation', 44: `forms and effects of donation', 45: `revocation of donations'

\

}
\end{minipage} 
\\ \hline 
\end{tabular}
\end{table}

\begin{table}[t!]
\centering
\caption{\textit{(cont.)} Unique headings of book subdivisions originally provided in the ICC and used in the attribute-aware article prediction evaluation task}
\label{tab:attributes_book3}
\begin{tabular}{|c|l|l|}
\hline 
 & attributes & attributes (translated in English)\\
 \hline \hline  
\textsl{Book-3}  & 
\begin{minipage}[t]{0.42\columnwidth}%
\tiny
{1: `beni', 2: `beni in generale', 3: `beni nell'ordine corporativo', 4: `beni immobili e mobili', 5: `frutti', 6: `beni appartenenti allo stato, agli enti pubblici e agli enti ecclesiastici', 7: `propriet\`{a}', 8: `disposizioni generali', 9: `propriet\`{a} fondiaria', 10: `riordinamento della propriet\`{a} rurale', 11: `bonifica integrale', 12: `vincoli idrogeologici e difese fluviali', 13: `propriet\`{a} edilizia', 14: `distanze nelle costruzioni, piantagioni e scavi, e muri, fossi e siepi interposti tra i fondi', 15: `luci e vedute', 16: `stillicidio', 17: `acque', 18: `modi di acquisto della propriet\`{a}', 19: `occupazione e invenzione', 20: `accessione, specificazione, unione e commistione', 21: `azioni a difesa della propriet\`{a}', 22: `superficie', 23: `enfiteusi', 24: `usufrutto, uso e abitazione', 25: `usufrutto', 26: `diritti nascenti dall'usufrutto', 27: `obblighi nascenti dall'usufrutto', 28: `estinzione e modificazioni dell'usufrutto', 29: `uso e abitazione', 30: `servit\`{u} prediali', 31: `servit\`{u} coattive', 32: `acquedotto e scarico coattivo', 33: `appoggio e infissione di chiusa', 34: `somministrazione coattiva di acqua a un edificio o a un fondo', 35: `passaggio coattivo', 36: `elettrodotto coattivo e passaggio coattivo di linee teleferiche', 37: `servit\`{u} volontarie', 38: `servit\`{u} acquistate per usucapione e per destinazione del padre di famiglia', 39: `esercizio delle servit\`{u}', 40: `estinzione delle servit\`{u}', 41: `azioni a difesa delle servit\`{u}', 42: `alcune servit\`{u} in materia di acque', 43: `servit\`{u} di presa o di derivazione di acqua', 44: `servit\`{u} degli scoli e degli avanzi di acqua', 45: `comunione', 46: `comunione in generale', 47: `condominio negli edifici', 48: `possesso', 49: `effetti del possesso', 50: `diritti e obblighi del possessore nella restituzione della cosa', 51: `possesso di buona fede di beni mobili', 52: `usucapione', 53: `azioni a difesa del possesso', 54: `denunzia di nuova opera e di danno temuto'

\

}
\end{minipage}
& 
\begin{minipage}[t]{0.42\columnwidth}%
\tiny
{
1: `assets', 2: `assets in general', 3: `assets in the corporate order', 4: `immovable and movable assets', 5: `civil fruits', 6: `assets belonging to the state, to public authorities and to ecclesiastical authorities', 7: `property', 8: `general provisions',  9: `landed property', 10: `rearrangement of rural property', 11: `integral reclamation', 12: `hydrogeological constraints and river defenses', 13: `building property', 14: `distances in buildings, plantations and excavations, and walls, ditches and hedges interposed between the bottoms', 15: `lights and views', 16: `dripping', 17: `waters', 18: `ways of purchasing property', 19 : `occupation and invention', 20: `accession, specification, union and admixture', 21: `actions in defense of property', 22: `surface', 23: `emphyteusis', 24: `usufruct, use and dwelling', 25: `usufruct', 26: `rights arising from usufruct', 27: `obligations arising from usufruct', 28: `extinction and modifications of usufruct', 29: `use and dwelling', 30: `predial servitude', 31: `coercive servitude', 32: `aqueduct and coercive drainage', 33: `support and lock driving', 34: `coercive administration of water to a building or ground', 35: `coercive passage', 36: `coercive power line and coercive passage of cableways', 37: `voluntary servitude', 38: `servitude purchased for usucapion and for the use of the father of a family', 39: `practice of the servitude', 40: `extinction of the servitude', 41:' actions in defense of the servitude', 42: `some servitude concerning water', 43: `servitude of water intake or diversion', 44: `servitude of drains and leftovers of water', 45: ` communion', 46: `communion in general', 47: `condominium in buildings', 48: `possession', 49:  'effects of possession', 50: `rights and obligations of the owner to return the property', 51: `possession of movable property in good faith', 52: `usucaption', 53: `actions in defense of possession', 54: `denunciation of new work and feared damage'

\

}
\end{minipage} 
\\ \hline 
\end{tabular}
\end{table}

\begin{table}[t!]
\centering
\caption{\textit{(cont.)} Unique headings of book subdivisions originally provided in the ICC and used in the attribute-aware article prediction evaluation task}
\label{tab:attributes_book4}
\begin{tabular}{|c|l|l|}
\hline 
 & attributes & attributes (translated in English)\\
 \hline \hline  
\textsl{Book-4}  & 
\begin{minipage}[t]{0.42\columnwidth}%
\tiny
{1: `obbligazioni in generale', 2: `disposizioni preliminari', 3: `adempimento delle obbligazioni', 4: `adempimento in generale', 5: `pagamento con surrogazione', 6: `mora del creditore', 7: `inadempimento delle obbligazioni', 8: `modi di estinzione delle obbligazioni diversi dall'adempimento', 9: `novazione', 10: `remissione', 11: `compensazione', 12: `confusione', 13: `impossibilit\`{a} sopravvenuta per causa non imputabile al debitore', 14: `cessione dei crediti', 15: `delegazione, espromissione e accollo', 16: `alcune specie di obbligazioni', 17: `obbligazioni pecuniarie', 18: `obbligazioni alternative', 19: `obbligazioni in solido', 20: `obbligazioni divisibili e indivisibili', 21: `contratti in generale', 22: `requisiti del contratto', 23: `accordo delle parti', 24: `causa del contratto', 25: `oggetto del contratto', 26: `forma del contratto', 27: `condizione nel contratto', 28: `interpretazione del contratto', 29: `effetti del contratto', 30: `disposizioni generali', 31: `clausola penale e caparra', 32: `rappresentanza', 33: `contratto per persona da nominare', 34: `cessione del contratto', 35: `contratto a favore di terzi', 36: `simulazione', 37: `nullit\`{a} del contratto', 38: `annullabilit\`{a} del contratto', 39: `incapacit\`{a}', 40: `vizi del consenso', 41: `azione di annullamento', 42: `rescissione del contratto', 43: `risoluzione del contratto', 44: `risoluzione per inadempimento', 45: `impossibilit\`{a} sopravvenuta', 46: `eccessiva onerosit\`{a}', 47: `contratti del consumatore', 48: `singoli contratti', 49: `vendita', 50: `obbligazioni del venditore', 51: `obbligazioni del compratore', 52: `riscatto convenzionale', 53: `vendita di cose mobili', 54: `vendita dei beni di consumo', 55: `vendita con riserva di gradimento, a prova, a campione', 56: `vendita con riserva della propriet\`{a}', 57: `vendita su documenti e con pagamento contro documenti', 58: `vendita a termine di titoli di credito', 59: `vendita di cose immobili', 60: `vendita di eredit\`{a}', 61: `riporto', 62: `permuta', 63: `contratto estimatorio', 64: `somministrazione', 65: `locazione', 66: `locazione di fondi urbani', 67: `affitto', 68: `affitto di fondi rustici', 69: `affitto a coltivatore diretto', 70: `appalto', 71: `trasporto', 72: `trasporto di persone', 73: `trasporto di cose', 74: `mandato', 75: `obbligazioni del mandatario', 76: `obbligazioni del mandante', 77: `estinzione del mandato', 78: `commissione', 79: `spedizione', 80: `contratto di agenzia', 81: `mediazione', 82: `deposito', 83: `deposito in generale', 84: `deposito in albergo', 85: `deposito nei magazzini generali', 86: `sequestro convenzionale', 87: `comodato', 88: `mutuo', 89: `conto corrente', 90: `contratti bancari', 91: `depositi bancari', 92: `servizio bancario delle cassette di sicurezza', 93: `apertura di credito bancario', 94: `anticipazione bancaria', 95: `operazioni bancarie in conto corrente', 96: `sconto bancario', 97: `rendita perpetua', 98: `rendita vitalizia', 99: `assicurazione', 100: `assicurazione contro i danni', 101: `assicurazione sulla vita', 102: `riassicurazione', 103: `disposizioni finali', 104: `giuoco e scommessa', 105: `fideiussione', 106: `rapporti tra creditore e fideiussore', 107: `rapporti tra fideiussore e debitore principale', 108: `rapporti tra pi\`{u} fideiussori', 109: `estinzione della fideiussione', 110: `mandato di credito', 111: `anticresi', 112: `transazione', 113: `cessione dei beni ai creditori', 114: `promesse unilaterali', 115: `titoli di credito', 116: `titoli al portatore', 117: `titoli all'ordine', 118: `titoli nominativi', 119: `gestione di affari', 120: `pagamento dell'indebito', 121: `arricchimento senza causa', 122: `fatti illeciti'

\

}
\end{minipage}
& 
\begin{minipage}[t]{0.42\columnwidth}%
\tiny
{
1: `obligations in general', 2: `preliminary provisions', 3: `fulfillment of obligations', 4: `fulfillment in general', 5: `payment with subrogation', 6: `default of creditor', 7: `default of obligations', 8: `ways of extinguishing obligations other than performance', 9: `novation', 10: `remission', 11: `set-off', 12: `confusion', 13: `impossibility occurred for reasons not attributable to the debtor', 14: `assignment of credits', 15: `delegation, expression and assumption', 16: `some kinds of obligations', 17: `pecuniary obligations', 18: `alternative obligations', 19: `joint and several obligations', 20: `divisible and indivisible obligations', 21: `contracts in general', 22: `contract requirements', 23: `agreement of the parties', 24 : `cause of the contract', 25: `object of the contract', 26: `form of the contract', 27: `condition in the contract', 28: `interpretation of the contract', 29: `effects of the contract', 30: `general provisions', 31: `penalty clause and deposit', 32: `representation', 33: `contract for person to be appointed', 34: `transfer of the contract', 35: `contract in favor of third parties', 36: `simulation', 37: `nullity of the contract', 38: `voidability of the contract', 39: `incapacity', 40: `vices of consent', 41: `action for annulment' , 42: `termination of the contract', 43: `resolution of the contract', 44: `termination due to non-fulfillment', 45: `occurred impossibility', 46: `excessive  onerousness', 47: `consumer contracts', 48: `individual contracts', 49: `sale', 50: `seller's obligations', 51: `buyer's obligations', 52: `conventional redemption', 53: `sale of movable things', 54: `sale of consumer goods', 55: `sale subject to approval, trial, sample', 56: `sale subject to ownership', 57: `sale on documents and with payment against documents', 58: `forward sale of debt securities', 59: `sale of real estate', 60: `sale of inheritance', 61: `return', 62: `exchange', 63: `estimation contract', 64: `administration', 65: `lease', 66: `lease of urban land', 67:  'rent', 68: `rent of rustic land', 69: `rent to direct farmer', 70:  'contract', 71: `transport', 72: `transport of people', 73: `transport of things', 74: `warrant', 75: `agent's obligations', 76:  'principal's obligations', 77: `termination of the warrant', 78:  'commission', 79: `shipment', 80: `agency contract', 81:  'mediation', 82: `deposit', 83: `deposit in general', 84: `deposit in hotel', 85: `deposit in general warehouses', 86: `conventional confiscation', 87: `free loan', 88: `mortgage', 89: `checking  account', 90: `bank contracts', 91: `bank deposits', 92: `safe deposit box banking service', 93: `bank credit opening', 94: `bank advance', 95: `checking account banking', 96: `bank discount', 97: `perpetual income', 98: `life annuity', 99: `insurance', 100: `damage insurance', 101: `life insurance', 102: `reinsurance', 103: `final provisions', 104: `game and bet', 105: `surety', 106: `relationship between creditor and guarantor', 107: `relationship between guarantor and principal debtor', 108: `relationship between several guarantors', 109: `termination of surety', 110: `credit warrant', 111: `anticresi', 112: `transaction', 113: `transfer of assets to creditors', 114: `unilateral promises', 115: `debt securities', 116: `bearer securities', 117: `order securities', 118: `registered securities', 119: `business management', 120: `undue payment', 121: `causeless enrichment', 122: `illegal deeds'

\ 

}
\end{minipage} 
\\ \hline 
\end{tabular}
\end{table}

\begin{table}[t!]
\centering
\caption{\textit{(cont.)} Unique headings of book subdivisions originally provided in the ICC and used in the attribute-aware article prediction evaluation task}
\label{tab:attributes_book5}
\begin{tabular}{|c|l|l|}
\hline 
 & attributes & attributes (translated in English)\\
 \hline \hline  
\textsl{Book-5}  & 
\begin{minipage}[t]{0.42\columnwidth}%
\tiny
{1: `disciplina delle attivit\`{a} professionali', 2: `disposizioni generali', 3: `ordinanze corporative e accordi economici collettivi', 4: `contratto collettivo di lavoro e norme equiparate', 5: `lavoro nell'impresa', 6: `impresa in generale', 7: `imprenditore', 8: `collaboratori dell'imprenditore', 9: `rapporto di lavoro', 10: `costituzione del rapporto di lavoro', 11: `diritti e obblighi delle parti', 12: `previdenza e assistenza', 13: `estinzione del rapporto di lavoro', 14: `disposizioni finali', 15: `tirocinio', 16: `impresa agricola', 17: `mezzadria', 18: `colonia parziaria', 19: `soccida', 20: `soccida semplice', 21: `soccida parziaria', 22: `soccida con conferimento di pascolo', 23: `disposizione finale', 24: `imprese commerciali e altre imprese soggette a registrazione', 25: `registro delle imprese', 26: `obbligo di registrazione', 27: `disposizioni particolari per le imprese commerciali', 28: `rappresentanza', 29: `scritture contabili', 30: `insolvenza', 31: `lavoro autonomo', 32: `professioni intellettuali', 33: `lavoro subordinato in particolari rapporti', 34: `lavoro domestico', 35: `societ\`{a}', 36: `societ\`{a} semplice', 37: `rapporti tra i soci', 38: `rapporti con i terzi', 39: `scioglimento della societ\`{a}', 40: `scioglimento del rapporto sociale limitatamente a un socio', 41: `societ\`{a} in nome collettivo', 42: `societ\`{a} in accomandita semplice', 43: `societ\`{a} per azioni', 44: `costituzione per pubblica sottoscrizione', 45: `promotori e soci fondatori', 46: `patti parasociali', 47: `conferimenti', 48: `azioni e altri strumenti finanziari partecipativi', 49: `assemblea', 50: `amministrazione e controllo', 51: `amministratori', 52: `collegio sindacale', 53: `revisione legale dei conti', 54: `sistema dualistico', 55: `sistema monistico', 56: `obbligazioni', 57: `libri sociali', 58: `bilancio', 59: `modificazioni dello statuto', 60: `patrimoni destinati ad uno specifico affare', 61: `societ\`{a} con partecipazione dello stato o di enti pubblici', 62: `societ\`{a} di interesse nazionale', 63: `societ\`{a} in accomandita per azioni', 64: `societ\`{a} a responsabilit\`{a} limitata', 65: `conferimenti e quote', 66: `amministrazione della societ\`{a} e controlli', 67: `decisioni dei soci', 68: `modificazioni dell'atto costitutivo', 69: `scioglimento e liquidazione delle societ\`{a} di capitali', 70: `direzione e coordinamento di societ\`{a}', 71: `trasformazione, fusione e scissione', 72: `trasformazione', 73: `fusione delle societ\`{a}', 74: `scissione delle societ\`{a}', 75: `societ\`{a} costituite all'estero', 76: `societ\`{a} cooperative e mutue assicuratrici', 77: `societ\`{a} cooperative', 78: `disposizioni generali di societ\`{a} cooperative a mutualit\`{a} prevalente', 79: `costituzione', 80: `quote e azioni', 81: `organi sociali', 82: `controlli', 83: `mutue assicuratrici', 84: `associazione in partecipazione', 85: `azienda', 86: `ditta e insegna', 87: `marchio', 88: `diritti sulle opere dell'ingegno e sulle invenzioni industriali', 89: `diritto di autore sulle opere dell'ingegno letterarie e artistiche', 90: `diritto di brevetto per invenzioni industriali', 91: `diritto di brevetto per modelli di utilit\`{a} e di registrazione per disegni e modelli', 92: `disciplina della concorrenza e dei consorzi', 93: `disciplina della concorrenza', 94: `concorrenza sleale', 95: `consorzi per il coordinamento della produzione e degli scambi', 96: `consorzi con attivit\`{a} esterna', 97: `consorzi obbligatori', 98: `controlli dell'autorit\`{a} governativa', 99: `disposizioni penali in materia di societ\`{a}, di consorzi e di altri enti privati', 100: `falsit\`{a}', 101: `illeciti commessi dagli amministratori', 102: `illeciti commessi mediante omissione', 103: `altri illeciti, circostanze attenuanti e misure di sicurezza patrimoniali'

\

}
\end{minipage}
& 
\begin{minipage}[t]{0.42\columnwidth}%
\tiny
{
1: `regulation of professional activities', 2: `general provisions', 3: `corporate regulations and collective economic agreements', 4: `collective bargaining agreement and equivalent rules', 5: `work in enterprise', 6: `enterprise in general', 7: `entrepreneur', 8: `collaborators of the entrepreneur', 9: `employment relationship', 10: `establishment of the employment relationship', 11: `rights and obligations of the parties', 12: `social security and assistance', 13: `termination of the employment relationship', 14: `final provisions', 15: `apprenticeship', 16: `agricultural enterprise', 17: `sharecropping', 18: `partial colony', 19: `soccida', 20:  'simple soccida', 21: `partial soccida', 22: `soccida with granting of pasture', 23: `final provision', 24: `commercial enterprises and other enterprises subject to registration', 25: `business register', 26: `obligation of registration', 27: `special provisions for commercial enterprises', 28: `representation', 29: `accounting records', 30: `insolvency', 31: `self-employment', 32: `intellectual professions', 33: `subordinate work in particular relationships', 34: `domestic work', 35: `company', 36: `simple company', 37: `relationships between shareholders', 38: `relationships with third parties', 39: `dissolution of the company', 40: `dissolution of the social relationship limited to one shareholder', 41: `collective denomination company', 42: `limited partnership company', 43: `joint stock company', 44: `constitution by public subscription', 45: `promoters and founding partners', 46: `shareholders' agreements', 47: `contributions', 48: `shares and other equity financial instruments', 49: `shareholders' meeting', 50: `administration and control', 51: `directors', 52: `board of statutory auditors', 53: `statutory audit', 54: `two-tier system', 55: `one-tier system',  56: `bonds', 57: `company books', 58: `balance sheet', 59:  'amendments to the statute', 60: `assets intended for a specific business',  61: `company with state participation or public authorities', 62: `company of national interest', 63: `company  limited by shares', 64: `company  with limited responsibility', 
65: `contributions and shares', 66: `company administration and controls', 67: `shareholders' decisions', 68: `amendments to the articles of association', 69: `dissolution and liquidation of corporations', 70: `management and coordination of companies', 71: `transformation, merger and spin-off', 72: `transformation', 73:  'merger of companies', 74: `spin-off of companies', 75: `companies  incorporated abroad', 76: `cooperative companies and mutual insurance companies', 77: `cooperative companies', 
78:' general provisions of cooperative societies with prevalent mutuality', 79: `incorporation', 80: `quotas and shares', 81: `corporate bodies', 82: `controls', 83: `mutual insurance companies', 84: `association in participation', 85: `firm', 86: `firm and sign', 87: `trademark', 88: `intellectual property rights and industrial inventions', 89: `copyright on literary and artistic intellectual works', 90: `patent law for industrial inventions', 91: `patent law for utility models and registration for designs and models', 92: `competition and consortium regulations', 93: `competition rules', 94: `unfair competition', 95: `consortia for the coordination of production and trade', 96: `consortia with external activities', 97: `compulsory consortia', 98: `controls by the governmental authority', 99: `criminal provisions concerning companies, consortia and other private authorities', 100: `falsity', 101: `offenses committed by administrators', 102: `offenses  committed by omission', 103: `other offenses, mitigating circumstances and asset security measures'

\ 

}
\end{minipage} 
\\ \hline 
\end{tabular}
\end{table}

\begin{table}[t!]
\centering
\caption{\textit{(cont.)} Unique headings of book subdivisions originally provided in the ICC and used in the attribute-aware article prediction evaluation task}
\label{tab:attributes_book6}
\begin{tabular}{|c|l|l|}
\hline 
 & attributes & attributes (translated in English)\\
 \hline \hline  
\textsl{Book-6}  & 
\begin{minipage}[t]{0.42\columnwidth}%
\tiny
{1: `trascrizione', 2: `trascrizione degli atti relativi ai beni immobili', 3: `pubblicit\`{a} dei registri immobiliari e responsabilit\`{a} dei conservatori', 4: `trascrizione degli atti relativi ad alcuni beni mobili', 5: `trascrizione relativamente alle navi, agli aeromobili e agli autoveicoli', 6: `trascrizione relativamente ad altri beni mobili', 7: `prove', 8: `disposizioni generali', 9: `prova documentale', 10: `atto pubblico', 11: `scrittura privata', 12: `scritture contabili delle imprese soggette a registrazione', 13: `riproduzioni meccaniche', 14: `taglie o tacche di contrassegno', 15: `copie degli atti', 16: `atti di ricognizione o di rinnovazione', 17: `prova testimoniale', 18: `presunzioni', 19: `confessione', 20: `giuramento', 21: `responsabilit\`{a} patrimoniale, cause di prelazione e conservazione della garanzia patrimoniale', 22: `privilegi', 23: `privilegi sui mobili', 24: `privilegi generali sui mobili', 25: `privilegi sopra determinati mobili', 26: `privilegi sopra gli immobili', 27: `ordine dei privilegi', 28: `pegno', 29: `pegno dei beni mobili', 30: `pegno di crediti e altri diritti', 31: `ipoteche', 32: `ipoteca legale', 33: `ipoteca giudiziale', 34: `ipoteca volontaria', 35: `iscrizione e rinnovazione delle ipoteche', 36: `iscrizione', 37: `rinnovazione', 38: `ordine delle ipoteche', 39: `effetti dell'ipoteca rispetto al terzo acquirente', 40: `effetti dell'ipoteca rispetto al terzo datore', 41: `riduzione delle ipoteche', 42: `estinzione delle ipoteche', 43: `cancellazione dell'iscrizione', 44: `modo di liberare i beni dalle ipoteche', 45: `rinunzia e astensione del creditore nell'espropriazione forzata', 46: `mezzi di conservazione della garanzia patrimoniale', 47: `azione surrogatoria', 48: `azione revocatoria', 49: `sequestro conservativo', 50: `tutela giurisdizionale dei diritti', 51: `esecuzione forzata', 52: `espropriazione', 53: `effetti del pignoramento', 54: `effetti della vendita forzata e dell'assegnazione', 55: `espropriazione di beni oggetto di vincoli di indisponibilit\`{a} o di alienazioni a titolo gratuito', 56: `esecuzione forzata in forma specifica', 57: `prescrizione e decadenza', 58: `prescrizione', 59: `sospensione della prescrizione', 60: `interruzione della prescrizione', 61: `termine della prescrizione', 62: `prescrizione ordinaria', 63: `prescrizioni brevi', 64: `prescrizioni presuntive', 65: `computo dei termini', 66: `decadenza'

\

}
\end{minipage}
& 
\begin{minipage}[t]{0.42\columnwidth}%
\tiny
{
1: `transcription', 2: `transcription of deeds relating to immovable properties', 3: `advertising of real estate registers and liability of conservators', 4: `transcription of deeds relating to some movable property', 5: `transcription relating to ships, aircraft and motor vehicles', 6: `transcription relating to other movable properties', 7: `evidence', 8: `general provisions', 9: `documentary evidence', 10: `public deed', 11: `private deed', 12: `accounting records of companies subject to registration', 13: `mechanical reproductions', 14: `sizes or mark notches', 15: `copies of deeds', 16: `acts of recognition or renewal', 17: `witness evidence', 18: `presumptions', 19: `confession', 20: `oath', 21: `property liability, causes of preemption and retention of the patrimonial guarantee', 22: `privileges', 23: `privileges on movables', 24: `general privileges on movables', 25: `privileges over certain movables', 26: `privileges over real estate', 27: `order of privileges', 28:  'pledge', 29: `pledge of movable property', 30: `pledge of credits and other rights', 31: `mortgages', 32: `legal mortgage', 33:  'judicial mortgage', 34: `voluntary mortgage', 35: `registration and renewal of mortgages', 36: `registration', 37: `renewal', 38: `mortgage order', 39: `effects of 'mortgage with respect to the third buyer', 40: `effects of the mortgage with respect to the third employer', 41: `reduction of mortgages', 42: `repayment of mortgages', 43: `cancellation of registration', 44: `way to release the assets from mortgages', 45: `renunciation and abstention of the creditor in forced expropriation', 46: `means of preserving the patrimonial guarantee', 47: `subrogation action', 48: `revocatory action', 49: `conservative seizure', 50: `judicial protection of rights', 51: `forced execution ', 52: `expropriation', 53: `effects of foreclosure', 54: `effects of forced sale and assignment', 55: `expropriation of assets subject to restrictions of unavailability or free disposal', 56: `specific forced execution', 57: `prescription and forfeiture', 58: `prescription', 59: `suspension of prescription', 60: `termination of prescription', 61: `limitation period of prescription', 62: `ordinary prescription', 63: `short prescription', 64: `presumptive prescription', 65: `calculation of time limits', 66: `forfeiture'

\

}
\end{minipage} 
\\ \hline 
\end{tabular}
\end{table}


\begin{thebibliography}{10}
\providecommand{\url}[1]{\texttt{#1}}
\providecommand{\urlprefix}{URL }
\providecommand{\doi}[1]{https://doi.org/#1}

\bibitem{Aletras2016}
Aletras, N., Tsarapatsanis, D., Preotiuc-Pietro, D., Lampos, V.: Predicting
  judicial decisions of the european court of human rights: a natural language
  processing perspective. PeerJ Computer Science  \textbf{2}, ~e93 (2016)

\bibitem{BahdanauCB14}
Bahdanau, D., Cho, K., Bengio, Y.: Neural machine translation by jointly
  learning to align and translate. In: Proc. ICLR (2015)

\bibitem{xClass}
Bengio, S., Dembczynski, K., Joachims, T., Kloft, M., Varma, M.: Extreme
  classification. Tech. rep., Report from Dagstuhl Seminar 18291 (2019).
  \doi{10.4230/DagRep.8.7.62}

\bibitem{Boella2011}
Boella, G., Caro, L.D., Humphreys, L.: {Using classification to support legal
  knowledge engineers in the Eunomos legal document management system}. In:
  Proc. Workshop on Juris-informatics (JURISIN) (2011)

\bibitem{BrantingWBPCFPY19}
Branting, K., Weiss, B., Brown, B., Pfeifer, C., Chakraborty, A., Ferro, L.,
  Pfaff, M., Yeh, A.S.: Semi-supervised methods for explainable legal
  prediction. In: Proc. Int. Conf. on Artificial Intelligence and Law (ICAIL).
  pp. 22--31 (2019)

\bibitem{BrantingYWMB17}
Branting, L.K., Yeh, A.S., Weiss, B., Merkhofer, E.M., Brown, B.: Inducing
  predictive models for decision support in administrative adjudication. In:
  {AI} Approaches to the Complexity of Legal Systems - {AICOL} Workshops
  2015-2017, vol. 10791, pp. 465--477 (2017)

\bibitem{ChalkidisAA19}
Chalkidis, I., Androutsopoulos, I., Aletras, N.: Neural legal judgment
  prediction in english. In: Proc. ACL. pp. 4317--4323. Association for
  Computational Linguistics (2019)

\bibitem{ChalkidisAM18}
Chalkidis, I., Androutsopoulos, I., Michos, A.: {Obligation and Prohibition
  Extraction Using Hierarchical RNNs}. In: Proc. Annual Meeting of the
  Association for Computational Linguistics ({ACL}). pp. 254--259 (2018)

\bibitem{Chalkidis-Extreme}
Chalkidis, I., Fergadiotis, M., Malakasiotis, P., Aletras, N., Androutsopoulos,
  I.: {Extreme Multi-Label Legal Text Classification: A case study in EU
  Legislation}. In: Proc. Natural Legal Language Processing (NLLP) Workshop of
  NAACL-HLT. pp. 78–--87 (2019)

\bibitem{legalBERT}
Chalkidis, I., Fergadiotis, M., Malakasiotis, P., Aletras, N., Androutsopoulos,
  I.: {LEGAL-BERT:} the muppets straight out of law school. CoRR
  \textbf{abs/2010.02559} (2020)

\bibitem{ChalkidisK19}
Chalkidis, I., Kampas, D.: Deep learning in law: early adaptation and legal
  word embeddings trained on large corpora. Artif. Intell. Law  \textbf{27}(2),
   171--198 (2019)

\bibitem{ConradB18}
Conrad, J.G., Branting, L.K.: Introduction to the special issue on legal text
  analytics. Artif. Intell. Law  \textbf{26}(2),  99--102 (2018)

\bibitem{Dadgostari2020}
Dadgostari, F., Guim, M., Beling, P., Livermore, M.A., Rockmore, D.: Modeling
  law search as prediction. Artif. Intell. Law  \textbf{29}(1),  3--34 (2021)

\bibitem{DevlinCLT19}
Devlin, J., Chang, M., Lee, K., Toutanova, K.: {BERT:} pre-training of deep
  bidirectional transformers for language understanding. In: Proc. NAACL-HLT.
  pp. 4171--4186 (2019)

\bibitem{DhillonM01}
Dhillon, I.S., Modha, D.S.: Concept decompositions for large sparse text data
  using clustering. Mach. Learn.  \textbf{42}(1/2),  143--175 (2001)

\bibitem{DoNTNN17}
Do, P., Nguyen, H., Tran, C., Nguyen, M., Nguyen, M.: Legal question answering
  using ranking {SVM} and deep convolutional neural network. CoRR
  \textbf{abs/1703.05320} (2017)

\bibitem{DuH18}
Du, C., Huang, L.: Text classification research with attention-based recurrent
  neural networks. Int. J. Comput. Commun. Control  \textbf{13}(1),  50--61
  (2018)

\bibitem{GanKYW21}
Gan, L., Kuang, K., Yang, Y., Wu, F.: Judgment prediction via injecting legal
  knowledge into neural networks. In: Proc. {AAAI}. pp. 12866--12874. {AAAI}
  Press (2021)

\bibitem{Goldberg17}
Goldberg, Y.: Neural network methods in natural language processing. Morgan and
  Claypool Publishers (2017)

\bibitem{Goodfellow16}
Goodfellow, I., Y, Y.B., Courville, A.: Deep learning. MIT Press, Cambridge
  (2016)

\bibitem{HackerKGN20}
Hacker, P., Krestel, R., Grundmann, S., Naumann, F.: Explainable {AI} under
  contract and tort law: legal incentives and technical challenges. Artif.
  Intell. Law  \textbf{28}(4),  415--439 (2020)

\bibitem{HuLT0S18}
Hu, Z., Li, X., Tu, C., Liu, Z., Sun, M.: Few-shot charge prediction with
  discriminative legal attributes. In: Proc. COLING. pp. 487--498. Association
  for Computational Linguistics (2018)

\bibitem{JainD88}
Jain, A.K., Dubes, R.C.: Algorithms for Clustering Data. Prentice-Hall (1988)

\bibitem{Jones04}
Jones, K.S.: A statistical interpretation of term specificity and its
  application in retrieval. J. Documentation  \textbf{60}(5),  493--502 (2004)

\bibitem{KimXG15}
Kim, M., Xu, Y., Goebel, R.: A convolutional neural network in legal question
  answering. In: Proc. Int. Workshop on Juris-informatics {JURISIN} (2015)

\bibitem{TextCNN}
Kim, Y.: Convolutional neural networks for sentence classification. In: Proc.
  EMNLP. pp. 1746–--1751 (2014)

\bibitem{TextRCNN}
Lai, S., Xu, L., Liu, K., Zhao, J.: Recurrent convolutional neural networks for
  text classification. In: Proc. AAAI. pp. 2267--–2273 (2015)

\bibitem{LiZYGF19}
Li, S., Zhang, H., Ye, L., Guo, X., Fang, B.: {MANN:} {A} multichannel
  attentive neural network for legal judgment prediction. {IEEE} Access
  \textbf{7},  151144--151155 (2019)

\bibitem{Lin2012}
Lin, W., Kuo, T., Chang, T., Yen, C., Chen, C., Lin, S.: Exploiting machine
  learning models for chinese legal documents labeling, case classification,
  and sentencing prediction. IJCLCLP  \textbf{17}(4) (2012)

\bibitem{Liu2006}
Liu, C., Hsieh, C.: Exploring phrase-based classification of judgment documents
  for criminal charges in chinese. In: Proc. ISMIS. pp. 681–--690 (2006)

\bibitem{BiLSTM}
Liu, P., Qiu, X., Huang, X.: Recurrent neural network for text classification
  with multi-task learning. In: Proc. IJCAI. pp. 2873–--2879 (2016)

\bibitem{LIU2015194}
Liu, Y.H., Chen, Y.L., Ho, W.L.: Predicting associated statutes for legal
  problems. Information Processing \& Management  \textbf{51}(1),  194--211
  (2015)

\bibitem{LongT0S19}
Long, S., Tu, C., Liu, Z., Sun, M.: Automatic judgment prediction via legal
  reading comprehension. In: Proc. Chinese Computational Linguistics. Lecture
  Notes in Computer Science, vol. 11856, pp. 558--572 (2019)

\bibitem{Luo2017}
Luo, B., Feng, Y., Xu, J., Zhang, X., Zhao, D.: Learning to predict charges for
  criminal cases with legal basis. In: Proc. EMNLP. pp. 2727–--2736 (2017)

\bibitem{tsne}
van~der Maaten, L., Hinton, G.: {Visualizing High-Dimensional Data Using
  t-SNE}. Journal of Machine Learning Research  \textbf{9},  2579--2605 (2008)

\bibitem{MedvedevaVW20}
Medvedeva, M., Vols, M., Wieling, M.: Using machine learning to predict
  decisions of the european court of human rights. Artif. Intell. Law
  \textbf{28}(2),  237--266 (2020)

\bibitem{MorimotoKSSM17}
Morimoto, A., Kubo, D., Sato, M., Shindo, H., Matsumoto, Y.: Legal question
  answering system using neural attention. In: Proc. 4th ICAIL Competition on
  Legal Information Extraction and Entailment ({COLIEE}). EPiC Series in
  Computing, vol.~47, pp. 79--89 (2017)

\bibitem{NallapatiM08}
Nallapati, R., Manning, C.D.: {Legal Docket Classification: Where Machine
  Learning Stumbles}. In: Proc. EMNLP. pp. 438--446 (2008)

\bibitem{NandaACBR17}
Nanda, R., Adebayo, K.J., Caro, L.D., Boella, G., Robaldo, L.: Legal
  information retrieval using topic clustering and neural networks. In: Proc.
  4th ICAIL Competition on Legal Information Extraction and Entailment
  ({COLIEE}). EPiC Series in Computing, vol.~47, pp. 68--78 (2017)

\bibitem{abs-2106-13405}
Nguyen, H., Nguyen, P.M., Vuong, T., Bui, Q.M., Nguyen, C.M., Dang, T.B., Tran,
  V., Nguyen, M.L., Satoh, K.: {JNLP} team: Deep learning approaches for legal
  processing tasks in {COLIEE} 2021. CoRR  \textbf{abs/2106.13405} (2021),
  \url{https://arxiv.org/abs/2106.13405}

\bibitem{NguyenNTSS17}
Nguyen, T., Nguyen, L., Tojo, S., Satoh, K., Shimazu, A.: {Single and multiple
  layer BI-LSTM-CRF for recognizing requisite and effectuation parts in legal
  texts}. In: Proc. ICAL Workshops (2017)

\bibitem{NguyenNTSS18}
Nguyen, T., Nguyen, L., Tojo, S., Satoh, K., Shimazu, A.: Recurrent neural
  network-based models for recognizing requisite and effectuation parts in
  legal texts. Artif. Intell. Law  \textbf{26}(2),  169--199 (2018)

\bibitem{ONeillBRO17}
O'Neill, J., Buitelaar, P., Robin, C., O'Brien, L.: Classifying sentential
  modality in legal language: a use case in financial regulations, acts and
  directives. In: Proc. Int. Conf. on Artificial Intelligence and Law (ICAIL).
  pp. 159--168 (2017)

\bibitem{Peters2018}
Peters, M.E., Neumann, M., Iyyer, M., Gardner, M., Clark, C., Lee, K.,
  Zettlemoyer, L.: Deep contextualized word representations. In: Proc.
  NAACL-HLT. pp. 2227--2237 (2018)

\bibitem{PolignanoBGSB19}
Polignano, M., Basile, P., de~Gemmis, M., Semeraro, G., Basile, V.: {AlBERTo:
  Italian {BERT} Language Understanding Model for {NLP} Challenging Tasks Based
  on Tweets}. In: Proc. 6th Italian Conf. on Computational Linguistics. {CEUR}
  Workshop Proceedings, vol.~2481. CEUR-WS.org (2019)

\bibitem{PuccinelliDD19}
Puccinelli, D., Demartini, S., D'Aoust, R.E.: Fixing comma splices in italian
  with {BERT}. In: Proc. 6th Italian Conf. on Computational Linguistics. {CEUR}
  Workshop Proceedings, vol.~2481. CEUR-WS.org (2019)

\bibitem{RabeloKG19}
Rabelo, J., Kim, M., Goebel, R.: Combining similarity and transformer methods
  for case law entailment. In: Proc. Int. Conf. on Artificial Intelligence and
  Law (ICAIL). pp. 290--296 (2019)

\bibitem{RabeloKGYKS20}
Rabelo, J., Kim, M., Goebel, R., Yoshioka, M., Kano, Y., Satoh, K.: {COLIEE}
  2020: Methods for legal document retrieval and entailment. In: Proc. of the
  JSAI-isAI 2020 Workshops, JURISIN, {LENLS} 2020 Workshops. New Frontiers in
  Artificial Intelligence, Lecture Notes in Computer Science, vol. 12758, pp.
  196--210. Springer (2020)

\bibitem{GPT}
Radford, A., Sutskever, I.: Improving language understanding by generative
  pre-training. In: arxiv (2018)

\bibitem{SanchezHMAML20}
Sanchez, L., He, J., Manotumruksa, J., Albakour, D., Martinez, M., Lipani, A.:
  Easing legal news monitoring with learning to rank and {BERT}. In: Proc.
  ECIR. Lecture Notes in Computer Science, vol. 12036, pp. 336--343. Springer
  (2020)

\bibitem{ShaoMLMSZM20}
Shao, Y., Mao, J., Liu, Y., Ma, W., Satoh, K., Zhang, M., Ma, S.: {BERT-PLI:}
  modeling paragraph-level interactions for legal case retrieval. In: Proc.
  IJCAI. pp. 3501--3507 (2020)

\bibitem{Sulea2017}
Sulea, O., Zampieri, M., Malmasi, S., Vela, M., Dinu, L.P., van Genabith, J.:
  Exploring the use of text classification in the legal domain. In: Proc. ICAIL
  Workshop on Automated Semantic Analysis of Information in Legal Texts (2017)

\bibitem{VaswaniSPUJGKP17}
Vaswani, A., Shazeer, N., Parmar, N., Uszkoreit, J., Jones, L., Gomez, A.N.,
  Kaiser, L., Polosukhin, I.: Attention is all you need. In: Proc. NIPS. pp.
  5998--6008 (2017)

\bibitem{Viola17}
Viola, L.: Interpretazione della legge con modelli matematici, vol.~26. Diritto
  Avanzato (2017)

\bibitem{WangYNZZN18}
Wang, P., Yang, Z., Niu, S., Zhang, Y., Zhang, L., Niu, S.: Modeling dynamic
  pairwise attention for crime classification over legal articles. In: Proc.
  {ACM} {SIGIR}. pp. 485--494 (2018)

\bibitem{YamakoshiKOT19}
Yamakoshi, T., Komamizu, T., Ogawa, Y., Toyama, K.: Japanese mistakable legal
  term correction using infrequency-aware {BERT} classifier. In: Proc. {IEEE}
  Int. Conf. on Big Data. pp. 4342--4351 (2019)

\bibitem{YangJZL19}
Yang, W., Jia, W., Zhou, X., Luo, Y.: Legal judgment prediction via
  multi-perspective bi-feedback network. In: Proc. IJCAI. pp. 4085--4091 (2019)

\bibitem{YangYDHSH16}
Yang, Z., Yang, D., Dyer, C., He, X., Smola, A.J., Hovy, E.H.: Hierarchical
  attention networks for document classification. In: Proc {NAACL}-{HLT}. pp.
  1480--1489. The Association for Computational Linguistics (2016)

\bibitem{Ye2018}
Ye, H., Jiang, X., Luo, Z., Chao, W.: Interpretable charge predictions for
  criminal cases: Learning to generate court views from fact descriptions. In:
  Proc. NAACL-HLT. pp. 1854--–1864 (2018)

\bibitem{0001KYS17}
Yin, W., Kann, K., Yu, M., Sch{\"{u}}tze, H.: Comparative study of {CNN} and
  {RNN} for natural language processing. CoRR  \textbf{abs/1702.01923} (2017)

\bibitem{YoshiokaAS21}
Yoshioka, M., Aoki, Y., Suzuki, Y.: {BERT}-based ensemble methods with data
  augmentation for legal textual entailment in {COLIEE} statute law task. In:
  Proc. of the 18th International Conference for Artificial Intelligence and
  Law (ICAIL). pp. 278--284. {ACM} (2021)

\bibitem{ZhaoK04}
Zhao, Y., Karypis, G.: Empirical and theoretical comparisons of selected
  criterion functions for document clustering. Mach. Learn.  \textbf{55}(3),
  311--331 (2004)

\bibitem{BiLSTMA}
Zhou, P., Shi, W., Tian, J., Qi, Z., Li, B., Hao, H., Xu, B.: Attention-based
  bidirectional long short-term memory networks for relation classification.
  In: Proc. ACL (2016)

\bibitem{ZhouZLSS19}
Zhou, X., Zhang, Y., Liu, X., Sun, C., Si, L.: Legal intelligence for
  e-commerce: Multi-task learning by leveraging multiview dispute
  representation. In: Proc. ACM SIGIR. pp. 315--324 (2019)

\end{thebibliography}
\end{document}